%% file: arxiv_accepted__3_.tex
\documentclass[10pt]{article} 
\usepackage[preprint]{tmlr}
\input{math_commands.tex}

\usepackage{hyperref}
\hypersetup{hidelinks}
\usepackage{url}
\usepackage{multirow}
\usepackage{booktabs}
\usepackage{amsmath,amssymb,amsfonts}
\usepackage{algorithmic}
\usepackage{graphicx}
\usepackage{textcomp}
\usepackage{pgf-pie}
\usepackage[dvipsnames]{xcolor}
\usepackage{array,tabularx,enumitem}
\usepackage{placeins}
\usepackage[utf8]{inputenc}
\usepackage{grffile}
\usepackage{subcaption}
\usepackage[percent]{overpic}
\usepackage[export]{adjustbox}
\usepackage[table]{xcolor}
\usepackage{comment}
\usepackage{ragged2e}
\usepackage{microtype}

\newcolumntype{M}[1]{>{\centering\arraybackslash}m{#1}}
\newcolumntype{C}[1]{>{\centering\arraybackslash}m{#1}}
\newcolumntype{L}[1]{>{\raggedright\arraybackslash}m{#1}}

\title{Foundation Models in Robotics: A Comprehensive Review of Methods, Models, Datasets, Challenges and Future Research Directions}

\author{%
\name Aggelos Psiris \email aggelospsiris@hua.gr \\
\addr Department of Informatics and Telematics, Harokopio University of Athens, Athens, Greece
\AND
\name Vasileios Argyriou \email Vasileios.Argyriou@kingston.ac.uk \\
\addr Department of Networks and Digital Media, Kingston University, London, United Kingdom
\AND
\name Evangelos K. Markakis \email emarkakis@hmu.gr \\
\addr Department of Electrical and Computer Engineering, Hellenic Mediterranean University, Heraklion, Greece
\AND
\name Panagiotis Sarigiannidis \email psarigiannidis@uowm.gr \\
\addr Department of Electrical and Computer Engineering, University of Western Macedonia, Kozani, Greece
\AND
\name Efstratios Gavves \email e.gavves@uva.nl \\
\addr University of Amsterdam, Amsterdam, Netherlands \\
\addr Archimedes, Athena Research Center, Athens, Greece
\AND
\name Kostas Bekris \email kb572@cs.rutgers.edu \\
\addr Computer Science Department, Rutgers University, New Brunswick, NJ, USA
\AND
\name Arash Ajoudani \email Arash.Ajoudani@iit.it \\
\addr Istituto Italiano di Tecnologia, Genova, Italy
\AND
\name Georgios Th. Papadopoulos \email g.th.papadopoulos@hua.gr \\
\addr Department of Informatics and Telematics, Harokopio University of Athens, Athens, Greece \\
\addr Archimedes, Athena Research Center, Athens, Greece
}

\def\month{07}  
\def\year{2026} 
\def\openreview{\url{https://openreview.net/forum?id=qF7vdPrpPk}} 

\begin{document}

\maketitle

\begin{abstract}
Over the recent years, the field of robotics has been undergoing a transformative paradigm shift from fixed, single-task, domain-specific solutions towards adaptive, multi-function, general-purpose agents, capable of operating in complex, open-world, dynamic environments. This tremendous advancement is primarily driven by the emergence of Foundation Models (FMs), i.e., large-scale neural-network architectures trained on massive, internet-scale, heterogeneous datasets that provide unprecedented capabilities in multi-modal understanding/reasoning, long-horizon planning, and cross-embodiment generalization. In this context, the current study provides a holistic, thorough, systematic, and in-depth review of the research landscape of FMs in robotics. In particular, the evolution in the field is initially delineated through five distinct research phases, spanning from the early incorporation of native Natural Language Processing (NLP) and Computer Vision (CV) models to the current frontier of multi-sensory generalization and real-world deployment. Subsequently, a highly-granular, multi-criteria, taxonomic investigation of the literature methods is performed, examining the following key aspects: a) The employed foundation model types (i.e., LLMs, VFMs, VLMs, and VLAs), b) The underlying neural network architectures, c) The adopted learning paradigms, d) The different learning stages of knowledge incorporation, e) The most common robotic tasks (including perception, planning, navigation, manipulation, and human-robot interaction), and f) The main real-world application domains. For each defined criterion/aspect, a methodical comparative analysis of the various categories of approaches and critical insights are provided. Moreover, a synthesis of the publicly available datasets, required for model training and evaluation, is provided, organized around the main recurring dataset families along with their typical uses and current gaps. Furthermore, a comprehensive and hierarchical discussion on the current open challenges and promising future research directions in the field is incorporated.

\end{abstract}

\section{Introduction}
\label{sec:Intro}

Over the recent years, the field of robotics has witnessed unprecedented and transformative technological advancements, which have boosted the evolution from fixed, single-task, domain-specific setups to adaptive, general-purpose, open-world solutions \citep{newbury2023deep,mokssit2023deep}. Apart from significant developments in hardware and material sciences, a key driving force for this technological rise comprises the progress in the fields of Artificial Intelligence (AI) and Machine Learning (ML) \citep{soori2023artificial}. In particular, nowadays robotic platforms exhibit increased levels of efficiency, dexterity, autonomy, precision, and adaptation across a wide set of diverse tasks, while operating in complex and dynamic environments \citep{billard2025roadmap}.

Robotic research has so far been dominated by two main (though not mutually exclusive) modeling paradigms, namely automatic control and machine learning approaches \citep{lee2024robot}. Classic automation control relies on the fundamental principle of initially defining a mathematical model of a system for predicting its behavior and, subsequently, designing a controller for enabling it to perform a specific task. Such approaches (often termed model-based) require explicit programming and have been proven to be efficient for implementing tasks in structured and predictable environments \citep{lee2024robot}. However, they are characterized by low adaptability (reprogramming is needed) and they are typically mathematically complex \citep{Rakhmatillaev_2025}. On the other hand, ML methods focus on enabling robots to learn from data and experiences. To this end, ML approaches are shown to exhibit high adaptability (e.g., tackling novel and previously unseen circumstances) and to be efficient in handling tasks in complex, unstructured, and dynamic (or even unknown) environments \citep{tang2025deep}. Nevertheless, ML methods are often computationally expensive and typically require large datasets for training purposes \citep{hu2023toward}. 

Foundation Models (FMs) constitute a recent and, yet, very powerful paradigm in the fields of AI and ML \citep{awais2025foundation}. In particular, FMs are constructed through training on massive, internet-scale, multi-modal datasets and can be adapted to a wide range of diverse downstream tasks, such as language, vision, and audio processing. In practice, FMs serve as a versatile and reusable basis for efficiently developing specialized or domain-specific (multi-task) applications, avoiding the need for training from scratch and using extensive training datasets. More recently, FMs have also been introduced in the field of robotics, exhibiting the following, among others, advantageous characteristics (that extend the ones of traditional ML-based approaches) \citep{firoozi2025foundation,xiao2025robot}: a) Improved transferability across related tasks, environments, and embodiments, b) More reusable and generalizable representations, c) Increased semantic understanding and open-world capabilities, d) Support for sim-to-real transfer and cross-domain adaptation, e) Multi-modal integration and semantic alignment, f) Enhanced versatility in language-conditioned and perception-driven robotic behavior, g) Interpretation of natural language instructions, h) Hierarchical and long-horizon task decomposition and planning, and i) Improved policy generalization. The above favorable attributes, though, are accompanied  by critical/unique challenges, including, indicatively \citep{firoozi2025foundation,xiao2025robot}: a) Inference latency and high computational cost, b) Limited real-time deployability, c) Lack of semantic and physical grounding, d) Data scarcity and embodiment bias, e) Safety risks and unforeseen failure modes, f) Limited interpretability, transparency, and diagnosability, and g) Ethical, alignment, and regulatory imperatives.

\newcommand{\headerbreak}[1]{%
  \begin{minipage}[c][0.7cm][c]{\linewidth}
    \centering\textbf{#1}
  \end{minipage}%
}

\newlist{tabitem}{itemize}{1}
\setlist[tabitem]{
  leftmargin=*,
  label=\textbullet,
  labelsep=0.0em,
  itemsep=0pt,
  parsep=0pt,
  topsep=0pt,
  partopsep=0pt,
  nosep,
  before=\vspace{0pt}\justifying,
  after=\vspace{-0.7em}
}

\begin{table}[t]
  \caption{Comparative analysis of recent surveys in the field of foundation models in robotics.}
  \label{tab:fm_surveys_rev}
  \centering
  \scriptsize

  \setlength{\aboverulesep}{0pt}
  \setlength{\belowrulesep}{0pt}
  \setlength{\tabcolsep}{2pt}
  \renewcommand{\arraystretch}{0.9}

  \setlist*[tabitem]{before=\vspace{2.2pt}\justifying, after=\vspace{2.2pt}}

  \rowcolors{2}{gray!25}{white}

  \begin{tabular}{@{}|
    >{\justifying\arraybackslash}m{2.5cm}|
    >{\raggedright\arraybackslash}m{5.2cm}|
    >{\raggedright\arraybackslash}m{4.8cm}|
  @{}}
    \toprule
    \rowcolor{gray!40}
    \headerbreak{Article} &
    \headerbreak{Strengths} &
    \headerbreak{Limitations} \\
    \midrule

    \citet{hu2023toward} (arXiv) &
    \begin{tabitem}
      \item Broad scope
      \item Empirical per-task performance meta-analysis
      \item CV/NLP vs.\ native FM separation
    \end{tabitem} &
    \begin{tabitem}
      \item Early works only
      \item No systematic comparison
      \item Misses recent advances
      \item No application domains
    \end{tabitem} \\ \midrule

    \citet{xu2024survey} (arXiv) &
    \begin{tabitem}
      \item Manipulation focus
      \item Planning/control taxonomy
    \end{tabitem} &
    \begin{tabitem}
      \item Narrow scope
      \item Limited coverage
      \item No systematic comparison
      \item No empirical benchmarking
      \item No application domains
    \end{tabitem} \\ \midrule

    \citet{ma2024survey} (arXiv) &
    \begin{tabitem}
      \item VLA/embodied-AI focus
      \item Systematic per-category comparison
    \end{tabitem} &
    \begin{tabitem}
      \item Only VLAs
      \item Not multi-criteria
      \item No empirical benchmarking
      \item No application domains
    \end{tabitem} \\ \midrule

    \citet{jang2024unlocking} (IJCAS) &
    \begin{tabitem}
      \item Autonomy focus
      \item Perception/planning/control taxonomy
      \item Platforms \& simulators
    \end{tabitem} &
    \begin{tabitem}
      \item Not multi-criteria
      \item No theoretical comparison
      \item No empirical benchmarking
      \item No application domains
    \end{tabitem} \\ \midrule

    \citet{kawaharazuka2024real} (AR) &
    \begin{tabitem}
      \item FM-component replacement focus
      \item Input-output analysis
    \end{tabitem} &
    \begin{tabitem}
      \item Not multi-criteria
      \item No theoretical comparison
      \item No empirical benchmarking
      \item No application domains
    \end{tabitem} \\ \midrule

    \citet{firoozi2025foundation} (IJRR) &
    \begin{tabitem}
      \item Broad robotics \& embodied-AI scope
      \item Decision/planning/control analysis
    \end{tabitem} &
    \begin{tabitem}
      \item Not multi-criteria
      \item No theoretical comparison
      \item No empirical benchmarking
      \item No application domains
    \end{tabitem} \\ \midrule

    \citet{xiao2025robot} (Neurocomputing) &
    \begin{tabitem}
      \item Robot-learning focus
      \item Systematic hierarchical analysis
    \end{tabitem} &
    \begin{tabitem}
      \item Not multi-criteria
      \item No theoretical comparison
      \item No empirical benchmarking
      \item No application domains
    \end{tabitem} \\ \midrule

    \citet{tayyab2025foundation} (arXiv) &
    \begin{tabitem}
      \item Deployment/integration focus
      \item Sim-to-real
      \item Feasibility analysis
    \end{tabitem} &
    \begin{tabitem}
      \item Not multi-criteria
      \item No theoretical comparison
      \item No empirical benchmarking
      \item No application domains
    \end{tabitem} \\ \midrule

    \citet{kawaharazuka2025vision} (IEEE Access) &
    \begin{tabitem}
      \item VLA full-stack (hardware \& software)
      \item Architecture \& learning analysis
    \end{tabitem} &
    \begin{tabitem}
      \item Only VLAs
      \item Coarse categorization
      \item No theoretical comparison
      \item No empirical benchmarking
      \item No application domains
    \end{tabitem} \\ \midrule

    \citet{sapkota2025vision} (arXiv) &
    \begin{tabitem}
      \item VLA evolution focus
      \item Application domains covered
    \end{tabitem} &
    \begin{tabitem}
      \item Only VLAs
      \item Not multi-criteria
      \item No theoretical comparison
      \item No empirical benchmarking
    \end{tabitem} \\ \midrule

    \textbf{Current survey} &
    \begin{tabitem}
      \item Holistic \& systematic review (6 databases, screening)
      \item 5 research evolution phases
      \item 6-criteria taxonomy
      \item Per-criterion comparison \& insights
      \item Challenges \& future directions
    \end{tabitem} &
    \begin{tabitem}
      \item No empirical benchmarking
      \item Snapshot of a fast-evolving field
    \end{tabitem} \\

    \bottomrule
  \end{tabular}%
\end{table}

As outlined above, FMs induce transformative effects on and lead to unprecedented performance/capability accomplishments in the field of robotics, fundamentally reforming robot design, learning, programming, and deployment practices. In this context, the current study aims to holistically and comprehensively investigate, map, and analyze in depth the research landscape of robotic FM methods. In particular, the main contributions of this work are:
\begin{itemize}
    \item Outline of the \textbf{evolution in robotic FM research}, focusing on describing the most common FMs proposed in the literature and the main observed phases, which comprise the following ones: a) Phase 1 (2018-2021): Integration of native Natural Language Processing (NLP) and Computer Vision (CV) models; b) Phase 2 (2021-2022): Grounded planning with Vision-Language (VL) representations; c) Phase 3 (2022-2023): Embodied Vision-Language-Action (VLA) policies; d) Phase 4 (2023-2024): Memory, autonomous task composition, and Web-to-robot transfer; and e) Phase 5 (2024-present): Multi-sensory generalization and real-world deployment;
    \item Holistic, thorough, systematic, highly-granular, multi-criteria, \textbf{taxonomic investigation of robotic FM approaches}, taking into account the following main criteria: a) The type of the employed FM with respect to the input-output modalities involved; b) The nature of the underlying Neural Network (NN) architecture; c) The learning paradigm adopted for developing a robotic FM; d) The learning stage at which knowledge is incorporated to a FM; e) The task controlled by a robotic FM; and f) The application domain where a robotic FM is used. For each defined criterion, a methodical comparative analysis of the various categories of approaches and critical insights are provided;
    \item Synthesis of the \textbf{public datasets/benchmarks} used for training/evaluation purposes, organized around the main recurring dataset families, along with their typical uses and current gaps;
    \item Extensive discussion of \textbf{current challenges} and \textbf{future research directions} in the field.
\end{itemize}

Regarding existing surveys in the field, Table \ref{tab:fm_surveys_rev} comparatively analyzes the current work with the relevant literature reviews of \citep{hu2023toward,xu2024survey,ma2024survey,jang2024unlocking,kawaharazuka2024real,firoozi2025foundation,xiao2025robot,tayyab2025foundation,kawaharazuka2025vision,sapkota2025vision}, summarizing their main strengths and limitations. Examining Table \ref{tab:fm_surveys_rev}, it can be seen that literature works exhibit the following limitations: a) They often remain relatively specific/narrow in scope (i.e., adopting a task-, model-, application-, learning-, or integration-oriented perspective), leading to a non-thorough investigation of the overall research landscape, b) They consider a single or very few literature analysis criteria, resulting into a non-comprehensive examination of the literature works, c) They commonly adopt a narrative-style discussion of the literature, avoiding to provide a systematic comparison of the different approaches, and d) with the exception of the meta-analysis of \citet{hu2023toward}, they do not include any empirical/experimental benchmarking of the surveyed methods. On the contrary, the current survey provides a thorough and in-depth analysis of the research landscape of the use of FMs in robotics, demonstrating the following key advantageous/distinctive characteristics: a) It targets a holistic investigation of the overall field, b) It adopts a structured and systematic literature review methodology, supporting the search in $6$ major databases, the use of specific inclusion/exclusion criteria, and the application of an iterative screening process, c) It documents the main $5$ distinct research evolution phases, as well as the key trends associated with each of them, d) It supports a highly-granular, multi-criteria ($6$), taxonomic investigation of the literature, examining the different FM types, NN architectures, learning paradigms, learning stages, robotic tasks, and application domains, e) It incorporates a per criterion methodical comparative analysis of the different approaches and facilitates the reporting of critical insights, and f) It provides a comprehensive and hierarchical discussion on the current challenges and future research directions in the field. At the same time though, the current survey does not include any empirical/quantitative benchmarking of the surveyed methods and it inevitably constitutes a snapshot of a rapidly evolving field. Moreover, a more detailed version of the above comparison is provided in Appendix~\ref{sec:comparative_surveys_details}, which reports, for each surveyed work, its survey scope, review methodology, primary contributions, and main limitations.

The remainder of the manuscript is organized as follows: Section \ref{sec:literature} outlines the adopted methodology for reviewing the relevant literature. Section \ref{sec:Evolution} describes the evolution in robotic FM research and indicates the most widely adopted models. Section \ref{sec:Taxonomy} delineates the criteria used for analyzing the literature, as well as the resulting categories of robotic FM methods structured in the form of a taxonomy. Sections \ref{sec:fm_types}-\ref{sec:app_domains} discuss in detail the various categories of robotic FM approaches, taking into account the type of the employed FM, the nature of the underlying NN architecture, the adopted learning paradigm, the learning stage at which knowledge is integrated to the FM, the performed robotic task, and the selected application domain, respectively. Section \ref{sec:Datasets} reviews the publicly available datasets for training and evaluating robotic FM methods, organized into the main recurring dataset families, along with their typical uses and current gaps. Sections \ref{sec:challenges} and \ref{sec:future_directions} discuss the current challenges and future research directions in the field, correspondingly, while Section \ref{sec:Conclusion} concludes the paper.

\section{Literature review methodology}
\label{sec:literature}

\textbf{Overview}: In order to efficiently and thoroughly identify/map the robotic FM literature, while at the same time detecting key concepts and trends, a structured and systematic review methodology was adopted, ensuring comprehensiveness and relevance of the selected research works. The main aspects of this methodology are summarized below.

\textbf{Scope and objectives}: In terms of scope and objectives, the survey targets approaches for various robotic tasks (namely, perception, planning, navigation, manipulation, and human-robot interaction) whose execution relies on the use of an underlying FM, emphasizing the following objectives: a) The main categories of methods based on multiple criteria, b) The utilized datasets, c) Current challenges, and d) Future research directions.

\textbf{Literature search}: The literature search involved querying several major scientific databases, namely IEEE Xplore, Google Scholar, Scopus, DBLP, arXiv, and Web of Science, by combining targeted keywords/terms (e.g., ``foundation model'', ``robotics'', ``vision-language-action'', ``large language model'', etc.) with Boolean operators. To guarantee contemporary relevance, the search primarily focused on research works published within the last five years, while certain earlier seminal studies were also included. It needs to be clarified that the actual search process was applied iteratively with successive keyword refinement, and was further supplemented by an extensive backward reference-checking/chaining procedure. This multi-pronged design renders the overall literature review process relatively robust against missing critical/important works in the field, including ones whose original terminology predates the current FM established one. Moreover, this keyword plus reference search strategy was deliberately favored over a pure embedding-based semantic retrieval over paper embeddings, since it offers improved reproducibility, transparency, and precision/recall control; embedding-based semantic search can, nevertheless, serve as a useful complementary tool.

\textbf{Screening and selection}: A multi-stage screening process was then applied, excluding duplicate records, non-English papers, and articles without full-text access, followed by title/abstract filtering and in-depth full-text review. The latter retained only studies where a FM comprises a key algorithmic component and which exhibit substantial theoretical and/or experimental contributions, with priority given to prominent robotics and AI/ML publication venues. Eventually, a total of $438$ articles were selected for analysis and were included as references in the current manuscript.

\textbf{LLM assistance}: Regarding the use of automated tools during the preparation process of the manuscript, LLM assistance was employed only on specific occasions and solely as a complementary aid for cross-checking potential gaps in the literature search described above (i.e., as a supplementary means of identifying relevant works that might have been missed by the database queries). However, all substantive scholarly work was carried out manually by the authors; in particular, the taxonomy design, the paper inclusion/exclusion decisions, all comparative methodical analyses, the study of individual works, and the inclusion and verification of all references were carefully performed and validated by the authors. The authors take full responsibility for the entire content of the manuscript.

\textbf{Additional details}: The full details of the review methodology, including the exact database queries and key bibliometric analytics (article types and most popular publication venues) of the selected literature, are provided in Appendix~\ref{sec:sup_methodology-details}, whereas in-depth analysis of the identified robotic FM works is provided in Sections~\ref{sec:Evolution}-\ref{sec:Datasets}.

\section{Robotic FM research evolution}
\label{sec:Evolution}

\subsection{Research phases}
\label{ssec:FMphases}

\begin{figure*}[!t]
    \centering
    \includegraphics[height=0.9\textheight]{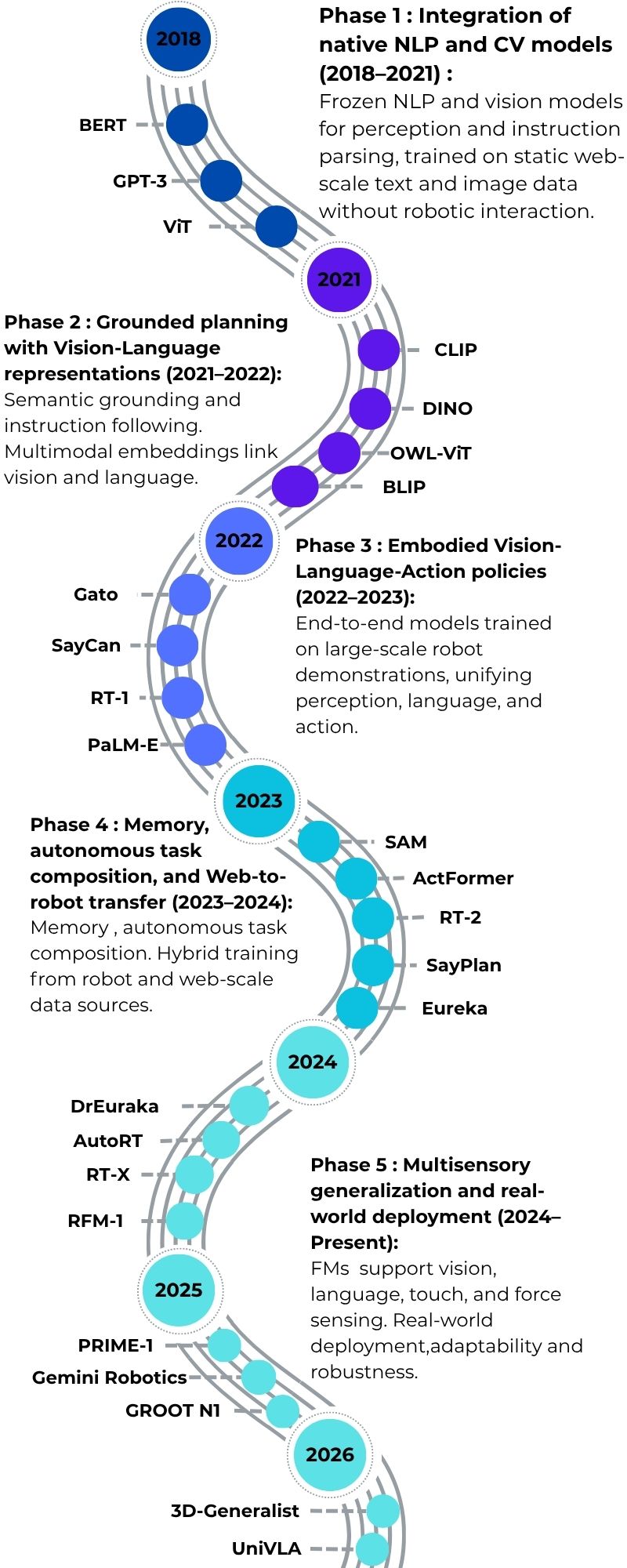}
    \caption{Main phases in robotic FM research and key/milestone works.}
    \label{f:Timeline}
\end{figure*}

Although foundation models have only relatively recently been introduced in the field of robotics, they have decisively contributed towards transformative effects, while gradual advancements and emerging research trends can be identified in the literature. In particular, the evolution of robotic FM research can be roughly classified into subsequent and distinct phases, each corresponding to a critical paradigm shift regarding how perception, reasoning, and control procedures are consolidated in a robotic system. The overall research progress concentrates on repositioning from isolated/modular designs to integrated/general-purpose agents \citep{gato,palme}. The considered research phases are graphically illustrated in Fig. \ref{f:Timeline}, along with key/milestone works associated with each of them, while they are explained in detail in the followings.

The delineation of the above phases follows a systematic rationale, where boundaries are combinatorially defined on the basis of the following parallel axes: a) The dominant source of training data (successively comprising Web text/image corpora, aligned image-text pairs, large-scale real robot trajectories, mixed Web and robot experiences, and multi-sensory simulation and real-world data), b) The level of system integration (ranging from modular pipelines relying on hand-engineered controllers to end-to-end, general-purpose policies), and c) The model output space (progressively encompassing perception/language grounding, task plans, and executable action sequences). Additionally, a given work is designated as a key/milestone one, in case that it first demonstrates a capability characteristic of a given phase, it is widely adopted as a reference point in the community, and it marks the transition to a subsequent phase. Moreover, it needs to be mentioned that the reported year ranges are approximate and partially overlapping, reflecting the gradual shifts in research and are anchored to the appearance of representative works (rather than to strict cut-offs).

Regarding alignment with the existing literature, it should be noted that current surveys (Table \ref{tab:fm_surveys_rev}) predominantly organize the field around capability- or model-centric taxonomies, rather than explicitly dated timelines. For the cases that prior works do address temporal progression, most notably the VLA-oriented reviews of \citet{kawaharazuka2025vision,ma2024survey,sapkota2025vision}, the described trajectory (from early vision-language grounding, to end-to-end VLA policies, and then to real-world generalist systems) is consistent with Phases $2$-$5$ of the current work. However, the explicit five-phase research decomposition introduced in the current study renders this implicit chronology concrete and extends it to more recent (2024-2026) developments in multi-sensory generalization and real-world deployment, which largely fall outside the coverage period of earlier surveys.

\subsubsection{Phase 1: Integration of native Natural Language Processing (NLP) and Computer Vision (CV) models (2018-2021)}
\label{sssec:FMphase1}

In the early attempts of incorporating FMs in robotic systems, native large-scale off-the-shelf NLP and CV models are used, aiming at enhancing robotic platforms with improved perception capabilities (for example, identifying and tracking objects in camera video sequences, translating human verbal requests to robot symbolic goals, etc.) \citep{hong2021vln}. The data required for training such FMs largely originate from standard Web text and image repositories (information sources that contain no robot action policies, no time-varying sensor streams, and no interactions with the physical world). These networks are integrated following a modular approach, requiring a separate conventional controller for realizing motion and action planning. Due to this fact, NLP and CV FMs serve a supportive role (boosting perception and language grounding tasks), while the robot’s behavior remains determined by a hand-engineered downstream pipeline \citep{shridhar2022cliport,hong2021vln}.

\subsubsection{Phase 2: Grounded planning with Vision-Language (VL) representations (2021-2022)}
\label{sssec:FMphase2}
With the emergence of vision-language and scalable language models, increased capabilities for integrating richer semantic priors and more flexible instruction grounding to robots are introduced. In particular, multimodal embeddings are used by robotic systems for linking visual input with natural language commands and generating (or evaluating) task sequences \citep{sun2022plate,saycan}. Such developments take advantage of broader access to aligned image-text data and pretrained language models \citep{clip}, although robotic data remains relatively limited. The latter constrains robots to utilize a relatively decreased number of own recorded/captured experiences; hence, most training samples originate from third-party datasets (e.g., curated demonstration benchmarks or Web-scraped image-text pairs), forcing models to generalize from proxy sources, rather than direct embodied trials \citep{shah2023lm}.

\subsubsection{Phase 3: Embodied Vision-Language-Action (VLA) policies (2022-2023)}
\label{sssec:FMphase3}
This phase marks the introduction of comprehensive and unified robot policies, derived directly from training using large-scale robot‐demonstration datasets. In particular, FMs process vision, language, and task context simultaneously, while outputting action sequences using the same architecture \citep{rt1,gato,palme}. This shift towards end-to-end training is enabled by the availability of real-world robotic data at scale \citep{openx}, collected considering hundreds of tasks and corresponding variations (e.g., different object types, lighting conditions, camera viewpoints, robot embodiments, etc.). As a consequence, robotic platforms are enhanced by incorporating the capability to generalize across goals and environments without explicit task engineering \citep{shridhar2023perceiver,rt2}.

\subsubsection{Phase 4: Memory, autonomous task composition, and Web-to-robot transfer (2023-2024)}
\label{sssec:FMphase4}
Building on prior advances, this phase concerns systems capable of long-horizon planning \citep{ajay2023compositional}, world-state tracking \citep{wu2023daydreamer}, and autonomous skill discovery \citep{nam2023lift}. In this respect, FMs are trained or adapted using both structured robot trajectories and unstructured Web-scale corpora \citep{openvla}. Additionally, data diversity increases, combining multimodal internet data with robot-collected experiences \citep{team2024octo}. Moreover, robotic agents synthesize and evaluate their own tasks based on FM reasoning pipelines, boosting semantic autonomy and self-improvement \citep{parakh2024lifelong}.

\subsubsection{Phase 5: Multi-sensory generalization and real-world deployment (2024-Present)}
\label{sssec:FMphase5}
More recently, research advancements have focused on building robust generalist robotic systems, capable of efficiently operating in unstructured real-world environments \citep{bjorck2025gr00t,team2025gemini}. The utilized FM-based solutions support multimodal inputs, including vision, language, touch, force, and proprioception \citep{li2026simultaneous}, as well as real-time adaptation \citep{routray2025vipra}. Additionally, robotic agents are trained and refined using a combination of simulation and real-world interaction data, significantly extending the boundaries in robustness, transferability, safety, and autonomy \citep{zhao2026sim2realvla,guo2026ctrlworld}. Consequently, FM-based solutions are being widely adopted across industrial, assistive, and mobile platforms \citep{covariant2024rfm1}.

\begin{table}[t]
  \centering
  \caption{Most common and widely adopted FMs in robotics.}
  \label{tab:main_fms}
  \scriptsize
  
  \setlength{\aboverulesep}{0pt}
  \setlength{\belowrulesep}{0pt}
  \setlength{\tabcolsep}{2pt}
  \renewcommand{\arraystretch}{0.9}

  \newlength{\Wmodel}\setlength{\Wmodel}{2.05cm}
  \newlength{\Wyear}\setlength{\Wyear}{0.62cm}
  \newlength{\Wcat}\setlength{\Wcat}{1.32cm}
  \newlength{\Wtype}\setlength{\Wtype}{0.72cm}
  \newlength{\Winput}\setlength{\Winput}{1.35cm}
  \newlength{\Woutput}\setlength{\Woutput}{1.35cm}
  \newlength{\Wemb}\setlength{\Wemb}{0.80cm}
  \newlength{\Wparam}\setlength{\Wparam}{1.06cm}
  \newlength{\Wpublic}\setlength{\Wpublic}{1.00cm}
  \newlength{\Wkey}\setlength{\Wkey}{4.99cm}

  \rowcolors{2}{gray!25}{white}

  \resizebox{\textwidth}{!}{%
  \begin{tabular}{@{}|
    >{\justifying\arraybackslash}m{\Wmodel}|
    >{\centering\arraybackslash}m{\Wyear}|
    >{\centering\arraybackslash}m{\Wcat}|
    >{\centering\arraybackslash}m{\Wtype}|
    >{\centering\arraybackslash}m{\Winput}|
    >{\centering\arraybackslash}m{\Woutput}|
    >{\centering\arraybackslash}m{\Wemb}|
    >{\centering\arraybackslash}m{\Wparam}|
    >{\centering\arraybackslash}m{\Wpublic}|
    >{\justifying\arraybackslash}m{\Wkey}|
  @{}}
    \toprule
    \rowcolor{gray!40}
    \headerbreak{Model} &
    \headerbreak{Year} &
    \headerbreak{Category} &
    \headerbreak{Type} &
    \headerbreak{Input} &
    \headerbreak{Output} &
    \headerbreak{Emb.} &
    \headerbreak{Param.} &
    \headerbreak{Public} &
    \headerbreak{Key innovation} \\
    \midrule

    BERT \citep{devlin2019bert} & 2019 & FM & LLM & Text & Text embedding & N & 110M & Y & Natural-language command interpretation and translation to action sequences \\ \midrule

    GPT-3 \citep{brown2020gpt3} & 2020 & FM & LLM & Text & Text & N & 175B & N & Zero/few-shot reasoning for instruction-to-plan/code translation \\ \midrule

    ViT \citep{vit} & 2021 & FM & VFM & Image & Class logit & N & 86M & Y & Visual perception capturing long-range dependencies and global context \\ \midrule

    CLIP \citep{clip} & 2021 & FM & VLM & Text, image & Embedding & N & 428M & Y & Zero-shot object grounding and task specification from free-form text \\ \midrule

    DINO \citep{caron2021emerging} & 2021 & FM & VFM & Image & Embedding & N & 86M & Y & Self-supervised object attention maps aiding manipulation \\ \midrule

    OWL-ViT \citep{minderer2022simple} & 2022 & FM & VLM & Text, image & Box, label & N & 100M & Y & Zero-shot open-vocabulary object detection from text \\ \midrule

    BLIP \citep{blip} & 2022 & FM & VLM & Text, image & Text, embedding & N & 480M & Y & Zero-shot vision-language understanding and generation \\ \midrule

    Gato \citep{gato} & 2022 & FM & VLA & Text, image, state & Action & Y & 1.18B & N & Generalist multi-task, multi-embodiment policy (600+ tasks) \\ \midrule

    SayCan \citep{saycan} & 2023 & System & LLM & Text, environment & Action & Y & -- & N & Affordance-grounded, feasible robotic action planning \\ \midrule

    RT-1 \citep{rt1} & 2023 & FM & VLA & Text, image & Action & Y & -- & Y & End-to-end mobile-manipulation policy from 130K+ real trajectories \\ \midrule

    PaLM-E \citep{palme} & 2023 & FM & VLA & Text, image & Text, action & Y & 562B & N & Embodied multimodal reasoning and long-horizon plan generation \\ \midrule

    SAM \citep{sam} & 2023 & FM & VFM & Image, prompt & Mask & N & 636M & Y & Promptable zero-shot image segmentation \\ \midrule

    ActFormer \citep{xu2023actformer} & 2023 & FM & VFM & Action class & 3D motion & N & -- & N & Action-conditioned 3D human motion generation (GAN transformer) \\ \midrule

    RT-2 \citep{rt2} & 2023 & FM & VLA & Text, image & Action & Y & -- & N & Web-scale and robot co-training for generalization to unseen tasks/objects \\ \midrule

    SayPlan \citep{rana2023sayplan} & 2023 & System & LLM & Text, scene graph & Plan & Y & -- & N & Scalable long-horizon planning over 3D scene graphs \\ \midrule

    Eureka \citep{ma2023eureka} & 2024 & System & LLM & Task description & Reward & N & -- & Y & Iterative LLM generation of robot reward-function code \\ \midrule

    DrEureka \citep{ma2024dreureka} & 2024 & System & LLM & Simulation, task configuration & Reward & N & -- & Y & Prompt-driven automation of the sim-to-real training pipeline \\ \midrule

    AutoRT \citep{ahn2024autort} & 2024 & System & VLA & Text, image & Action & Y & -- & N & Safety-constrained orchestration of robot action generation \\ \midrule

    RT-X \citep{openx} & 2024 & FM & VLA & Text, image & Action & Y & -- & Y & Cross-embodiment policy across 22 platforms (527 skills) \\ \midrule

    RFM-1 \citep{covariant2024rfm1} & 2024 & FM & VLA & Text, image, video & Action & Y & 8B & N & Multimodal physics-informed reasoning for complex tasks \\ \midrule

    PRIME-1 \citep{ambirobotics2025prime1} & 2025 & FM & VLA & Image & 3D features, action & Y & -- & N & Real-world adaptive control for multi-task operational settings \\ \midrule

    Gemini robotics \citep{team2025gemini} & 2025 & FM & VLA & Text, image & Action & Y & -- & N & On-device multimodal reasoning for dexterous bi-manual tasks \\ \midrule

    GR00T N1 \citep{bjorck2025gr00t} & 2025 & FM & VLA & Text, image & Action & Y & 2B & Y & Single, multi-task, general-purpose architecture for humanoid robots \\ \midrule

    3D-GENERALIST \citep{sun20253d} & 2026 & FM & VLM & Text, image & Action code, 3D scene & N & -- & N & VLM-as-policy generation of simulation-ready 3D environments \\ \midrule

    UniVLA \citep{wang2025unified} & 2026 & FM & VLA & Text, image, video & Action, image & Y & 8.5B & Y & Unified token-space vision-language-action modeling with video world-model post-training \\

    \bottomrule
  \end{tabular}%
  }
\end{table}

\subsection{Key foundation models}
\label{ssec:FMmodels}
Throughout the different research phases described in Section \ref{ssec:FMphases}, key FM architectures have been introduced, which, on the one hand, have led to significant technological advancements and capabilities, and, on the other hand, have served as the basis for numerous methods in the field. In particular, the most common and widely adopted robotic FMs, along with their main characteristics, are briefly summarized in Table~\ref{tab:main_fms}. For each model, the latter reports its year of publication, category, type, input and output modalities, adoption of embodied design (`Emb.'), number of parameters (`Param.'), public availability of an implementation (`Public'), and key innovation. Regarding the category criterion, a distinction is made between standalone robotic foundation models (denoted as `FM') and FM-based robotic systems (denoted as `System'); the latter comprise modular/compositional approaches that orchestrate one or more existing FMs, rather than constituting standalone models themselves. Regarding the embodiment criterion, a model is considered embodied (`Emb.'~$=$~Y) only when it is trained on, or directly produces, robot-executable actions/observations of a physical or simulated agent; hence, general-purpose models that are widely adopted in robotics but operate on generic vision/language data are not characterized as embodied themselves.

\section{Key criteria and main categories of robotic FM methods}
\label{sec:Taxonomy}

This section provides a systematic overview of the landscape of robotic FM methods. For facilitating the analysis, a set of complementary and diverse criteria are defined (each focusing on a specific/key aspect of a robotic FM system), resulting into the classification of the literature works into a corresponding set of main categories. The different criteria used and the resulting categories are graphically illustrated in Fig.~\ref{f:FMTaxonomy} and detailed as follows:

\begin{itemize}
    \item \textbf{Foundation model type}: FMs in robotics can be grouped taking into account the number and the nature of the input-output modalities involved, which critically dictates their overall capabilities and how they manage robot perception, reasoning, and action procedures. The main types of FMs are:
    \begin{itemize}
        \item \textbf{Large Language Models (LLMs)}: These receive textual streams of data as input (sometimes also multimodal information) and, subsequently, generate high-level action policies. Their fundamental functionality relies on translating natural language instructions into task goals, programs, or supervision signals \citep{saycan,liang2023code,palme}.
        \item \textbf{Vision Foundation Models (VFMs)}: These receive as input visual information streams (e.g., RGB, LiDAR, thermal, etc.) and output corresponding delicate representations (e.g., object detection masks, depth maps, dense embeddings, etc.) for facilitating subsequent robotic tasks \citep{sam,zhang2022dino}.  
        \item \textbf{Vision-Language Models (VLMs)}: These exploit correlations and inter-dependencies among the input visual and textual data, in order to enable complex robot operations (e.g., visual grounding, language-conditioned mapping, question answering, etc.) \citep{clip,liu2024grounding}.  
        \item \textbf{Vision-Language-Action models (VLAs)}: These comprise native robotic models that map multi-modal inputs directly/end-to-end to generalist action policies, across multiple types of tasks and embodiments \citep{rt2,openvla,bjorck2025gr00t}.  
    \end{itemize}

    \item \textbf{Neural Network (NN) architecture}: The type of the underlying neural network architecture that is employed in any FM solution largely defines its capabilities, efficiency, and limitations in robotic applications. The main NN types used in robotic FM methods are:
    \begin{itemize}
        \item \textbf{Transformers}: These rely on self-attention to capture long-range dependencies and to model complex multimodal inputs in a unified way \citep{vaswani2017attention,palme,rt2}.
        \item \textbf{State-Space Models (SSMs)}: Their fundamental functionality is grounded on the use of a set of first-order differential or difference equations for modeling complex, dynamic operations, typically involving  multiple input and output signals \citep{gu2023mamba,liu2024robomamba,lieber2024jamba}.
        \item \textbf{Diffusion models}: These employ an iterative denoising process, in order to generate diverse and realistic data samples (e.g., action policies) \citep{ho2020denoising,chi2025diffusion}.
        \item \textbf{Convolutional and hybrid encoders}: Convolutional Neural Network (CNN) encoders are efficient in learning hierarchical feature representations and modeling local patterns in the input data, while they are often combined with transformer or diffusion networks for further enhancing visual perception \citep{nair2022r3m,rt1,chi2025diffusion}.
        \item \textbf{Graphical models}: These allow the processing of data that exhibit irregular and/or complex relations, while also enabling the generation of context-aware representations \citep{rana2023sayplan,gu2024conceptgraphs,patel2025get}.
    \end{itemize}

    \begin{figure*}[t]
    \centering
    \includegraphics[width=1.0\textwidth]{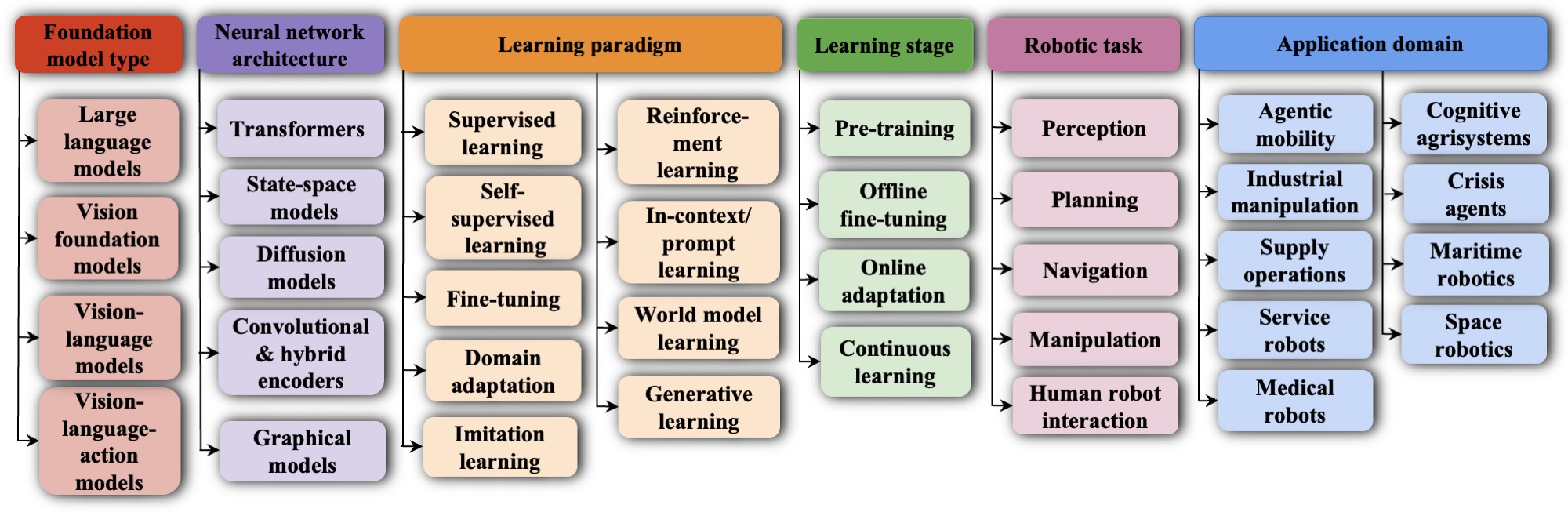}
    \caption{Key criteria and main resulting categories of robotic FM methods.}
    \label{f:FMTaxonomy}
    \end{figure*}

    \item \textbf{Learning paradigm}: In order to develop a robust robotic FM solution, different/diverse learning techniques, principles, and approaches can be adopted. The most commonly met learning paradigms in the literature, which are typically combined in a comprehensive learning methodology, are the following:
    \begin{itemize}
        \item \textbf{Supervised Learning (SL)}: This relies on training a FM on large-scale, labeled datasets of input-output pairs (e.g., observations/instructions mapped to target outputs or actions), so that the model directly learns the desired input-to-output mapping from explicit, human-provided supervisory signals \citep{rt2, palme, team2025gemini}.
        \item \textbf{Self-Supervised Learning (SSL)}: This enables ML models to generate their own supervisory signals directly from raw, unlabeled data, thereby circumventing the need for costly external human-provided labels \citep{gao2025adaworld,nazeri2024vertiencoder}.
        \item \textbf{Fine-tuning}: This allows generalist FMs to efficiently adapt to specific tasks, environments, and robotic platforms, while demonstrating significant efficiency and cost-effectiveness \citep{team2024octo, yadav2026robustfinetuning}.
        \item \textbf{Domain Adaptation (DA)}: This aims at adjusting a FM, originally trained on a source domain, to maintain its accuracy and performance when applied to a new target domain \citep{li2026metavla,zheng2026xvla}. 
        \item \textbf{Imitation Learning (IL)}: Also known as Learning from Demonstration (LfD) or Robot programming by Demonstration (PbD), it relies on the consideration of an autonomous agent learning to execute tasks or acquiring new skills, by observing/emulating demonstrations provided by an expert \citep{wan2024lotus, fu2025icrt, cai2023groot}.
        \item \textbf{Reinforcement Learning (RL)}: This is grounded on the use of an agent learning to make sequential decisions and to adjust its behavior through trial-and-error interactions with its surrounding environment, taking into account feedback received in the form of rewards or penalties for its actions \citep{ma2023eureka,tziafas2024lifelong,wang2024rl}.

        \item \textbf{In-context/prompt learning}: This paradigm enables inference-time adaptation of FMs, by guiding model behavior through demonstrations, examples, or task-specific instructions. As such, it supports flexible task adaptation and behavior specification, allowing pre-trained models to generalize to new scenarios through contextual conditioning alone \citep{huang2023instruct2act,grigorev2025verifyllm,yin2024context}.
        \item \textbf{World Model (WM) learning}: WMs enable the grounding of FM knowledge into physically, real-world, plausible predictions, by modeling environmental dynamics and predicting the consequences of robot actions \citep{gao2025adaworld,zhou2025dino}.
        \item \textbf{Generative Learning (GL)}: This facilitates towards reducing the reliance on extensive real-world datasets during training, by artificially synthesizing diverse and high-quality robot experiences \citep{zhao2026sim2realvla,heppert2026real2gen}.
    \end{itemize}
    
    \item \textbf{Learning stage}: The particular phase, during the overall learning process, at which knowledge is incorporated to a FM, largely defines the type/nature of the acquired knowledge, algorithmic/development details, and key assumptions about the model behavior. In this respect, the main learning stages identified in the literature are summarized as follows:    
    \begin{itemize}
        \item \textbf{Pre-training}: This is the first and by-far the most computationally intensive step in the FM development life-cycle, which involves the processing of massive, internet-scale, diverse, and usually multi-modal datasets for learning general-purpose feature representations \citep{rt2,palme}.    
        \item \textbf{Offline fine-tuning}: Following generic pre-training, this step focuses on adjusting the FM knowledge structures to the requirements/nuances of particular application domains or downstream tasks, making use of (minimal) additional training data \citep{team2024octo,li2026metavla}. 
        \item \textbf{Online adaptation}: This corresponds to real-time adjustments of a robot's behavior for maintaining performance during deployment, involving the acquisition of new skills, response to novel tasks, or handling of unforeseen environmental conditions \citep{wang2024rl,grigorev2025verifyllm}.
        \item \textbf{Continuous learning}: This aims at enabling FMs to continuously acquire new skills, to refine existing ones, and to maintain performance in dynamic, real-world environments in the long term \citep{wan2024lotus,murillo2025actionflowmatching}. 
    \end{itemize}
    
    \item \textbf{Robotic task}: The introduction of FMs has induced transformative effects in the materialization and execution of all core robotic tasks, i.e., specific jobs, actions, or functions that robots perform in order to achieve a goal. The most common, pronounced robotic tasks, where FMs are applied to, are as follows:
    \begin{itemize}
        \item \textbf{Perception}: FMs equip robots with enhanced capabilities to perceive and reason about their surrounding environment, largely relying on visual information processing streams and often combined with additional modalities (e.g., natural language inputs) \citep{clip,10655587,yamazaki2024open,nguyen2024open3dis}.
        \item \textbf{Planning}: FMs enable robots to interpret complex, high-level, human-like commands (e.g., in natural language form) and to subsequently translate them into (long-horizon) sequences of low-level, executable, and discrete actions \citep{saycan,liang2023code,chen2024autotamp,singh2023progprompt}.    
        \item \textbf{Navigation}: FMs significantly boost robot navigation capabilities, by moving away from traditional, task-specific models and heading towards more generalized, adaptable schemes for efficient operation in complex, unstructured, and dynamic environments \citep{shah2023vint,wang2024navformer,huang2023visual,xu2024drivegpt4}.
        \item \textbf{Manipulation}: Often equally termed motor control, it is enhanced by the use of FMs by shifting away from task-specific programming to more generalized, adaptable approaches, also supporting more dexterous and precise manipulation tasks \citep{rt1,rt2,palme,bjorck2025gr00t}.
        \item \textbf{Human-Robot Interaction (HRI)}: FMs enable robotic platforms to understand and interact/respond with/to humans in a more intuitive, natural, flexible, and human-like way \citep{izquierdo2024plancollabnl,liu2024dragon,barmann2024incremental,irfan2024recommendations}.
    \end{itemize}

    \item \textbf{Application domain}: FMs have significantly enhanced several aspects of robot capabilities (e.g., autonomy, complex decision-making, human-robot interaction, etc.) in challenging real-world settings; hence, further boosting their widespread use, including the following main/common application domains:
    \begin{itemize}
        \item \textbf{Agentic mobility}: FMs significantly extend the capabilities of the conventional autonomous driving stack (i.e., perception, prediction, planning, and control), by transforming it into a single, cohesive, end-to-end decision-making framework \citep{wu2024vision,xu2024drivegpt4,wang2024drive}.
        \item \textbf{Industrial manipulation}: FMs revolutionize industrial automation pipelines, by converting rigid, task-specific solutions into flexible, general-purpose agents that are capable of handling dynamic, unstructured manufacturing tasks \citep{covariant2024rfm1,ambirobotics2025prime1,openvla,rt2}.
        \item \textbf{Supply operations}: FMs dramatically increase flexibility, intelligence, and generalization, enabling robots to move beyond repetitive, structured tasks to handle unstructured, dynamic, and complex operations \citep{nicoletti2024green,xu2024multi,nicoletti2025foundation}.
        \item \textbf{Service robots}: Robots are more efficient in operating safely and intelligently in challenging household environments, while greatly capitalizing on their ability of receiving instructions in natural language \citep{wu2023tidybot,mon2025embodied}.
        \item \textbf{Medical robots}: Robotic platforms incorporate comprehensive, fine-grained medical knowledge, enabling them to robustly undertake high-stake, high-variability tasks, to provide context-aware intelligence, and to support consistent, precise assistance \citep{cui2024surgical,zeinoddin2024dares,he2024foundation}.
        \item \textbf{Cognitive agrisystems}: Robotic platform capabilities evolve from conventional, field-level task execution to efficient, resilient, and sustainable precision farming, i.e., moving beyond simple automation to genuine, autonomous intelligence \citep{yin2025foundation}.
        
        \item \textbf{Crisis agents}: FMs enable robot operations to elaborate from conventional, remote-controlled settings to the handling of autonomous reasoning and adaptation circumstances in unpredictable, highly dangerous environments \citep{palme,openvla,rt2}.

        \item \textbf{Maritime robotics}: Robots are reinforced with advanced capabilities so as to efficiently overcome typical, extreme environmental challenges in under- and open-water settings, in principle relying on robust, generalized, and multi-sensorial intelligence/reasoning pipelines \citep{zheng2024marineinst}.
        
        \item \textbf{Space robotics}: FMs enable the functioning of robots under extreme operating conditions, involving limited resource availability and highly variable, unknown environments, largely relying on their enhanced autonomous decision-making capabilities \citep{giannakis2023deep,zhao2024crater}.

    \end{itemize}
\end{itemize}

\section{Foundation model types}
\label{sec:fm_types}

FMs used in robotics can be organized into groups with respect to the number and the nature of the input-output modalities involved. The latter also largely affects their exhibited capabilities. In particular, the main types of FMs are: a) Large Language Models (LLMs), b) Vision Foundation Models (VFMs), c) Vision-Language Models (VLMs), and d) Vision-Language-Action models (VLAs), as discussed in Section \ref{sec:Taxonomy} and further detailed below.

\subsection{Large Language Models (LLMs)}
\label{ssec:llm_in_robotics}

In the context of robotics, LLMs are primarily used as high-level, cognitive task planners and reasoning engines that generate the sequence of operations that are necessary for accomplishing a stated goal \citep{saycan,liang2023code,chen2024autotamp}. In practice, they translate high-level natural language inputs to low-level robot behaviors, turning free-form instructions and short state summaries into typed goals, multi-step plans, executable code, constraints, and run-time feedback, which renders them particularly useful for tasks requiring sophisticated reasoning and complex decision-making \citep{li2025large}.

In terms of supported functionality/operation, LLM-based systems can be classified into the following main categories:
\begin{itemize}
    \item \underline{Goal/constraint grounding and context}: LLMs map abstract human instructions (goal and context) to low-level robot actions (grounding). SayCan \citep{saycan} filters skills via PaLM-based value affordances to stay within the robot's capabilities, while LM-Nav \citep{shah2023lm} converts instructions into visually grounded way-points for a navigation planner.
    \item \underline{Command interpretation and code synthesis}: LLMs act as natural language to code translators, letting robots accept high-level instructions instead of hand-written code. Code-as-Policies \citep{liang2023code} compiles instructions into inspectable, reusable robot API code, while AutoTAMP \citep{chen2024autotamp} translates requests into TAMP specs checked by a symbolic planner for feasibility.
    \item \underline{Task planning and long-horizon reasoning}: LLMs decompose abstract goals into sequential, grounded actions with contextual awareness across steps. SELP \citep{wu2025selp} maps instructions to temporal logic via constrained decoding so that plans meet safety and efficiency constraints, while LLM-GROP \citep{zhang2025llm} cross-checks instructions against motion feasibility in cluttered settings.
    \item \underline{Perception-aware and multimodal integration}: LLM reasoning is enhanced by accounting for the robot's visual and physical surroundings. PaLM-E \citep{palme} injects visual and proprioceptive streams into inference so that decisions match real-world observations, while Chain-of-Modality \citep{wang2025chain} derives a plan and control parameters from human videos and auxiliary signals.
    \item \underline{Navigation and spatial understanding}: LLMs turn abstract navigation commands into grounded paths via structured textual scene representations. LM-Nav \citep{shah2023lm} links instructions to landmarks and routes through CLIP-based grounding, while SayPlan \citep{rana2023sayplan} plans over a $3$D scene graph and re-plans on infeasible steps for large-scale missions.
    \item \underline{Conversational interfaces and teleoperation}: LLMs create conversational interfaces and shared-control teleoperation, making robots accessible to non-experts. TidyBot \citep{wu2023tidybot} learns user-specific tidying conventions and transfers them to new homes, while LAMS \citep{tao2025lams} predicts intent and auto-switches teleoperation modes to lower cognitive load.
    \item \underline{Execution-time validation, error handling, and recovery}: LLMs recast run-time failures as semantic problems for dynamic recovery, instead of brittle pre-defined routines. STATLER \citep{yoneda2024statler} interprets robot state and tool feedback to suggest targeted repairs without restarting, while CAPE \citep{raman2024cape} re-prompts and proposes fixes for precondition failures.
    \item \underline{Adaptation, efficiency, and safety}: LLMs facilitate robots to adapt and to operate safely under novel situations and disruptions. Eureka \citep{ma2023eureka} writes and refines GPT-$4$ reward code to speed skill acquisition across platforms, while DrEureka \citep{ma2024dreureka} co-designs rewards and domain randomization for efficient sim-to-real transfer.
    \item \underline{Knowledge retrieval and memory}: LLMs access and store external knowledge and past episodes to overcome limitations of immediate sensor data. ELLMER \citep{mon2025embodied} couples GPT-$4$ with retrieval-augmented memory so that a mobile manipulator incorporates context, adapts plans on the fly, and completes multi-step household tasks as conditions change.
\end{itemize}

\subsection{Vision Foundation Models (VFMs)}
\label{ssec:vfm_in_robotics}

The ultimate goal of VFMs is to address the perceptual requirements of embodied AI systems, by providing generalized, high-quality visual representations necessary for interaction with the physical world \citep{sam,oquab2023dinov2}. In the robotics setting, VFMs distill raw pixel information into rich, transferable visual features or embeddings, enabling a robust and generalized visual understanding that serves as the input information stream for modulating downstream control policies \citep{shangtheia}.

In terms of supported functionality/operation, VFM-based
systems can be classified into the following main categories:
\begin{itemize}
    \item \underline{Object recognition}: VFMs enable generalized visual recognition via transferable representations that boost task-specific training. SAM \citep{sam} yields class-agnostic, promptable segmentation masks for isolating objects, while DINOv2 \citep{oquab2023dinov2} provides robust dense features that transfer across scenes under domain shift.

    \item \underline{Localization}: VFMs improve localization with robust, semantic, globally consistent representations beyond geometric methods. DINO-VO \citep{azhari2025dino} uses DINOv2 features and ViT-based keypoints for robust monocular visual odometry at high throughput, while LiteVLoc \citep{jiao2025litevloc} enables long-range re-localization for image-goal navigation.

    \item \underline{Object tracking}: VFMs support trackers rich semantic understanding and long-term memory, improving robustness to occlusion and viewpoint change. \citet{zhong2024empowering} extract text-prompted segmentation masks to train a recurrent policy via offline RL, while open-vocabulary cues enable instance tracking of novel objects over time \citep{guo2025openvis}.
    
    \item \underline{Depth perception}: VFMs enable robust, high-fidelity depth estimation, where sensors or traditional methods under-perform. Metric$3$D v$2$ \citep{hu2024metric3d} uses geometric priors for zero-shot metric depth estimation across diverse cameras, while Prompt-Depth-Anything \citep{lin2025prompting} demonstrates that a small LiDAR `metric prompt' can steer a FM to accurate, high-resolution depth estimation.
    
    \item \underline{Semantic map creation}: VFMs supply robust, semantic-aware features that improve map accuracy and informativeness. \citet{busch2025one} build reusable open-vocabulary feature maps with probabilistic-semantic updating, while combining VFM features with Gaussian-splatting supports robust long-horizon missions in dynamic environments \citep{zheng2025wildgs}.
    
    \item \underline{Visual-inertial fusion}: VFMs enhance the visual part of visual-inertial odometry (VIO) systems, which is crucial for drift correction and estimation of metric scale. Specifically, features that improve VO \citep{azhari2025dino} or metric priors from depth FMs \citep{hu2024metric3d} are combined with Inertial Measurement Unit (IMU) data in standard VIO estimators.
    \item \underline{Environment mapping}: VFMs construct maps by producing dense visual embeddings fused into persistent, semantically enriched scenes. FMGS \citep{zuo2025fmgs} combines FM features with $3$D Gaussian splatting for semantic reconstruction and open-vocabulary understanding, while OpenGS-SLAM \citep{yang2025opengs} adds FM-derived semantic features for robust real-time tracking and mapping.
\end{itemize}

\subsection{Vision-Language Models (VLMs)}
\label{ssec:vlms_in_robotics}

VLMs combine computer vision with natural language processing capabilities for establishing a coherent, concrete semantic understanding of the world, enabling interpretation and generation of language descriptions of the observed visual entities \citep{clip}. In the context of robotics, VLMs enable robots to simultaneously interpret visual data and natural language commands, allowing for intuitive human-robot interaction and robust task execution in unstructured environments \citep{zhou2025physvlm}.

With respect to supported functionalities/operations, VLM-based systems can be classified into the following main categories:
\begin{itemize}
    \item \underline{Manipulation grounding and control signals}: VLMs map high-level semantic intents into concrete, actionable constraints near the robot. OmniManip \citep{pan2025omnimanip} converts VLM reasoning into object-centric primitives with dual closed-loop planning and execution for precise $3$D constraints generation, while RoboGround \citep{huang2025roboground} feeds grounded target and placement masks into a low-level policy.
    
    \item \underline{Semantic mapping, referring expressions, and navigation}: VLMs build human-readable maps, interpret spatial language, and ground navigation goals. One-Map-to-Find-Them-All \citep{busch2025one} forms a reusable open-vocabulary map for zero-shot multi-object-based navigation, while VLFly \citep{zhang2025grounded} performs grounded vision-language UAV navigation without active ranging sensors.
    
    \item \underline{Execution-time check and progress verification}: VLMs enable closed-loop semantic self-monitoring, turning sensor feedback into human-understandable checks. ExploreVLM \citep{lou2025explorevlm} integrates perception, planning, and execution validation in real time, while \citet{ahmad2025unified} verify skill pre- and post-conditions and suggest recovery skills for failure handling.
    
    \item \underline{Closed-loop mobile manipulation}: VLMs supply continuous feedback and adaptation for long-horizon execution in unstructured, dynamic environments. COME-robot \citep{zhi2025closed} uses GPT-$4$ for situated reasoning and iterative feedback to recover from failures, while HomeRobot \citep{yenamandra2023homerobot} navigates homes to grasp novel objects and to place them on receptacles.
\end{itemize}

\subsection{Vision-Language-Action models (VLAs)}
\label{ssec:vla_in_robotics}
VLAs aim at integrating multi-modal understanding with direct physical execution, targeting to serve as the basis for autonomous embodied task execution \citep{ma2024survey}. In particular, a VLA model receives multi-modal inputs (typically, vision, language, and robot state) and generates real-world physical actions or control policies in real-time, often designed in an end-to-end fashion \citep{sapkota2025vision}. 

In terms of supported functionality/operation, VLA-based
systems can be classified into the following main categories:
\begin{itemize}
    \item \underline{Scaling and web-to-robot transfer}: VLAs transfer internet-scale semantic and visual knowledge to control policies, boosting cross-task and cross-embodiment generalization. RT-$1$ \citep{rt1} scales imitation learning with language-tied tokenized actions for long-tail tasks, while RT-$2$ \citep{rt2} adds web-scale vision-language pretraining for open-vocabulary transfer to real robots.
    \item \underline{Fusion and action parameterization}: VLAs unify perception, reasoning, and control by pairing a VLM with an action decoder. GR$00$T N$1$ \citep{bjorck2025gr00t} couples a VLM with a diffusion transformer for real-time motor actions, while $\pi_{0.5}$ \citep{intelligence2025pi_} casts action generation as flow matching for stable continuous control at moderate cost.
    
    \item \underline{Specialization, adapters, and mixture-of-experts}: VLAs exploit massive pre-trained backbones without running the full network per action. MoRE \citep{zhao2025more} activates a few sparse LoRA experts per step to expand capacity at no extra inference cost, while OpenVLA \citep{openvla} pairs a Llama $2$ model with a visual encoder and efficient fine-tuning for generalizable visuo-motor policies.
    \item \underline{Navigation and locomotion}: VLAs act as semantic navigation planners translating language into movements for mobile and legged robots, often hierarchically. NaVILA \citep{ChengA-RSS-25} generates mid-level actions as language that drive a visual locomotion RL policy, while VAMOS \citep{castro2025vamos} decouples a generalist planner from a specialist affordance model encoding physical constraints.
    \item \underline{Operations, deployment, and safety}: VLAs combine flexible, generalized reasoning with safety guarantees for real-world deployment. SafeVLA \citep{zhang2025safevla} frames safety alignment as constraint learning so that operation respects task rules beyond post-hoc filtering, while VLATest \citep{wang2025vlatest} generates robotic manipulation scenes for systematically testing VLAs.
\end{itemize}

\begin{table}[!t]
  \caption{Foundation model types: Comparative analysis and key insights.}
  \label{tab:fm_types_summary}
  \centering
  \scriptsize
  
  \setlength{\aboverulesep}{0pt}
  \setlength{\belowrulesep}{0pt}
  \setlength{\tabcolsep}{2pt}
  \renewcommand{\arraystretch}{0.9}

  \setlist*[tabitem]{before=\vspace{2.2pt}\justifying, after=\vspace{2.2pt}}

  \rowcolors{2}{gray!25}{white}

  \newlength{\Waspect}\setlength{\Waspect}{1.35cm}
  \newlength{\Wllm}\setlength{\Wllm}{4cm}
  \newlength{\Wvfm}\setlength{\Wvfm}{4cm}
  \newlength{\Wvlm}\setlength{\Wvlm}{4cm}
  \newlength{\Wvla}\setlength{\Wvla}{4cm}

  \resizebox{\textwidth}{!}{%
  \begin{tabular}{@{}|
    >{\justifying\arraybackslash}m{\Waspect}|
    >{\raggedright\arraybackslash}m{\Wllm}|
    >{\raggedright\arraybackslash}m{\Wvfm}|
    >{\raggedright\arraybackslash}m{\Wvlm}|
    >{\raggedright\arraybackslash}m{\Wvla}|
  @{}} 
    \toprule
    \rowcolor{gray!40}
    \headerbreak{Aspect} &
    \headerbreak{LLMs} &
    \headerbreak{VFMs} &
    \headerbreak{VLMs} &
    \headerbreak{VLAs} \\
    \midrule

    Primary functions &
    \begin{tabitem}
      \item Cognitive task planning
      \item Symbolic reasoning
      \item Language-to-action translation
      \item Task decomposition
    \end{tabitem} &
    \begin{tabitem}
      \item Visual representations
      \item Dense features/embeddings
      \item Object/instance differentiation
      \item Geometric reconstruction
    \end{tabitem} &
    \begin{tabitem}
      \item Visual language grounding
      \item Open-vocabulary recognition
      \item Visual reasoning
      \item Semantic alignment
    \end{tabitem} &
    \begin{tabitem}
      \item Policy execution
      \item Hardware actuation
      \item Embodied task execution
      \item Multi-modal physical alignment
    \end{tabitem} \\
    \midrule

    Input &
    \begin{tabitem}
      \item Text tokens
      \item NL instructions/goals
      \item Reasoning traces
      \item Code snippets
      \item Environment descriptions
      \item Conversation/memory
    \end{tabitem} &
    \begin{tabitem}
      \item RGB images
      \item RGB-D video
      \item 3D point clouds
      \item LiDAR
      \item Camera specs
    \end{tabitem} &
    \begin{tabitem}
      \item Image-text pairs
      \item Visual prompts
      \item RGB-D video
      \item Scene descriptions
    \end{tabitem} &
    \begin{tabitem}
      \item RGB-D video
      \item Text instructions
      \item Proprioceptive states
      \item Haptic/tactile feedback
      \item Action trajectories
      \item Success/failure signals
    \end{tabitem} \\
    \midrule

    Output &
    \begin{tabitem}
      \item Logic-based sub-goals
      \item Code snippets
      \item Symbolic plans
      \item Safety constraints
      \item Feedback messages
    \end{tabitem} &
    \begin{tabitem}
      \item Visual features/embeddings
      \item Segmentation masks
      \item Detection/tracking
      \item Depth estimates
      \item Surface keypoints
    \end{tabitem} &
    \begin{tabitem}
      \item Image/video captions
      \item Semantic descriptions
      \item VQA answers
      \item Grounded maps
      \item Multi-modal alignment
    \end{tabitem} &
    \begin{tabitem}
      \item Motor commands
      \item End-effector poses
      \item Action policies
      \item Failure management
    \end{tabitem} \\
    \midrule

    Strengths &
    \begin{tabitem}
      \item Instruction translation
      \item Task decomposition/sequencing
      \item Strong generalization
      \item Easy interaction
    \end{tabitem} &
    \begin{tabitem}
      \item Transfer learning
      \item Open-world perception
      \item Spatial awareness
      \item Distortion robustness
    \end{tabitem} &
    \begin{tabitem}
      \item Semantic understanding
      \item Open-vocabulary recognition
      \item Flexible perception
      \item Novel-entity generalization
    \end{tabitem} &
    \begin{tabitem}
      \item End-to-end simplicity
      \item Cross-platform generalization
      \item Action-reasoning integration
    \end{tabitem} \\
    \midrule

    Limitations &
    \begin{tabitem}
      \item No embodiment/grounding
      \item Hallucinations
      \item High latency
      \item Input bias/sensitivity
    \end{tabitem} &
    \begin{tabitem}
      \item Domain specificity
      \item Weak physics modeling
      \item High compute cost
    \end{tabitem} &
    \begin{tabitem}
      \item No precise actions
      \item Incomplete grounding
      \item Needs external policy
    \end{tabitem} &
    \begin{tabitem}
      \item Large data demand
      \item Cross-platform inefficiency
      \item High latency
      \item Complex failure handling
    \end{tabitem} \\
    \midrule

    Indicative models &
    \begin{tabitem}
      \item BERT \citep{devlin2019bert}, GPT-3 \citep{brown2020gpt3}, Llama 3 \citep{grattafiori2024llama}, DeepSeek-V3 \citep{liu2024deepseek}
    \end{tabitem} &
    \begin{tabitem}
      \item ViT \citep{vit}, DINOv2 \citep{oquab2023dinov2}, SAM \citep{sam}, R3M \citep{nair2022r3m}, VC-1 \citep{majumdar2023we}
    \end{tabitem} &
    \begin{tabitem}
      \item CLIP \citep{clip}, OWL-ViT \citep{minderer2022simple}, BLIP \citep{blip}, SigLIP \citep{zhai2023sigmoid}
    \end{tabitem} &
    \begin{tabitem}
      \item RT-1 \citep{rt1}, PaLM-E \citep{palme}, RT-2 \citep{rt2}, OpenVLA \citep{openvla}, Octo \citep{team2024octo}
    \end{tabitem} \\
    \bottomrule
  \end{tabular}%
  }
\end{table}

\begin{figure}[!t]
    \centering

    \begin{minipage}[t]{0.47\textwidth}
        \centering
        \includegraphics[width=\linewidth]{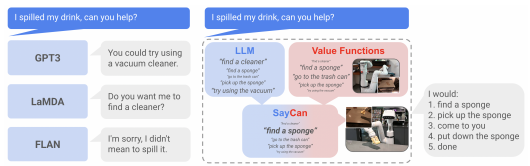}\\[0.3em]
        (a)
    \end{minipage}
    \hfill
    \begin{minipage}[t]{0.47\textwidth}
        \centering
        \includegraphics[width=\linewidth]{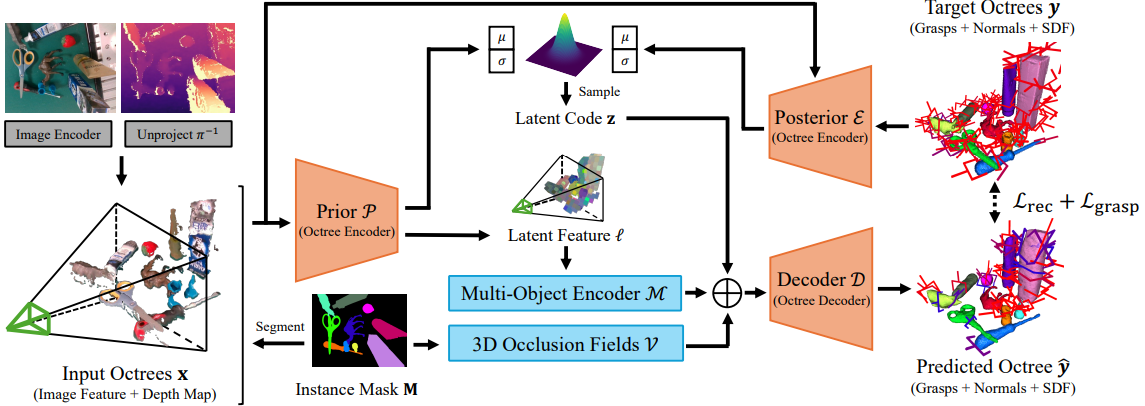}\\[0.3em]
        (b)
    \end{minipage}

    \vspace{1em}

    \begin{minipage}[t]{0.47\textwidth}
        \centering
        \includegraphics[width=\linewidth]{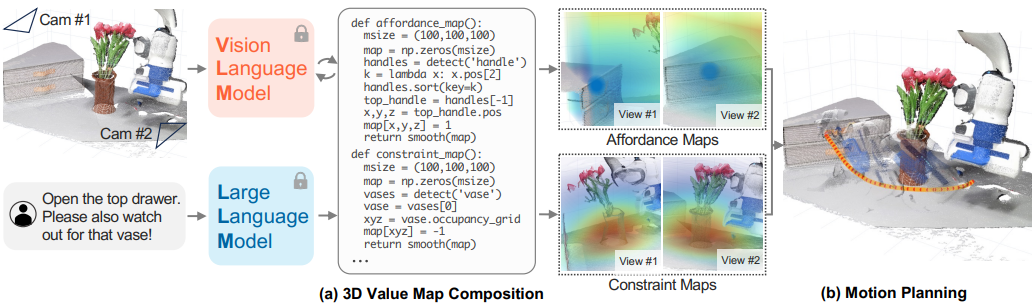}\\[0.3em]
        (c)
    \end{minipage}
    \hfill
    \begin{minipage}[t]{0.47\textwidth}
        \centering
        \includegraphics[width=\linewidth]{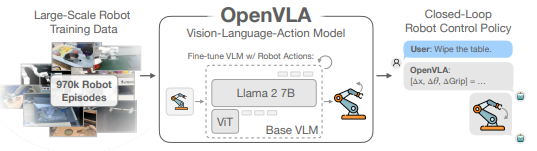}\\[0.3em]
        (d)
    \end{minipage}

    \caption{Representative literature methods per FM type: (a) LLMs (SayCan \citep{saycan}), (b) VFMs (ZeroGrasp \citep{iwase2025zerograsp}), (c) VLMs (VoxPoser \citep{huang2023voxposer}), and (d) VLAs (OpenVLA \citep{openvla}).}
    \label{fig:fm_modalities}
\end{figure}

\subsection{Comparative analysis and key insights}
\label{ssec:FM_comparison}

Having discussed in detail the various types of FMs (Sections \ref{ssec:llm_in_robotics}-\ref{ssec:vla_in_robotics}), this section systematically examines the literature methods, providing a comparative analysis and critical insights for each methodological category. In this respect, Table \ref{tab:fm_types_summary} summarizes for each type of FM its: a) Primary functions, b) Input types, c) Output types, d) Key strengths, e) Critical limitations, and f) Indicative models. Moreover, representative literature methods per FM type are illustrated in Fig. \ref{fig:fm_modalities}.

Based on Table \ref{tab:fm_types_summary}, the following main observations can be drawn: a) LLMs excel at high-level, long-horizon planning and symbolic reasoning, operating on textual inputs but lacking physical grounding, b) VFMs are optimized for real-time, low-level perception, offering strong transfer and robustness, yet remaining domain-specific, c) VLMs bridge language and visual perception, but cannot define precise actions and need external policy generators, d) VLAs provide complete, end-to-end policies mapping multi-modal inputs to actions, at the cost of high data demands and latency, and e) overall, semantic breadth (LLMs, VLMs) trades off against real-time, embodied execution (VFMs, VLAs), motivating hybrid systems.

The full details of the FM types discussed in this section are provided in Appendix~\ref{sec:fm_types_details}. In particular, for each FM type (i.e., LLMs, VFMs, VLMs, and VLAs), the appendix contains the complete category definition, the corresponding main advantages and limitations, as well as the extensive list of representative methods per sub-category of supported functionality.

\section{Neural network architectures}
\label{sec:architectures}

The type of the underlying neural network architecture that is employed in a FM solution largely dictates its capabilities. The main categories of NNs used in robotic FM methods are: a) Transformers, b) State-Space Models (SSMs), c) Diffusion Models (DMs), d) Convolutional and hybrid encoders, and e) Graphical models, as discussed in Section \ref{sec:Taxonomy} and further detailed below.

\subsection{Transformers}
\label{ssec:architectures_transformers}
In the context of robotics, transformers are widely used for diverse tasks (e.g., high-level task planning, low-level policy learning, perception, and human-robot interaction), relying on the fundamental principle of formalizing them as a sequence modeling problem \citep{firoozi2025foundation}. By converting input data (e.g., states, actions, and images) into numerical tokens that are processed as a sequence, they enable long-range dependency modeling, architectural homogenization across tasks and modalities, and efficient high-level reasoning \citep{sanghai2024advances}.

With respect to the input data modality, transformer-based systems can be classified into the following main categories:
\begin{itemize}

    \item \underline{Vision Transformers (ViTs)}: ViTs comprise the fundamental architecture of multiple VFMs in robotics, using self-attention to capture global image relationships, unlike traditional local convolutional methods \citep{vit}. DINOv$2$ \citep{oquab2023dinov2} applies self-supervised learning on web-scale datasets to estimate general-purpose visual features that transfer across tasks without extensive fine-tuning, supporting perception pipelines for scene understanding, semantic mapping, and low-level geometric control.

    \item \underline{Text transformers}: Text transformers act as the language interface, planner, programmer, memory, and supervisor across robotic pipelines, providing a common semantic layer that connects high-level intent to grounded perception and control. Decoder-only models such as PaLM \citep{palm} and GPT-$4$ \citep{gpt4} benefit from scale to improve robotic reasoning and planning, translating free-form language into structured plans, executable policy code, and run-time corrections.

    \item \underline{Multi-modal transformers}: These models transform each sensorial stream into a sequence of tokens, project them to a shared latent space, and align them via a cross-attention or gated fusion mechanism, so that a single backbone can understand scenes, follow instructions, and select actions \citep{openvla,rt2}. They also fuse vision with proprioceptive, geometric, tactile, audio, and thermal cues to generate embodiment-aware policies that scale across diverse robots and tasks.
\end{itemize}

\subsection{State-space models}
\label{ssec:architectures_ssm}
SSMs are increasingly adopted in robotics for realizing real-time control and long-horizon reasoning, by treating the sensorial input streams as latent states and producing output predictions in a step-by-step way \citep{liu2024robomamba}. By learning end-to-end system matrices with hardware-aware modeling \citep{gu2023mamba}, they offer linear-complexity scaling, stable long-horizon memory, and deployment efficiency suitable for embedded robotic solutions \citep{gu2021efficiently}.

With respect to the input modalities, SSM-based systems can be classified into the following main categories:
\begin{itemize}
    \item \underline{Visual SSMs}: Visual SSMs replace attention mechanisms with selection-based ones, resulting into linear-complexity encoders that can be plugged into recognition, dense prediction, and tracking heads \citep{liu2024vmamba}. Long temporal context can be modeled at constant cost, which is particularly suitable for long video streams and multi-camera setups \citep{park2024videomamba}.

    \item \underline{Policy/control SSMs}: SSMs enable data-driven nonlinear reduction in complex systems, such as modeling hysteresis and memory effects. FlowRAM \citep{wang2025flowram} adopts region-aware selective-state policies with flow-matching objectives to learn precise skills from limited demonstrations, while vision-driven locomotion incorporates depth and proprioception through stacked selective-state formalisms and end-to-end RL \citep{wang2025locomamba}.

    \item \underline{Multimodal SSMs}: A single SSM trunk can be pretrained on long videos, robot logs, and demonstrations to support perception, planning interfaces, and action heads. SSM-based VLA models fuse vision, language, and proprioception as token streams to generate actions with lower latency and memory than attention-only implementations \citep{tsuji2025mamba}, making long-horizon policies more practical \citep{liu2024robomamba}.
\end{itemize}

\subsection{Diffusion models}
\label{ssec:architectures_dms}
DMs generate robot behaviours by reversing a gradual noising process, where a forward pass adds Gaussian noise to the input data and a learned reverse model denoises back to the original space (e.g., actions, trajectories, and sub-goals) \citep{ho2020denoising}. Implemented as conditional denoising policies, they are robust across manipulation tasks and serve as generative heads on pretrained backbones, offering multi-modal modelling, composite conditioning, and trajectory-level decision making \citep{chi2025diffusion,liang2025diffusion}.

With respect to the conditioning type, DM-based systems can be classified into the following main categories:
\begin{itemize}

\item \underline{Vision-conditioned DMs}: DMs translate visual goals into structured information for control. Image-goal generation and rearrangement priors estimate object- and scene-level targets that downstream controllers follow \citep{kapelyukh2023dall}, while pretrained image-editing DMs generate sub-goal images from language instructions and current camera views to guide goal-conditioned policies in real-world settings \citep{black2023zero}.

\item \underline{Proprioception-, force-, and haptic-conditioned DMs}: Visuomotor diffusion policies treat action sequences as denoised samples conditioned on images and robot states, handling multi-modal actions and improving stability for manipulation \citep{chi2025diffusion}. When contact requirements are present, conditioning on haptics and force signals can be incorporated, e.g., in visual-tactile slow-fast policies for contact-rich skills \citep{shukla2025learning}.

\item \underline{Language-conditioned DMs}: Textual inputs serve as a guiding signal, where the denoising mechanism modulates goals, trajectories, or sub-goals, simplifying task setup and execution \citep{bjorck2025gr00t}. Further works demonstrate the growing use of language prompts for manipulation and planning tasks based on diffusion backbones \citep{wolf2025diffusion}.

\item \underline{Human behaviour-conditioned DMs}: Diffusion objectives can target early-stage human motion prediction to infer intent, improving intuitiveness and comfort in human-robot interaction without changing the controller structure. The Legibility Diffuser \citep{bronars2024legibility} generates intent-expressive collaborative motions that humans find easier to understand, while still completing the task efficiently \citep{ng2023diffusion}.

\end{itemize}

\subsection{Convolutional and hybrid encoders}
\label{ssec:architectures_cnns}
Visual encoders typically comprise the main perception module of any robotic solution, translating raw pixel data into latent representations that a robot can use for subsequent planning \citep{nair2022r3m}. The choice between CNNs and hybrid CNN-transformer implementations trades off local spatial precision against global context modeling, offering zero-shot generalization, robustness to noise, and reduced need for training data \citep{rt1}.

With respect to the encoder and integration type, the following main categories can be identified:
\begin{itemize}

\item \underline{CNN encoders}: CNNs excel at capturing low-level spatial details such as edges, textures, and object boundaries, due to their local receptive fields. R$3$M \citep{nair2022r3m} freezes a ResNet-50 trained on Ego$4$D to improve manipulation in both simulation and real-world scenarios, while EfficientNet-B3 is employed as a visual encoder for real-time, goal-conditioned navigation and exploration \citep{sridhar2024nomad}.

\item \underline{CNN-transformer hybrids}: Hybrid architectures combine a CNN for pixel-level information with a transformer for context and action aspects. RT-$1$ \citep{rt1} encodes frames with a FiLM-conditioned EfficientNet, compresses them with TokenLearner, and predicts discrete actions using a transformer, while BC-Z/PaLM-SayCan \citep{saycan} couple a lightweight ResNet with a shallow attention network for instruction-conditioned policies.

\item \underline{CNN tokenizers inside generalist agents}: Generalist agents convert high-resolution images into compact tokens prior to sequence modeling. Gato \citep{gato} employs a small ResNet image tokenizer feeding visual, text, and proprioception information to a single transformer, while RoboCat \citep{bousmalis2023robocat} uses a pretrained VQ-GAN tokenizer and a transformer to adapt across robots and tasks via self-improvement cycles.

\item \underline{CNN-conditioned diffusion policies}: Diffusion policies condition a temporal U-Net on CNN features to generate diverse, yet feasible action chunks. Diffusion Policy \citep{chi2025diffusion} employs ResNet-$18$ features for manipulation while preserving low added latency, whereas DiffuserLite \citep{dong2024diffuserlite} uses progressive refinement with a frozen MobileNet-V$3$ encoder to reach real-time prediction rates on embedded platforms.

\end{itemize}

\subsection{Graphical models}
\label{ssec:architectures_graphs}

In the context of robotic FM methods, graphs introduce additional capabilities for connecting low-level, raw sensorial data with high-level, structured reasoning \citep{maggio2024clio}. Unlike architectures that process data as matrices or sequences, graphs model the environment as a set of interconnected entities (e.g., scene graphs with nodes for objects and edges for spatial, semantic, or functional relationships), offering combinatorial generalization, permutation invariance, sample efficiency, and increased explainability \citep{gu2024conceptgraphs}.

With respect to the graph and function type, the following main categories can be identified:
\begin{itemize}

  \item \underline{Scene graphs}: Open-vocabulary $3$D scene graphs associate vision-language features with real-world entities, remaining compact compared to dense maps and enabling robots to query targets, to reason about relationships, and to define sub-goals to planners \citep{gu2024conceptgraphs}. More recently, graphs are constructed online from RGB-D streams, using hierarchical structures for language-grounded navigation \citep{werby2024hierarchical}.

  \item \underline{Shared graphs}: Compressed-form scene graphs allow bandwidth-limited sharing and map merging, while maintaining open-vocabulary query capabilities. Decentralized visual FMs estimate peer poses and produce local Bird's-Eye View maps on embedded hardware, reducing communication requirements without losing key semantic information \citep{blumenkamp2025covis,gu2025mr}.

  \item \underline{Graph neural networks}: Graph Neural Networks (GNNs) enable message passing over task, object, and agent graphs for allocation, scheduling, and policy conditioning in a data-driven way. Recent hybrid cognitive pipelines couple GNN-based scene graphs with LLM or symbolic planners, keeping plans physically feasible while still following language goals \citep{tong2026gnn,strader2025language}.

  \item \underline{Embodiment graphs}: Embodiment graphs encode robot joint information and the links between them, allowing a single learned policy to adapt across platforms. Attention or message passing follows the learned graph connectivity, boosting zero-shot transfer to new morphologies and supporting reusable controllers across different robots \citep{patel2025get}.

\end{itemize}

\begin{table}[t]
  \caption{Neural network architectures: Comparative analysis and key insights.}
  \label{tab:nn_architectures_summary}
  \centering
  \scriptsize
  
  \setlength{\aboverulesep}{0pt}
  \setlength{\belowrulesep}{0pt}
  \setlength{\tabcolsep}{2pt}
  \renewcommand{\arraystretch}{0.9}

  \setlist*[tabitem]{before=\vspace{2.2pt}\justifying, after=\vspace{2.2pt}}

  \rowcolors{2}{gray!25}{white}

  \newlength{\Warch}\setlength{\Warch}{1.8cm}
  \newlength{\Wtrans}\setlength{\Wtrans}{3cm}
  \newlength{\Wssm}\setlength{\Wssm}{3cm}
  \newlength{\Wdm}\setlength{\Wdm}{3cm}
  \newlength{\Wcnn}\setlength{\Wcnn}{3cm}
  \newlength{\Wgraph}\setlength{\Wgraph}{3cm}

  \resizebox{\textwidth}{!}{%
  \begin{tabular}{@{}|
    >{\justifying\arraybackslash}m{\Warch}|
    >{\raggedright\arraybackslash}m{\Wtrans}|
    >{\raggedright\arraybackslash}m{\Wssm}|
    >{\raggedright\arraybackslash}m{\Wdm}|
    >{\raggedright\arraybackslash}m{\Wcnn}|
    >{\raggedright\arraybackslash}m{\Wgraph}|
  @{}} 
    \toprule
    \rowcolor{gray!40}
    \headerbreak{Architecture} &
    \headerbreak{Transformers} &
    \headerbreak{SSMs} &
    \headerbreak{DMs} &
    \headerbreak{CNNs/hybrid} &
    \headerbreak{Graphical models} \\
    \midrule

    Primary functions &
    \begin{tabitem}
      \item Multi-modal alignment
      \item High-level reasoning
      \item Task decomposition
      \item Cross-embodiment transfer
    \end{tabitem} &
    \begin{tabitem}
      \item Sequence modeling
      \item Real-time edge control
      \item State estimation
      \item Contextual memory
    \end{tabitem} &
    \begin{tabitem}
      \item Action generation
      \item Precise manipulation
      \item Receding-horizon control
      \item Score estimation
    \end{tabitem} &
    \begin{tabitem}
      \item Feature detection
      \item Spatial grounding
      \item Multi-objective perception
      \item Object classification
    \end{tabitem} &
    \begin{tabitem}
      \item Causal reasoning
      \item Structured planning
      \item Relational grounding
      \item State transitions
    \end{tabitem} \\
    \midrule

    Main mechanisms &
    \begin{tabitem}
      \item Global self-attention
      \item Positional encoding
      \item Autoregressive prediction
      \item Chain-of-thought
    \end{tabitem} &
    \begin{tabitem}
      \item Dynamics discretization
      \item Selective scan operators
      \item Hardware-aware kernels
      \item Input-dependent gating
    \end{tabitem} &
    \begin{tabitem}
      \item Score-based denoising
      \item Langevin dynamics
      \item Action chunking
      \item Latent space diffusion
    \end{tabitem} &
    \begin{tabitem}
      \item Convolutional layers
      \item Local connectivity
      \item Parameter sharing
      \item Pooling operators
    \end{tabitem} &
    \begin{tabitem}
      \item Graph neural networks
      \item Symbolic reasoning
      \item Entity masking
      \item Scene graph serialization
    \end{tabitem} \\
    \midrule

    Strengths &
    \begin{tabitem}
      \item Long-range dependency
      \item Architectural homogenization
      \item Planning efficiency
    \end{tabitem} &
    \begin{tabitem}
      \item Linear scaling
      \item Long-horizon memory
      \item Deployment efficiency
    \end{tabitem} &
    \begin{tabitem}
      \item Multi-modal modelling
      \item Composite conditioning
      \item Trajectory decision-making
    \end{tabitem} &
    \begin{tabitem}
      \item Zero-shot generalization
      \item Noise robustness
      \item Data efficiency
    \end{tabitem} &
    \begin{tabitem}
      \item Combinatorial generalization
      \item Permutation invariance
      \item Sample efficiency
    \end{tabitem} \\
    \midrule

    Limitations &
    \begin{tabitem}
      \item Quadratic complexity
      \item Discrete tokenization
      \item Context contradiction
    \end{tabitem} &
    \begin{tabitem}
      \item Weak cross-token check
      \item Limited global context
      \item Hybrid-design need
    \end{tabitem} &
    \begin{tabitem}
      \item Sampling latency/cost
      \item No safety guarantees
      \item Conditioning drift
    \end{tabitem} &
    \begin{tabitem}
      \item High latency
      \item Fine-detail loss
      \item Distribution shifts
    \end{tabitem} &
    \begin{tabitem}
      \item Computational overhead
      \item Dynamic topology
      \item Latent space integration
    \end{tabitem} \\
    \midrule

    Indicative models &
    \begin{tabitem}
      \item RT-2 \citep{rt2}, PaLM-E \citep{palme}, Gato \citep{gato}, OpenVLA \citep{openvla}, Octo \citep{team2024octo}
    \end{tabitem} &
    \begin{tabitem}
      \item RoboMamba \citep{liu2024robomamba}, Mamba \citep{gu2023mamba}, AnoleVLA \citep{takagi2026anolevla}, Decision Mamba \citep{huang2024decision}
    \end{tabitem} &
    \begin{tabitem}
      \item Diffusion Policy \citep{chi2025diffusion}, Diffuser \citep{janner2022planning}, Motion Planning Diffusion \citep{carvalho2023motion}, M2Diffuser \citep{yan2025m}
    \end{tabitem} &
    \begin{tabitem}
      \item RT-1 \citep{rt1}, R3M \citep{nair2022r3m}, VC-1 \citep{majumdar2023we}, MVP \citep{wei2022mvp}
    \end{tabitem} &
    \begin{tabitem}
      \item GRID \citep{ni2024grid}, ConceptGraphs \citep{gu2024conceptgraphs}, HOV-SG \citep{werby2024hierarchical}, Open3DSG \citep{koch2024open3dsg}
    \end{tabitem} \\
    \bottomrule
  \end{tabular}%
  }
\end{table}

\subsection{Comparative analysis and key insights}
\label{sec:architectures_discussion}

\begin{figure}[!t]
    \centering

    \begin{minipage}[t]{0.46\textwidth}
        \centering
        \includegraphics[width=\linewidth]{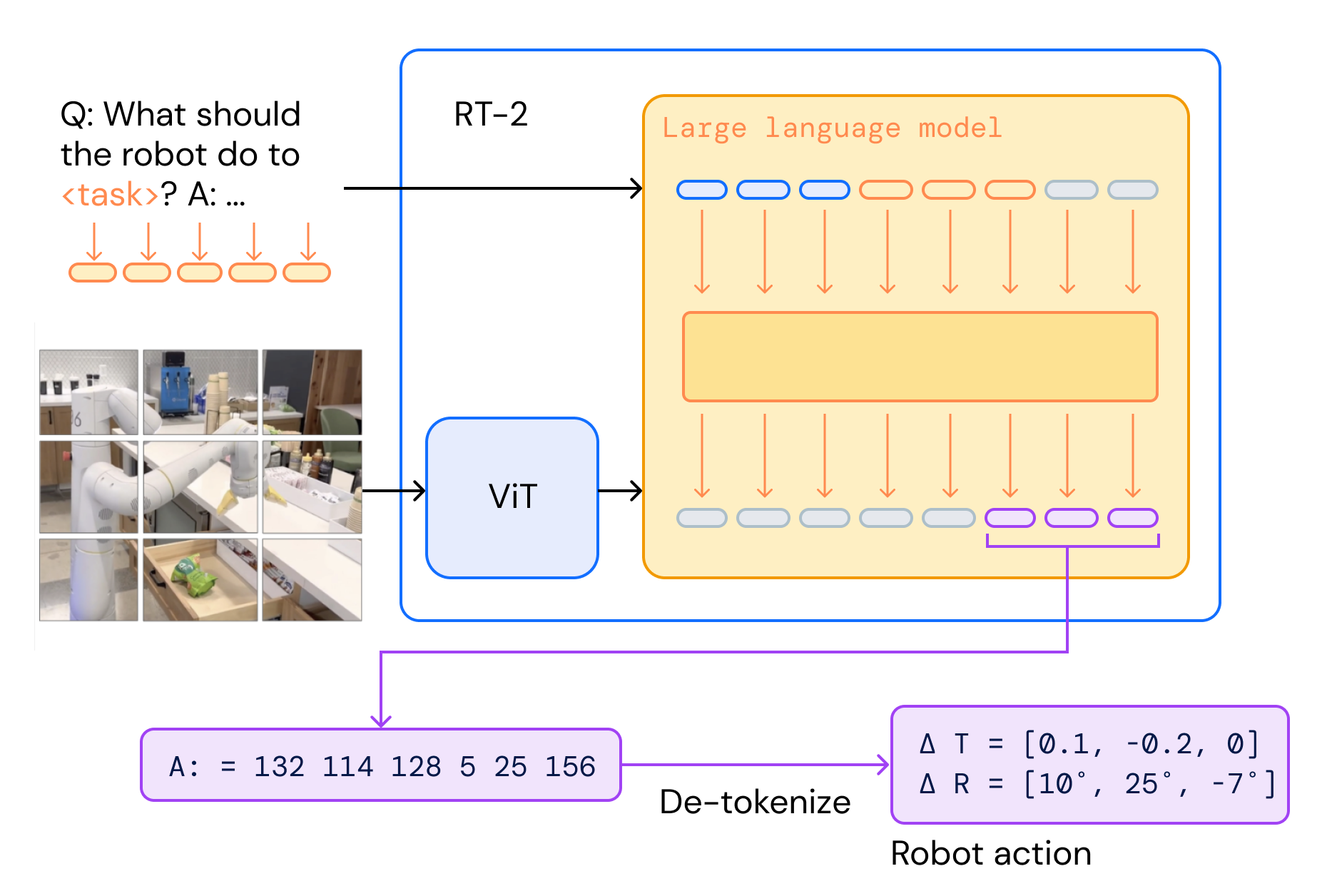}\\[0.3em]
        (a)
    \end{minipage}
    \hspace{0.04\textwidth}
    \begin{minipage}[t]{0.46\textwidth}
        \centering
        \includegraphics[width=\linewidth]{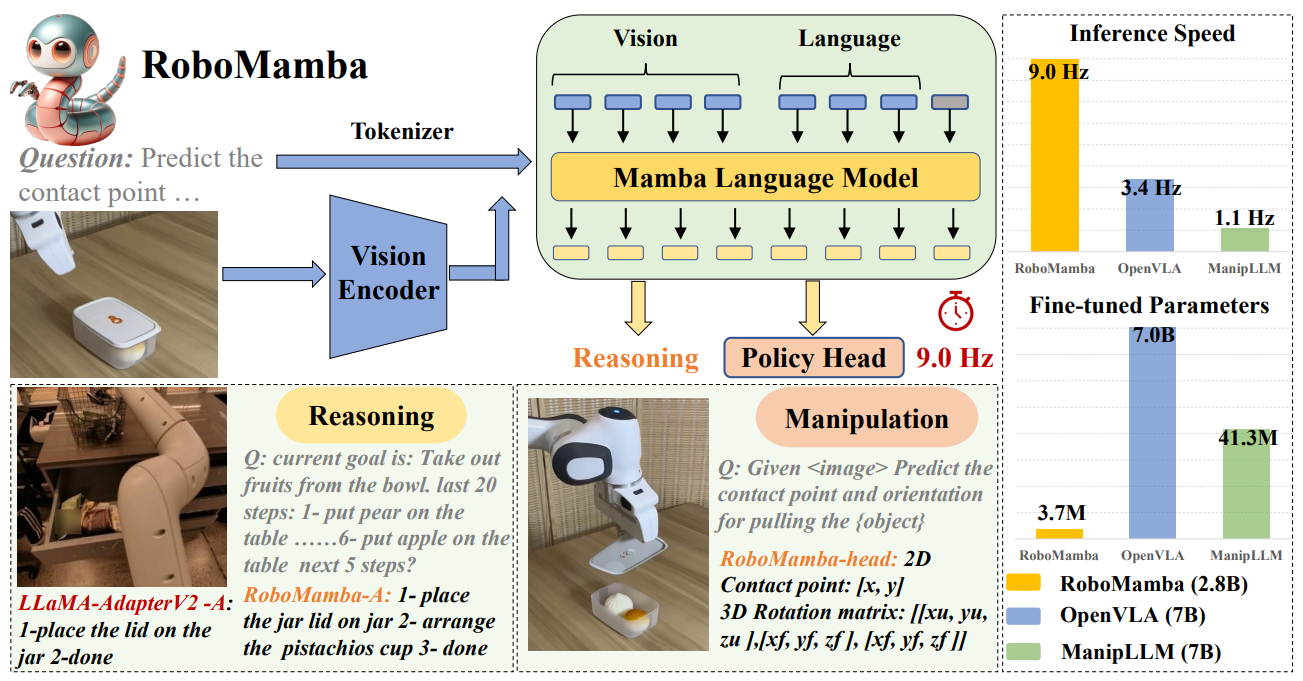}\\[0.3em]
        (b)
    \end{minipage}

    \vspace{1.2em}

    \begin{minipage}[t]{0.46\textwidth}
        \centering
        \includegraphics[width=\linewidth]{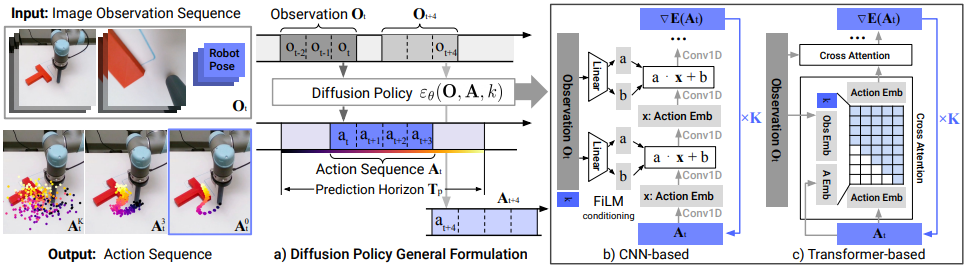}\\[0.3em]
        (c)
    \end{minipage}
    \hspace{0.04\textwidth}
    \begin{minipage}[t]{0.46\textwidth}
        \centering
        \includegraphics[width=\linewidth]{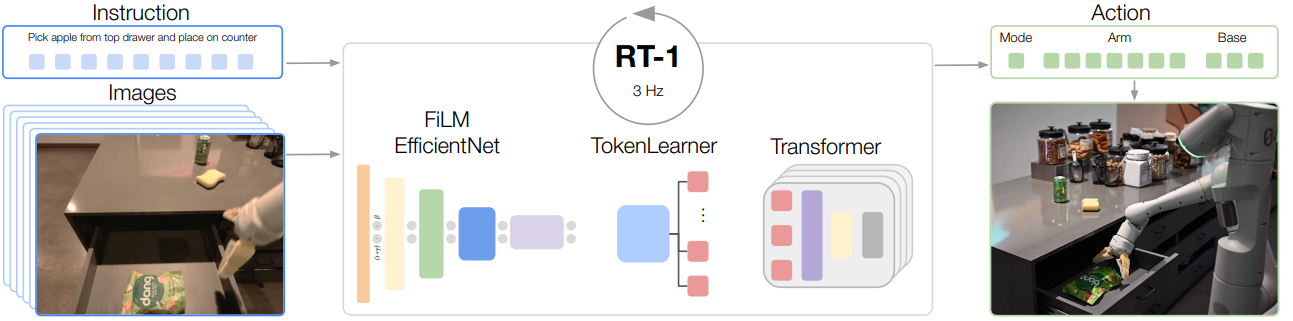}\\[0.3em]
        (d)
    \end{minipage}

    \vspace{1.2em}

    \begin{minipage}[t]{0.46\textwidth}
        \centering
        \includegraphics[width=\linewidth]{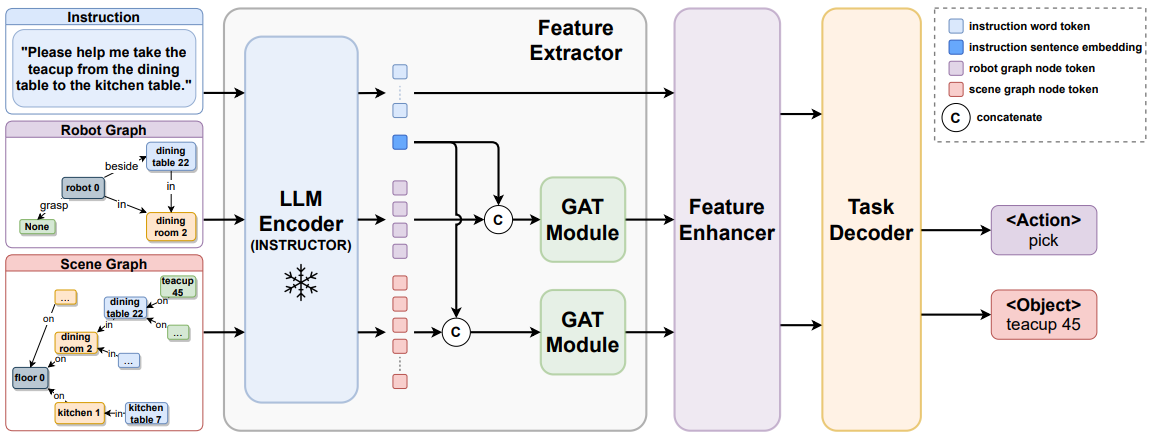}\\[0.3em]
        (e)
    \end{minipage}

    \caption{Representative literature methods incorporating different NN architecture types: (a) Transformers (RT-2 \citep{rt2}), (b) State-space models (RoboMamba \citep{liu2024robomamba}), (c) Diffusion models (Diffusion Policy \citep{chi2025diffusion}), (d) Convolutional and hybrid encoders (RT-1 \citep{rt1}), and (e) Graphical models (GRID \citep{ni2024grid}).}
    \label{fig:nn_architectures_modalities}
\end{figure}

Having discussed in detail the various types of neural network architectures (Sections \ref{ssec:architectures_transformers}-\ref{ssec:architectures_graphs}), this section systematically examines the literature methods, providing a comparative analysis and critical insights for each category. In this respect, Table \ref{tab:nn_architectures_summary} summarizes for each type of architecture its: a) Primary functions, b) Main mechanisms, c) Key strengths, d) Critical limitations, and e) Indicative models. Moreover, representative literature methods incorporating different NN architecture types are illustrated in Fig. \ref{fig:nn_architectures_modalities}.

Based on Table \ref{tab:nn_architectures_summary}, the following main observations can be drawn: a) Transformers excel at multi-modal alignment and high-level reasoning through global self-attention, but their quadratic complexity and discrete tokenization hinder real-time control, b) SSMs provide linear-complexity temporal modeling and stable long-horizon memory for real-time edge control, yet offer weaker global context than full attention, c) DMs generate multi-modal, high-precision actions via score-based denoising, at the cost of sampling latency and the absence of built-in safety guarantees, d) CNN and hybrid encoders serve as robust, zero-shot perceptual backbones for pixel-level grounding, but suffer high latency and sensitivity to distribution shifts, and e) Graphical models support structured, relational, and causal reasoning with strong sample efficiency, though they incur message-passing overhead and struggle with dynamic topologies.

The full details of the neural network architectures discussed in this section are provided in Appendix~\ref{sec:nn_architectures_details}. In particular, for each architecture type (i.e., transformers, SSMs, DMs, convolutional and hybrid encoders, and graphical models), the appendix contains the complete category definition, the corresponding main advantages and limitations, as well as the extensive list of representative methods per sub-category.

\section{Learning paradigms}
\label{sec:learningParadigm}

In order to develop robust, real-world robotic FM solutions, different/diverse learning techniques, principles, and approaches can be adopted. The most commonly met learning paradigms in the literature, which are typically combined in a comprehensive learning methodology, are: a) Supervised Learning (SL), b) Self-Supervised Learning (SSL), c) Fine-tuning, d) Domain Adaptation (DA), e) Imitation Learning (IL), f) Reinforcement Learning (RL), g) In-Context/prompt Learning (ICL), h) World Model (WM) learning, and i) Generative Learning (GL), as discussed in Section \ref{sec:Taxonomy} and further detailed below.

\subsection{Supervised learning}
\label{ssec:paradigms_supervised}
Supervised learning trains a FM on large-scale, labeled datasets, directly fitting a mapping from inputs (e.g., images, language instructions, and proprioceptive states) to target outputs (e.g., class labels, tokens, or action commands) under explicit human-provided supervision \citep{xiao2025robot}. This offers high task accuracy, stable optimization, and effective transfer of broad semantic priors, when paired with internet-scale labeled corpora \citep{rt2}. In this context, PaLM-E \citep{palme} is trained on web-scale labeled multi-modal data jointly with embodied experiences, while RT-$2$ \citep{rt2} couples web-scale vision-language supervision with labeled robot manipulation data to enable open-vocabulary, instruction-following control.

\subsection{Self-supervised learning}
\label{ssec:paradigms_ssl}

SSL techniques employ `pretext tasks' (e.g., predicting the next video frame or reconstructing a masked image patch) and large quantities of unlabeled data streams to enable FMs to acquire common-sense knowledge regarding physics, object permanence, and spatial relationships \citep{he2022masked,oquab2023dinov2}. This offers massive scalability, sample efficiency, zero-shot generalization, and autonomous improvement, without requiring human supervision \citep{nair2022r3m,assran2025v}.

The most common SSL techniques used for developing robotic FM solutions are:

\begin{itemize}
  \item \underline{Masked Autoencoder (MAE)}: The model learns by reconstructing missing or corrupted parts of the input data, modeling robust spatial and temporal features. Image MAE \citep{he2022masked} and its video counterpart VideoMAE \citep{tong2022videomae} are widely used for producing robust visual backbone networks.

  \item \underline{Contrastive learning}: This aligns matched views (and separates mismatched ones) to create discriminative features and to support open-vocabulary grounding, linking language to perception. CLIP \citep{clip} establishes vision-language alignment at scale, while egocentric robot features such as R$3$M \citep{nair2022r3m} improve manipulation sample-efficiency from human-performing videos.

  \item \underline{Autoregressive sequence modeling}: This predicts the next token in vision, language, or action streams to capture long-horizon structure and to enable unified perception-to-policy modeling. Generalist agents, such as Gato \citep{gato}, condition on images, text, and proprioception to produce actions across many tasks, while LLMs, like GPT-$4$ \citep{gpt4}, showcase cross-domain reasoning in embodied pipelines.

  \item \underline{World model learning}: Also reported as an individual paradigm (Section \ref{ssec:worldmodel}), WMs learn to predict how the world changes in response to specific actions, encoding causal relations. Representative methods include Dreamer-style agents on physical robots \citep{wu2023daydreamer} and compositional video world models such as RoboDreamer \citep{zhou2024robodreamer}.

\end{itemize}

\subsection{Fine-tuning}
\label{ssec:paradigms_fine-tuning}

Fine-tuning adapts the internet-scale acquired knowledge of a pretrained FM to specific physical-world execution settings, adjusting its common-sense knowledge structures to a specific robot, sensor suite, or task, using a smaller, in-domain, annotated dataset \citep{yu2025survey}. This offers sample efficiency, domain adaptation, and improved precision for the target task \citep{team2024octo}.

The most common fine-tuning techniques in robotics are:
\begin{itemize}
    \item\underline{Full fine-tuning}: This updates all FM weights using a relatively small, in-domain robot dataset, to re-target a pretrained policy to a new robot, sensor setup, or task. Recent generalist policies report fast adaptation to new observation and action spaces on standard GPUs, using full fine-tuning as the baseline learning method \citep{team2024octo,openvla}.

    \item\underline{Low-Rank Adaptation (LoRA)}: This keeps the original FM weights unchanged and adds two low-rank matrices that incorporate the required modifications. OpenVLA \citep{openvla} demonstrates LoRA-based tuning on the large-scale Open-X-Embodiment dataset, while more recent quantized variants target adaptation on resource-constrained platforms \citep{williams2026litevla}.

    \item\underline{Quantized parameter-efficient fine-tuning}: This extends PEFT by combining low-precision weights with LoRA-style adapters to keep latency and memory low on embedded hardware, while retaining most full-precision performance. LiteVLA \citep{williams2026litevla} reports that $4$-$8$ bit quantization plus adapters preserves high recognition rates, while enabling real-time control on smaller platforms \citep{williams2025lite}.

    \item\underline{Action space remapping}: This integrates lightweight encoders/decoders or tokenizers so that the core policy can be fine-tuned to new sensors or actuators without retraining the FM from scratch. Generalist policies, such as Octo \citep{team2024octo} and RT-$2$ \citep{rt2}, re-target a FM to multiple robots and grippers with modest additional data.

\end{itemize}

\subsection{Domain adaptation}
\label{ssec:paradigms_domain-adaptation}

Domain Adaptation (DA) bridges the critical gap between FMs pre-trained on massive internet-scale data and their deployment in real-world environments, equipping them with the required physical-world grounding for given application scenarios (the `sim-to-real' transfer challenge) \citep{da2025survey}. This reduces the need for real-world samples, improves robustness to sensor shifts, and lowers the risk of damage during the learning phase \citep{tayyab2025foundation}.

The most common DA techniques in robotics are:
\begin{itemize}

  \item \underline{Sim-to-real transfer}: This trains a FM using large corpora of simulated/synthetic data, evaluates performance in simulation suites, and eventually calibrates on real-world hardware (typically using some real-world data). Humanoid and manipulation solutions demonstrate the efficiency of this approach \citep{luo2026simvla,deng2025graspvla}.

  \item \underline{Real-to-sim-to-real transfer}: This replays real-world robot trajectories in high-fidelity simulations, diversifies the scenes and objects, synthesizes new training data (often with domain randomization), and eventually calibrates the FM to real-world specifications. Various works demonstrate the validity of this approach in diverse operational settings \citep{zhu2025vr,fang2025rebot}.
\end{itemize}

\subsection{Imitation learning}
\label{ssec:paradigms_imitation-learning}

Imitation Learning (IL) enables a model to learn directly from a (human) expert via teleoperation or video demonstrations of the desired skills, providing efficient multi-modal alignment that maps high-level language instructions and visual inputs to low-level motor commands \citep{zare2024survey}. This requires no reward-signal definition, needs relatively few demonstrations, and offers easy training supervision \citep{kawaharazuka2025vision}.

The most common IL techniques in robotics are:
\begin{itemize}
  \item \underline{Behavioral cloning}: This is a particular type of Supervised Fine-Tuning (SFT), imitating human expert demonstrations. RT-$1$ \citep{rt1} learns a direct mapping of observations and language goals to actions using large-scale transformers, while RT-$2$ \citep{rt2} extends this with web-scale vision-language pretraining to open-vocabulary, instruction-following control.

  \item \underline{Diffusion-based IL}: This represents a robot's behavior as a conditional denoising process, treating actions as a data distribution iteratively refined from random noise. Diffusion Policy \citep{chi2025diffusion} and Diff-Dagger \citep{lee2025diff} generate action sequences that match expert behavior and handle multi-modal inputs, improving stability for long-horizon manipulation.

  \item \underline{In-context IL}: This enables zero- or few-shot task adaptation on the fly, without updating model weights. ICRT \citep{fu2025icrt} uses next-token prediction over sensorimotor streams for real-robot in-context imitation, while prompt demonstrations are augmented with explicit visual reasoning traces to infer task intent more reliably in ambiguous environments \citep{nguyen2026iclr}.

  \item \underline{Continual IL}: This addresses the long-term memory and evolution challenges of robotic FMs, enabling a robot to sequentially acquire new skills over time without forgetting previously learned ones. LOTUS \citep{wan2024lotus} introduces a continual imitation learning framework for skill acquisition by a real robot.
\end{itemize}

\subsection{Reinforcement learning}
\label{ssec:paradigms_reinforcement-learning}
Reinforcement Learning (RL) serves as the optimization formalism that bridges the gap between high-level, semantic reasoning (supported by FMs) and low-level, physical robot actions, involving the FM in a continuous, self-improvement cycle of perception, action, and evaluation \citep{tang2025deep}. This enables self-improvement beyond the training data, increased generalization, and efficient sim-to-real implementation \citep{ter2025taxonomy}.

The most common RL techniques in robotics are:
\begin{itemize}
  \item \underline{SFT-to-RL}: This is a two-stage process where BC first learns a policy and RL subsequently improves it. RT-$1$ \citep{rt1} and RT-$2$ \citep{rt2} show how large-scale BC produces powerful priors refined through interaction, while ExploRLLM \citep{ma2025explorllm} combines an LLM-guided exploration policy with a residual RL head to improve sample efficiency.

  \item \underline{LLM-guided reward design}: This leverages LLMs to write/refine RL reward code and tune domain randomization, improving robustness and transfer. Eureka \citep{ma2023eureka} automates reward design and outperforms expert rewards on multiple tasks, while DrEureka \citep{ma2024dreureka} extends this to the sim-to-real setting by jointly optimizing rewards and randomization.

  \item \underline{Preference-based RL}: This replaces detailed/numeric rewards with preferences produced by VLMs (or adapted ones with small-scale human intervention). RL-VLM-F \citep{wang2024rl} models rewards from VLM comparisons over image observations and text, while VARP \citep{singh2025varp} regularizes VLM-derived preferences with the agent's own rollouts to reduce misalignment.

  \item \underline{Offline-to-online RL}: This trains a model offline on massive datasets, refines it via offline RL, and eventually applies online RL for real-world deployment. Embodied visual tracking is combined with a text-promptable encoder and offline RL for improved perception \citep{zhong2024empowering}, while FLaRe \citep{hu2025flare} applies large-scale RL fine-tuning on a pre-trained VLA for adaptive manipulation.

  \item \underline{World-model RL}: World-model RL learns a generative world model with language-aware structure and uses it for RL-based policy improvement. RoboDreamer \citep{zhou2024robodreamer} factors video generation into compositional parts conditioned by language and visual goals, exhibiting robust performance on long-horizon tasks.

\end{itemize}

\subsection{In-context/prompt learning}
\label{ssec:paradigms_in-context-learning}
ICL and prompt learning adapt FMs at inference time by conditioning their behavior on demonstrations, examples, or task-specific instructions, without updating the underlying model weights \citep{fu2025icrt,yin2024context}. This offers generalization to novel settings, high-level task planning from natural-language guidance, and flexible multimodal task specification \citep{yao2023react}.

The most common in-context/prompt learning techniques in robotics are:
\begin{itemize}
  \item \underline{Language prompting}: This uses natural language instructions to guide a robot's behavior, decision-making, and physical actions. Few-shot language prompts can encode demonstrations or templates, so that an LLM can output low-level actions or acquire new skills \citep{yin2024context,liang2023code}.

  \item \underline{Reason-act prompting (ReAct)}: This interleaves natural language reasoning with physical actions, allowing a robot to decompose complex goals, validate its progress, and dynamically adjust its plan. Planning, execution, and re-planning can be performed in a single loop, including an LLM-based verification step \citep{yao2023react,grigorev2025verifyllm}.

  \item \underline{In-context imitation}: This enables a FM to perform a novel task by observing a few videos or sensorimotor demonstrations, without permanent weight changes. A causal policy can parse short teleoperation trajectories as a prompt and predict the next action for new tasks without fine-tuning \citep{fu2025icrt}.
\end{itemize}

\subsection{World model learning}
\label{ssec:worldmodel}

World Models (WMs) allow robots to predict environmental changes in response to their actions, decoupling perception from action so that, instead of operating only on pixel values, the robot learns the underlying physics of the world \citep{li2025comprehensive}. This incorporates `imagined' experiences that reduce physical-world trials, models the rules of physics, and reduces operational delays \citep{zhang2025step}.

The most common WM learning techniques in robotics are:
\begin{itemize}
  \item \underline{Feature-space WMs}: Instead of predicting each pixel, which is expensive and noisy, feature-space WMs map visual inputs to an abstract feature space and predict the future there. Future DINOv2 patch embeddings are predicted from offline trajectories and action sequences are optimized in the embedding space for zero-shot planning \citep{zhou2025dino}.

  \item \underline{Latent-action WMs}: These learn to model the underlying physics and intent of actions, rather than specific skills. Continuous latent actions are discovered from videos and an auto-regressive WM conditioned on those actions transfers skills across scenes and embodiments with small-scale finetuning \citep{gao2025adaworld}.

  \item \underline{Compositional video WMs}: These factorize the environment into its constituent parts (objects, relationships, and action primitives) and recombine them to generate future scenarios. Videos are factorized into objects and relations so the model can synthesize plans for unseen goal-scene combinations and guide long-horizon decisions \citep{zhou2024robodreamer}.

  \item \underline{JEPA-style WMs}: These predict an abstract meaning of what will happen next, enabling robots to plan complex tasks without being distracted by irrelevant noise. Joint predictions of short-horizon actions and abstract observations couple imitation with predictive learning to reduce control-error accumulation \citep{vujinovic2025act}.
\end{itemize}

\subsection{Generative learning}
\label{ssec:paradigms_generative-learning}

Generative Learning (GL) enables robots to imagine future states, to synthesize training data, and to propose complex action sequences, leveraging the capability of generative FMs to produce large quantities of data samples and alleviating the need for extensive high-quality robotic interaction data \citep{zhang2025generative}. This offers increased zero-shot generalization, multi-modality handling, and long-horizon planning \citep{liu2024rdt}.

The most common GL techniques in robotics are:

\begin{itemize}
    \item \underline{Autoregressive sequence modeling}: This predicts the next action or state based on previous observations. PACT \citep{bonatti2023pact} trains a causal transformer to predict the next observation-action token, so that a single model captures long-horizon structure across tasks, while long-horizon manipulation is modeled through sequential generation of action tokens \citep{zhang2025autoregressive}.

    \item \underline{Diffusion-based action policies}: This employs a diffusion model to generate a chunk of actions at once via a gradual denoising process. Diffusion Policy \citep{chi2025diffusion} learns a conditional denoising process that samples action sequences for multi-modal behaviors and stable visuomotor control, while the Legibility Diffuser \citep{bronars2024legibility} is an intent-expressive variant.

    \item \underline{Generative video and scene synthesis}: This creates a model of physical reality, enabling robots to imagine, simulate, and plan actions prior to real-world execution. RoboDreamer \citep{zhou2024robodreamer} employs compositional video WMs, while ReBot \citep{fang2025rebot} uses a real-to-sim-to-real synthesis approach.

\end{itemize}

\begin{table}[!t]
  \caption{Learning paradigms: Comparative analysis and key insights.}
  \label{tab:learning_paradigms_summary}
  \centering
  \scriptsize
  
  \setlength{\aboverulesep}{0pt}
  \setlength{\belowrulesep}{0pt}
  \setlength{\tabcolsep}{2pt}
  \renewcommand{\arraystretch}{0.9}

  \setlist*[tabitem]{before=\vspace{2.2pt}\justifying, after=\vspace{2.2pt}}

  \rowcolors{2}{gray!25}{white}

  \newlength{\Wlp}\setlength{\Wlp}{1.7cm}
  \newlength{\Wfunc}\setlength{\Wfunc}{2.0cm}
  \newlength{\Wmech}\setlength{\Wmech}{2.0cm}
  \newlength{\Wdata}\setlength{\Wdata}{2.0cm}
  \newlength{\Wstr}\setlength{\Wstr}{3.0cm}
  \newlength{\Wlim}\setlength{\Wlim}{3.0cm}
  \newlength{\Wmod}\setlength{\Wmod}{3.0cm}

  \resizebox{\textwidth}{!}{%
  \begin{tabular}{@{}|
    >{\justifying\arraybackslash}m{\Wlp}|
    >{\raggedright\arraybackslash}m{\Wfunc}|
    >{\raggedright\arraybackslash}m{\Wmech}|
    >{\raggedright\arraybackslash}m{\Wdata}|
    >{\raggedright\arraybackslash}m{\Wstr}|
    >{\raggedright\arraybackslash}m{\Wlim}|
    >{\raggedright\arraybackslash}m{\Wmod}|
  @{}}
    \toprule
    \rowcolor{gray!40}
    \headerbreak{Learning\\paradigm} &
    \headerbreak{Primary\\function} &
    \headerbreak{Main\\mechanisms} &
    \headerbreak{Primary\\data source} &
    \headerbreak{Strengths} &
    \headerbreak{Limitations} &
    \headerbreak{Indicative\\models} \\
    \midrule

    Supervised learning &
    \begin{tabitem}
      \item General representation learning
    \end{tabitem} &
    \begin{tabitem}
      \item Large-scale labeled supervision
    \end{tabitem} &
    \begin{tabitem}
      \item Internet-scale data
    \end{tabitem} &
    \begin{tabitem}
      \item Data efficiency
      \item Zero-shot generalization
      \item Knowledge transfer
      \item Emergent reasoning
    \end{tabitem} &
    \begin{tabitem}
      \item Embodiment gap
      \item Sim-to-real gap
      \item High compute cost
      \item Safety concerns
    \end{tabitem} &
    \begin{tabitem}
      \item PaLM-E \citep{palme}, RT-2 \citep{rt2}, Octo \citep{team2024octo}, OpenVLA \citep{openvla}
    \end{tabitem} \\
    \midrule

    Self-supervised learning &
    \begin{tabitem}
      \item Label-free representation learning
    \end{tabitem} &
    \begin{tabitem}
      \item Pretext tasks
    \end{tabitem} &
    \begin{tabitem}
      \item Unlabeled robot trajectories
      \item Egocentric videos
      \item Raw sensor data
    \end{tabitem} &
    \begin{tabitem}
      \item Scalability
      \item Sample efficiency
      \item Zero-shot generalization
      \item Autonomous improvement
    \end{tabitem} &
    \begin{tabitem}
      \item Embodiment gap
      \item High compute cost
      \item Hallucinations
    \end{tabitem} &
    \begin{tabitem}
      \item R3M \citep{nair2022r3m}, MVP \citep{wei2022mvp}, DINOv2 \citep{oquab2023dinov2}, Masked Autoencoders \citep{he2022masked}
    \end{tabitem} \\
    \midrule

    Fine-tuning &
    \begin{tabitem}
      \item Task/domain adaptation
    \end{tabitem} &
    \begin{tabitem}
      \item Supervised update
    \end{tabitem} &
    \begin{tabitem}
      \item Action-labeled demonstrations
      \item Domain-specific instructions
    \end{tabitem} &
    \begin{tabitem}
      \item Sample efficiency
      \item Domain adaptation
      \item Improved precision
    \end{tabitem} &
    \begin{tabitem}
      \item Catastrophic forgetting
      \item Overfitting
      \item Annotation need
    \end{tabitem} &
    \begin{tabitem}
      \item OpenVLA \citep{openvla}, Octo \citep{team2024octo}, RoboCat \citep{bousmalis2023robocat}, RT-2 \citep{rt2}
    \end{tabitem} \\
    \midrule

    Domain adaptation &
    \begin{tabitem}
      \item Sim-to-real bridging
    \end{tabitem} &
    \begin{tabitem}
      \item Feature alignment
      \item Adversarial training
      \item Distribution reweighting
    \end{tabitem} &
    \begin{tabitem}
      \item Synthetic \& sparse real data
    \end{tabitem} &
    \begin{tabitem}
      \item Reduced real-world data
      \item Sensor-shift robustness
      \item Transferability
    \end{tabitem} &
    \begin{tabitem}
      \item Negative transfer
      \item Training instability
      \item Low predictability
    \end{tabitem} &
    \begin{tabitem}
      \item DrEureka \citep{ma2024dreureka}, Gen2Sim \citep{katara2024gen2sim}, ReBot \citep{fang2025rebot}, VR-Robo \citep{zhu2025vr}
    \end{tabitem} \\
    \midrule

    Imitation learning &
    \begin{tabitem}
      \item Expert behavior replication
    \end{tabitem} &
    \begin{tabitem}
      \item Behavioral cloning
    \end{tabitem} &
    \begin{tabitem}
      \item Expert teleoperation
      \item Kinesthetic teaching
      \item Human videos
    \end{tabitem} &
    \begin{tabitem}
      \item No reward signal
      \item Data efficiency
      \item Easy supervision
    \end{tabitem} &
    \begin{tabitem}
      \item Data-quality dependence
      \item Covariate shift
      \item Causal confusion
    \end{tabitem} &
    \begin{tabitem}
      \item RT-1 \citep{rt1}, Gato \citep{gato}, Octo \citep{team2024octo}, BC-Z \citep{jang2022bc}
    \end{tabitem} \\
    \midrule

    Reinforcement learning &
    \begin{tabitem}
      \item Interaction-based policy learning
    \end{tabitem} &
    \begin{tabitem}
      \item Reward-driven optimization
    \end{tabitem} &
    \begin{tabitem}
      \item Simulation/ real interaction data
    \end{tabitem} &
    \begin{tabitem}
      \item Self-improvement
      \item Generalization
      \item Efficient sim-to-real
    \end{tabitem} &
    \begin{tabitem}
      \item Sample inefficiency
      \item Reward engineering
      \item Credit assignment
    \end{tabitem} &
    \begin{tabitem}
      \item Eureka \citep{ma2023eureka}, DrEureka \citep{ma2024dreureka}, RL-VLM-F \citep{wang2024rl}, ExploRLLM \citep{ma2025explorllm}
    \end{tabitem} \\
    \midrule

    In-context/ prompt learning &
    \begin{tabitem}
      \item Prompt-based task adaptation
    \end{tabitem} &
    \begin{tabitem}
      \item Frozen-model inference
    \end{tabitem} &
    \begin{tabitem}
      \item Multi-modal instructions
      \item Observation-action pairs
    \end{tabitem} &
    \begin{tabitem}
      \item Novel-setting generalization
      \item Task planning
      \item Multi-modal flexibility
    \end{tabitem} &
    \begin{tabitem}
      \item Prompt sensitivity
      \item Limited grounding
      \item Inference latency
    \end{tabitem} &
    \begin{tabitem}
      \item SayCan \citep{saycan}, VIMA \citep{jiang2022vima}, ICRT \citep{fu2025icrt}, Instruct2Act \citep{huang2023instruct2act}
    \end{tabitem} \\
    \midrule

    World model learning &
    \begin{tabitem}
      \item Environment dynamics prediction
    \end{tabitem} &
    \begin{tabitem}
      \item Latent transition functions
    \end{tabitem} &
    \begin{tabitem}
      \item Interaction data
    \end{tabitem} &
    \begin{tabitem}
      \item Imagined experiences
      \item Physics modeling
      \item Reduced delays
    \end{tabitem} &
    \begin{tabitem}
      \item Hallucinations
      \item Cumulative errors
      \item High compute cost
    \end{tabitem} &
    \begin{tabitem}
      \item Dreamer \citep{hafner2019dream}, RoboDreamer \citep{zhou2024robodreamer}, ACT-JEPA \citep{vujinovic2025act}, AdaWorld \citep{gao2025adaworld}
    \end{tabitem} \\
    \midrule

    Generative learning &
    \begin{tabitem}
      \item Data/plan/ trajectory synthesis
    \end{tabitem} &
    \begin{tabitem}
      \item Data distribution modeling
    \end{tabitem} &
    \begin{tabitem}
      \item Massive multi-modal datasets
    \end{tabitem} &
    \begin{tabitem}
      \item Zero-shot generalization
      \item Multi-modality handling
      \item Long-horizon planning
    \end{tabitem} &
    \begin{tabitem}
      \item Sim-to-real gap
      \item Hallucinations
      \item Inference latency
    \end{tabitem} &
    \begin{tabitem}
      \item Diffuser \citep{janner2022planning}, Diffusion Policy \citep{chi2025diffusion}, DALL-E-Bot \citep{kapelyukh2023dall}, Gen2Sim \citep{katara2024gen2sim}
    \end{tabitem} \\
    \bottomrule
  \end{tabular}%
  }
\end{table}

\subsection{Comparative analysis and key insights}
\label{sec:learning_paradigms_discussion}

Having discussed in detail the various types of learning paradigms (Sections \ref{ssec:paradigms_supervised}-\ref{ssec:paradigms_generative-learning}), this section systematically examines the literature methods, providing a comparative analysis and critical insights for each type. In this respect, Table \ref{tab:learning_paradigms_summary} summarizes for each learning paradigm type its: a) Primary function, b) Main mechanisms, c) Primary data source, d) Key strengths, e) Critical limitations, and f) Indicative models.

Based on Table \ref{tab:learning_paradigms_summary}, the following main observations can be drawn: a) Supervised and self-supervised learning build general-purpose representations, b) Fine-tuning and domain adaptation ground generalist models to specific hardware and tasks, at the risk of catastrophic forgetting and negative transfer, c) Imitation learning offers easy, reward-free supervision from demonstrations, yet suffers covariate shift and data-quality dependence, d) Reinforcement learning enables self-improvement beyond demonstrations, but is sample-inefficient and reward-engineering intensive, and e) In-context, world-model, and generative learning support continual adaptation and future-state reasoning, though they face prompt sensitivity, hallucinations, and inference latency.

The full details of the learning paradigms discussed in this section are provided in Appendix~\ref{sec:learning_paradigms_details}. In particular, for each learning paradigm (i.e., supervised learning, self-supervised learning, fine-tuning, domain adaptation, imitation learning, reinforcement learning, in-context/prompt learning, world model learning, and generative learning), the appendix contains the complete category definition, the corresponding main advantages and limitations, as well as the extensive list of representative methods per sub-category.

\section{Learning stages}
\label{sec:learningStage}

During the overall learning process of a robotic FM, the particular phase at which knowledge is incorporated largely defines the type/nature of the acquired skills, algorithmic/development details, and key assumptions about the model behavior. In this context, the main learning stages identified in the literature are: a) Pre-training, b) Offline fine-tuning, c) Online adaptation, and d) Continuous learning, as discussed in Section \ref{sec:Taxonomy} and further detailed below.

\subsection{Pre-training}
\label{ssec:stages_pre-training}

The pre-training stage estimates robust, general-purpose representations of robotic data by processing massive (often internet-scale) amounts of diverse data from multiple platforms, modeling the cross-correlations among vision, language, and action \citep{li2024foundation}. This offers increased generalization, robust zero-shot capability, and accurate multi-modal mapping across words, visual concepts, and physical actions \citep{kawaharazuka2025vision}.

The most common learning paradigms adopted during the pre-training stage are:

\begin{itemize}

  \item \underline{Supervised learning}: Supervised learning equips a model with a foundational understanding of the world from high-quality, diverse, labeled data. PaLM-E \citep{palme} is jointly trained on web-scale multi-modal data and embodied experiences, while RT-$2$ \citep{rt2} is constructed using web and robot manipulation data.

  \item \underline{Self-supervised learning}: SSL enables robots to operate beyond narrow, task-specific programming, towards generalized intelligence capabilities. DINOv$2$ \citep{oquab2023dinov2} and VideoMAE \citep{tong2022videomae} learn perception priors from unlabeled data, while R$3$M \citep{nair2022r3m} extends this by incorporating egocentric features.

  \item \underline{Imitation learning}: IL learns a prior distribution of successful behaviors directly from expert demonstrations. RT-$1$ \citep{rt1} treats robot control as next-token prediction over multi-modal streams to inherit semantic and sensorimotor skills, while Octo \citep{team2024octo} uses the Open-X-Embodiment trajectories to derive a generalist policy.

\end{itemize}

\subsection{Offline fine-tuning}
\label{ssec:stages_fine-tuning}

Offline fine-tuning bridges the knowledge gap between the general-purpose representations learned during pre-training and the specificities of a given physical-world application, targeting task and embodiment specialization \citep{hu2023toward}. This reduces the need for training data, increases training stability, and enables knowledge distillation of only the necessary general-purpose representations \citep{firoozi2025foundation}.

The most common learning paradigms adopted during the offline fine-tuning stage are:
\begin{itemize}

  \item \underline{Imitation learning}: IL equips pre-trained models with the low-level precision skills for a specific application, learning specific motor commands by mimicking expert demonstrations. Octo \citep{team2024octo} and OpenVLA \citep{openvla} employ large-scale Open-X-Embodiment pretraining and then focus on new platforms using LoRA-style adapters, while LiteVLA \citep{williams2025lite} shows NF4 quantized LoRA can be tuned on CPU-only hardware.

  \item \underline{Reinforcement learning}: RL enables robots to learn from a reward signal, focusing on actions that lead to successful task executions. A recurrent tracker is trained with conservative offline RL on VFM-annotated trajectories \citep{zhong2024empowering}, while FLaRe \citep{hu2025flare} applies large-scale RL fine-tuning to transformer-based policies for long-horizon mobile manipulation.

  \item \underline{Generative learning}: GL specializes pre-trained models to specific environments, especially under sparse target-task constraints. ReBot \citep{fang2025rebot} replays real trajectories in simulation and composes them into inpainted real backgrounds to adapt to new domains, while RoboDreamer \citep{zhou2024robodreamer} uses compositional WMs to generate imagined video plans as additional training data.

\end{itemize}

\subsection{Online adaptation}
\label{ssec:stages_adaptation}

Online adaptation equips robots with routines for learning in real-time from their own experiences, handling the distribution shift between offline training data and what is encountered during online deployment \citep{firoozi2025foundation}. This offers high-precision performance in the adapted environments, continuous improvement, and robustness to distribution shifts \citep{yuan2025survey}.

The most common learning paradigms adopted during the online adaptation stage are:
\begin{itemize}

    \item \underline{Domain adaptation}: This handles the physical-world constraints and sensorial noise of a specific deployment scenario, re-calibrating the model's knowledge structures to the perceived environment. TTA-Nav \citep{piriyajitakonkij2024tta} adds a reconstruction decoder on a pre-trained policy to denoise corrupted frames without gradient updates, while Phys2Real \citep{wang2025phys2real} bridges sim-to-real gaps by combining FM priors with interaction-based estimations.

    \item \underline{Reinforcement learning}: RL allows the robot to perform micro-adjustments to its general-purpose knowledge, based on sensory feedback and exploration. Self-improving embodied FMs refine pre-trained policies from reward and success estimation across a robot fleet \citep{ghasemipour2025self}, while RL-VLM-F \citep{wang2024rl} estimates rewards using a VLM that compares trajectory snippets with language goals.

    \item \underline{In-context/prompt learning}: This enables zero- or few-shot specialization at deployment time, without updating model weights. ICRT \citep{fu2025icrt} uses in-context imitation policies conditioned on a few recent demonstration trajectories, while LLM-based control stacks interleave reasoning and acting through ReAct-style prompting to monitor progress and to revise plans \citep{yao2023react}.

\end{itemize}

\subsection{Continuous learning}
\label{ssec:stages_continuous_learning}

Continuous Learning (CL), often termed lifelong learning, enables robots to acquire new skills or to adapt to new environments incrementally, without full retraining from scratch, employing conventional paradigms (e.g., IL, RL) in slower outer loops \citep{xiao2025robot}. This offers increased adaptability for long-term deployment, improved scalability across a fleet, and reduced downtime \citep{firoozi2025foundation}.

The most common learning paradigms adopted during the continuous learning stage are:

\begin{itemize}

  \item \underline{Domain adaptation}: DA enables a model to continuously adjust its internal knowledge to dynamic, real-world settings. Action Flow Matching for Lifelong Learning \citep{murillo2025actionflowmatching} incrementally aligns robot dynamics across sequential tasks for safe continual adaptation, while VR-Robo \citep{zhu2025vr} builds photorealistic digital twins from logged data to retrain and transfer navigation or locomotion policies.

  \item \underline{Imitation learning}: IL constantly bridges the gap between a model's general knowledge and the high-precision requirements of a real-world case, given few expert demonstrations. SkillsCrafter \citep{wang2026lifelong} realizes lifelong language-conditioned learning across sequential manipulation skills while reducing forgetting via symbolic skill distillation, while LOTUS \citep{wan2024lotus} refines manipulation skills from demonstration streams.


 \item \underline{Reinforcement learning}: RL supports continuous learning by enabling policies to improve through repeated interaction, autonomous practice, and reward-driven post-training over extended deployment horizons. Self-improving embodied FMs refine pretrained policies via autonomous practice from self-predicted rewards across a robot fleet \citep{ghasemipour2025self}, while LiReN \citep{stachowicz2024lifelong} shows navigation FMs can improve lifelong learning through online RL in open-world settings.

\end{itemize}

\begin{table}[!t]
  \caption{Learning stages: Comparative analysis and key insights.}
  \label{tab:learning_stages_summary}
  \centering
  \scriptsize
  
  \setlength{\aboverulesep}{0pt}
  \setlength{\belowrulesep}{0pt}
  \setlength{\tabcolsep}{2pt}
  \renewcommand{\arraystretch}{0.9}

  \setlist*[tabitem]{before=\vspace{2.2pt}\justifying, after=\vspace{2.2pt}}

  \rowcolors{2}{gray!25}{white}

  \newlength{\Wstage}\setlength{\Wstage}{1.8cm}
  \newlength{\Wpre}\setlength{\Wpre}{3.7cm}
  \newlength{\Woff}\setlength{\Woff}{3.7cm}
  \newlength{\Won}\setlength{\Won}{3.7cm}
  \newlength{\Wcont}\setlength{\Wcont}{3.7cm}

  \resizebox{\textwidth}{!}{%
  \begin{tabular}{@{}|
    >{\justifying\arraybackslash}m{\Wstage}|
    >{\raggedright\arraybackslash}m{\Wpre}|
    >{\raggedright\arraybackslash}m{\Woff}|
    >{\raggedright\arraybackslash}m{\Won}|
    >{\raggedright\arraybackslash}m{\Wcont}|
  @{}}
    \toprule
    \rowcolor{gray!40}
    \headerbreak{Learning\\stage} &
    \headerbreak{Pre-training} &
    \headerbreak{Offline\\fine-tuning} &
    \headerbreak{Online\\adaptation} &
    \headerbreak{Continuous\\learning} \\
    \midrule

    Primary function &
    \begin{tabitem}
      \item World-knowledge learning
    \end{tabitem} &
    \begin{tabitem}
      \item Task/embodiment grounding
    \end{tabitem} &
    \begin{tabitem}
      \item Real-time adjustment
    \end{tabitem} &
    \begin{tabitem}
      \item Lifelong retention
    \end{tabitem} \\
    \midrule

    Data requirements &
    \begin{tabitem}
      \item Internet-scale data
    \end{tabitem} &
    \begin{tabitem}
      \item Expert demonstrations
    \end{tabitem} &
    \begin{tabitem}
      \item Interaction history
    \end{tabitem} &
    \begin{tabitem}
      \item Sequential streams
    \end{tabitem} \\
    \midrule

    Learning paradigms &
    \begin{tabitem}
      \item Supervised learning
      \item Self-supervised learning
      \item Imitation learning
    \end{tabitem} &
    \begin{tabitem}
      \item Imitation learning
      \item Reinforcement learning
      \item Generative learning
    \end{tabitem} &
    \begin{tabitem}
      \item Domain adaptation
      \item Reinforcement learning
      \item In-context/prompt learning
    \end{tabitem} &
    \begin{tabitem}
      \item Domain adaptation
      \item Imitation learning
      \item Reinforcement learning
    \end{tabitem} \\
    \midrule

    Computational requirements &
    \begin{tabitem}
      \item Extreme GPU
    \end{tabitem} &
    \begin{tabitem}
      \item Moderate-to-high GPU
    \end{tabitem} &
    \begin{tabitem}
      \item Low (edge)
    \end{tabitem} &
    \begin{tabitem}
      \item Moderate (incremental)
    \end{tabitem} \\
    \midrule

    Generalization capability &
    \begin{tabitem}
      \item Zero-shot cross-domain
    \end{tabitem} &
    \begin{tabitem}
      \item Task-specific precision
    \end{tabitem} &
    \begin{tabitem}
      \item Local adjustment
    \end{tabitem} &
    \begin{tabitem}
      \item Cross-task transfer
    \end{tabitem} \\
    \midrule

    Strengths &
    \begin{tabitem}
      \item Generalization
      \item Zero-shot capability
      \item Multi-modal mapping
    \end{tabitem} &
    \begin{tabitem}
      \item Data efficiency
      \item Training stability
      \item Knowledge distillation
    \end{tabitem} &
    \begin{tabitem}
      \item High precision
      \item Continuous improvement
      \item Shift robustness
    \end{tabitem} &
    \begin{tabitem}
      \item Adaptability
      \item Scalability
      \item Reduced downtime
    \end{tabitem} \\
    \midrule

    Limitations &
    \begin{tabitem}
      \item High compute cost
      \item Weak real-world performance
      \item Safety concerns
    \end{tabitem} &
    \begin{tabitem}
      \item Distribution shift
      \item Data-quality dependence
      \item No online exploration
    \end{tabitem} &
    \begin{tabitem}
      \item Catastrophic forgetting
      \item Latency
      \item Noise sensitivity
    \end{tabitem} &
    \begin{tabitem}
      \item Catastrophic forgetting
      \item Stability-plasticity dilemma
      \item Memory overhead
    \end{tabitem} \\
    \midrule

    Indicative models &
    \begin{tabitem}
      \item Open X-Embodiment/RT-X \citep{openx}, GR00T N1 \citep{bjorck2025gr00t}, PaLM-E \citep{palme}, CLIP \citep{clip}
    \end{tabitem} &
    \begin{tabitem}
      \item Octo \citep{team2024octo}, OpenVLA \citep{openvla}, BC-Z \citep{jang2022bc}, RT-2 \citep{rt2}
    \end{tabitem} &
    \begin{tabitem}
      \item TTA-Nav \citep{piriyajitakonkij2024tta}, Phys2Real \citep{wang2025phys2real}, RL-VLM-F \citep{wang2024rl}, ICRT \citep{fu2025icrt}
    \end{tabitem} &
    \begin{tabitem}
      \item DrEureka \citep{ma2024dreureka}, VR-Robo \citep{zhu2025vr}, LOTUS \citep{wan2024lotus}, Self-Improving Embodied FMs \citep{ghasemipour2025self}
    \end{tabitem} \\
    \bottomrule
  \end{tabular}%
  }
\end{table}

\subsection{Comparative analysis and key insights}
\label{sec:learning_stages_discussion}

Having discussed in detail the different learning stages (Sections \ref{ssec:stages_pre-training}-\ref{ssec:stages_continuous_learning}), this section systematically examines the literature methods, providing a comparative analysis and critical insights for each stage. In this respect, Table \ref{tab:learning_stages_summary} summarizes for each learning stage its: a) Primary function, b) Data requirements, c) Learning paradigms, d) Computational requirements, e) Generalization capability, f) Key strengths, g) Critical limitations, and h) Indicative models.

Based on Table \ref{tab:learning_stages_summary}, the following main observations can be drawn: a) Pre-training builds general-purpose, zero-shot world knowledge from internet-scale data, but at extreme computational cost and with limited real-world grounding, b) Offline fine-tuning grounds these representations to specific tasks and embodiments with moderate data, yet lacks online exploration and it is sensitive to distribution shift, c) Online adaptation adjusts policies in real-time on edge hardware, trading low compute for sensitivity to noise and catastrophic forgetting, d) Continuous learning enables lifelong skill retention and fleet-level scalability, but faces the stability-plasticity dilemma and memory overhead, and e) Overall, the stages trade off general-purpose knowledge acquisition against specialized, real-time precision refinement.

The full details of the learning stages discussed in this section are provided in Appendix~\ref{sec:learning_stages_details}. In particular, for each learning stage (i.e., pre-training, offline fine-tuning, online adaptation, and continuous learning), the appendix contains the complete category definition, the corresponding main advantages and limitations, as well as the extensive list of representative methods per adopted learning paradigm.

\section{Robotic tasks}
\label{sec:tasks}

The introduction of FMs has led to transformative effects in the materialization and execution of all core robotic tasks, mainly by shifting the field from task-specific programming to general-purpose, multi-task agents. In this context, the main robotic tasks identified in the literature, where FM-based solutions have been applied, are: a) Perception, b) Planning, c) Navigation, d) Manipulation, and e) Human-robot interaction, as discussed in Section \ref{sec:Taxonomy} and further detailed below.

\subsection{Perception}
\label{ssec:robotic_tasks-perception}

Perception creates rich, semantic maps of the surrounding environment that enable robots to execute individual actions, realizing semantic grounding, object-affordance discovery, and contextual awareness \citep{kawaharazuka2024real}. The incorporation of FMs offers open-vocabulary recognition, zero-shot generalization, multi-modal fusion, and robustness to noise \citep{hu2023toward}.

The main categories of perception methods are:
\begin{itemize}
  \item \underline{Language-grounded detection and segmentation}: This identifies (detection) and precisely outlines (segmentation) objects in the robot's environment based on natural language prompts. Grounding DINO \citep{liu2024grounding} provides phrase-based detections robust to clutter and in the zero-shot setting, while SAM-style promptable segmenters \citep{sam} incorporate box, click, or text prompts into control pipelines interactively.

  \item \underline{Open-vocabulary $3$D semantic mapping}: This allows robots to perceive and localize objects of previously unseen categories in $3$D space using natural language inputs. ConceptFusion \citep{jatavallabhulaconceptfusion} builds open-set, language-searchable maps supporting multi-modal queries, while ConceptGraphs \citep{gu2024conceptgraphs} estimates object nodes and relations so that planners can operate on a semantic scene graph instead of raw pixels.

  \item \underline{Pose estimation and affordance prediction}: Pose estimation aligns an object's local coordinate system to the world frame, while affordance prediction detects the ways an object can be manipulated, with FMs linking semantics to spatial geometry. OV$9$D \citep{cai2024ov9d} estimates category-agnostic $9$-DoF pose without CAD models, while OpenAD \citep{nguyen2023open} models zero-shot $3$D affordances in a shared vision-language embedding space.

  \item \underline{Contact-centric and visuotactile perception}: This analyzes and models the physics of robot interactions, with contact-centric approaches, using junction points as state and visuotactile ones integrating vision with tactile sensing. Tactile-VLA \citep{huang2025tactile} combines tactile streams with vision and language for insertion and assembly, while NeuralFeels \citep{suresh2024neuralfeels} enhances in-hand pose and shape estimation, when visual cues are uncertain.

  \item \underline{Long-term object tracking}: This locates a given object or point across video frames, maintaining stability over time and recalling objects after occlusion, as required for long-horizon tasks. OVTrack \citep{li2023ovtrack} employs language and diffusion priors to generalize multi-object tracking to unseen categories, while DINO-MOT \citep{lee2024dino} combines DINOv$2$ features with a memory mechanism for robust pedestrian tracking.
\end{itemize}

\subsection{Planning}
\label{ssec:robotic_tasks-planning}

Planning serves as the fundamental bridge between high-level semantic reasoning and low-level motor control, with its primary usefulness lying in long-horizon task decomposition of a goal into sequential sub-goals prior to execution \citep{hu2023toward}. The incorporation of FMs offers increased interpretability, generalization, reduced need for training data, and safety-constraint integration into the planning loop \citep{firoozi2025foundation}.

The main categories of planning methods are:
\begin{itemize}
  \item \underline{Language-driven task decomposition}: This breaks down high-level, long-horizon goals into logical sequences of primitive actions, often as structured, executable code. SayCan \citep{saycan} grounds each action step on an affordance map so that abstract sub-goals map to concrete objects, while Code-as-Policies \citep{liang2023code} generates short Python-like programs that render planning easier to inspect, to test, and to modify.

  \item \underline{Neuro-symbolic closed-loop reasoning}: This combines the general-purpose knowledge of a FM with formal, logical checking to guarantee successful real-world operation. ISR-LLM \citep{zhou2024isr} converts instructions to PDDL and iteratively refines plans via symbolic validation, while AutoTAMP \citep{chen2024autotamp} translates instructions into TAMP representations using autoregressive re-prompting to correct errors.

  \item \underline{Multi-modal policy generation}: This generates high-level plans or actions, by integrating language, vision, and embodied state. PaLM-E \citep{palme} combines language, vision, and proprioception for embodied reasoning, while RT-$2$ \citep{rt2} shows that language-aligned visual representations transfer web-scale semantic knowledge to real-world control.

  \item \underline{Execution-time validation and failure recovery}: This monitors plans during deployment, verifying preconditions, detecting failures, and triggering corrective re-planning. Code-as-Monitor \citep{zhou2025code} introduces constraint-aware visual programming for reactive and proactive failure detection, while VLM-based monitoring frameworks, such as Guardian \citep{pacaud2025guardian}, support execution-time failure recovery in manipulation.

  \item \underline{Semantic multi-robot coordination}: This capitalizes on the broad knowledge of FMs to coordinate multi-robot setups, reasoning about task dependencies, resources, and scheduling. LiP-LLM \citep{obata2024lip} builds a skill list and dependency graph, and uses linear programming to allocate tasks, while SMART-LLM \citep{kannan2024smart} assigns role-based tasks from a single high-level instruction.
\end{itemize}

\subsection{Navigation}
\label{ssec:robotic_tasks-navigation}

The usefulness of FMs in robot navigation lies in providing the necessary spatial common-sense knowledge in embodied AI settings, by processing the environment as a high-level semantic space instead of rigid spatial maps \citep{pan2025mixed}. The incorporation of FMs offers open-world navigation, cross-embodiment transfer, and semantic reasoning that combines vision with language for instruction interpretation \citep{firoozi2025foundation}.

The main categories of navigation methods are:
\begin{itemize}

  \item \underline{Semantic spatial grounding}: This registers the objects in the environment, their relative position, and a concrete action plan for reaching them in natural-language form. VLMaps \citep{huang2023visual} builds CLIP-indexed spatial memories that query objects and rooms without task-specific retraining, while open-vocabulary mapping methods, such as One Map to Find Them All \citep{busch2025one}, support multi-object navigation, dynamic environments, and functional queries.

  \item \underline{Instruction-following policies}: This leverages the capability of FMs to interpret natural language instructions without a pre-defined map or hard-coded scripts. Generalist navigation models, such as ViNT \citep{shah2023vint}, formalize navigation as sequence prediction over images and poses across robots, while NaviLLM \citep{zheng2024towards} unifies instruction following and embodied QA with schema-tuned prompts.

  \item \underline{End-to-end policies}: This maps raw sensorial data directly to motor commands, instead of the conventional modular design of separate mapping, localization, and path planning. ViNT \citep{shah2023vint} learns a generalizable visuomotor navigation policy across robots and environments, while DriveGPT-4 \citep{xu2024drivegpt4} predicts low-level control signals directly from visual- and language-conditioned inputs.

\end{itemize}

\subsection{Manipulation}
\label{ssec:robotic_tasks-manipulation}

The utility of FMs in robot manipulation lies in translating high-level, semantic, human-like instructions into the precise forces and movements needed to manipulate an object, with cross-embodiment learning as their main contribution \citep{li2024foundation}. The incorporation of FMs offers increased generalization to unseen objects, improved robustness via real-time feedback, and enhanced understanding of objects' physical properties \citep{sapkota2025vision}.

The main categories of manipulation methods are:
\begin{itemize}

  \item \underline{Language-to-action models}: This interprets the semantic meaning of a natural language command and maps it to the physical world, without task-specific programming. RT-$2$ \citep{rt2} approaches manipulation as sequence modeling over multi-modal tokens, while OpenVLA \citep{openvla} is a large vision-language-action policy that adapts to new platforms via small-scale fine-tuning.

  \item \underline{Retrieval-augmented imitation learning}: This receives guidance from a large database of relevant previous demonstrations to predict future actions. DINOBot \citep{di2024dinobot} detects similar demonstrations, via DINO feature correspondence, to estimate dense trajectories for one- and few-shot generalization, while STRAP \citep{memmelstrap} retrieves sub-trajectories to augment few-shot imitation learning.

  \item \underline{Constraint-aware policy synthesis}: This employs a FM to generate high-level control code or objectives bounded by physical, safety, and environmental constraints. CoPa \citep{huang2024copa} detects task-relevant parts with a multi-modal LLM and estimates spatial constraints translated into $6$-DoF actions, while ReKep \citep{huang2025rekep} represents tasks as relational keypoint constraints optimized hierarchically for assembly.

  \item \underline{Semantic spatial maps}: This generates representations combining $3$D spatial geometry with semantic information about the objects. VoxPoser \citep{huang2023voxposer} estimates constraints and affordances from language, creates $3$D value maps, and applies zero-shot motion planning, while AdaRPG \citep{zhang2025adaptive} leverages VLMs to infer part affordances guiding articulated-object manipulation.
\end{itemize}

\subsection{Human-robot interaction}
\label{ssec:robotic_tasks-human-robot-interaction}

The usefulness of FMs in HRI is grounded on their ability for semantic, human-like reasoning and interpreting human intent, enabling robots to react to conversational instructions, to ask clarification questions, and to explain their actions \citep{zhao2025multimodal}. The incorporation of FMs offers rapid generalization, intuitive control, increased safety alignment, and efficient error recovery from user feedback \citep{xiao2025robot}.

The main categories of HRI methods are:
\begin{itemize}

  \item \underline{Conversational policy alignment}: This uses natural language dialogue to dynamically adjust a robot's behavior in real-time to match a human's intent, preferences, and safety boundaries. DRAGON \citep{liu2024dragon} is a dialogue-based navigation framework that grounds free-form commands in visual landmarks and asks clarification questions, while PlanCollabNL \citep{izquierdo2024plancollabnl} translates spoken instructions into editable collaborative plans.

  \item \underline{Reciprocal social tuning}: This develops a semantic communication framework where humans and robots continuously adjust their behaviors, social cues, and expectations to harmonize. TidyBot \citep{wu2023tidybot} learns user-specific clean-up preferences from a few examples and generalizes them to new scenes, while LAMS \citep{tao2025lams} extends this to assistive teleoperation, by switching control modes from user feedback.

  \item \underline{Active alignment and mitigation}: This keeps a robot synchronized with human intent, while proactively handling errors or deviations. RoboVQA \citep{sermanet2024robovqa} queries egocentric video to check preconditions and to request human intervention on failures, while embodied LLM controllers incorporate human feedback to adjust plans when the state drifts from the goal \citep{barmann2024incremental}.

\end{itemize}

\begin{table}[!t]
  \caption{Robotic tasks: Comparative analysis and key insights.}
  \label{tab:robotic_tasks_summary}
  \centering
  \scriptsize
  
  \setlength{\aboverulesep}{0pt}
  \setlength{\belowrulesep}{0pt}
  \setlength{\tabcolsep}{2pt}
  \renewcommand{\arraystretch}{0.9}

  \setlist*[tabitem]{before=\vspace{2.2pt}\justifying, after=\vspace{2.2pt}}

  \rowcolors{2}{gray!25}{white}

  \newlength{\Wtask}\setlength{\Wtask}{1.3cm}
  \newlength{\Wperc}\setlength{\Wperc}{3.15cm}
  \newlength{\Wplan}\setlength{\Wplan}{3.15cm}
  \newlength{\Wnav}\setlength{\Wnav}{3.15cm}
  \newlength{\Wman}\setlength{\Wman}{3.15cm}
  \newlength{\Whri}\setlength{\Whri}{3.55cm}

  \resizebox{\textwidth}{!}{%
  \begin{tabular}{@{}|
    >{\justifying\arraybackslash}m{\Wtask}|
    >{\raggedright\arraybackslash}m{\Wperc}|
    >{\raggedright\arraybackslash}m{\Wplan}|
    >{\raggedright\arraybackslash}m{\Wnav}|
    >{\raggedright\arraybackslash}m{\Wman}|
    >{\raggedright\arraybackslash}m{\Whri}|
  @{}}
    \toprule
    \rowcolor{gray!40}
    \headerbreak{Robotic\\task} &
    \headerbreak{Perception} &
    \headerbreak{Planning} &
    \headerbreak{Navigation} &
    \headerbreak{Manipulation} &
    \headerbreak{Human-robot\\interaction} \\
    \midrule

    Primary function &
    \begin{tabitem}
      \item Semantic scene recognition
      \item Object detection
      \item 3D spatial grounding
    \end{tabitem} &
    \begin{tabitem}
      \item Goal decomposition
      \item Long-horizon reasoning
      \item Constraint-aware sequencing
    \end{tabitem} &
    \begin{tabitem}
      \item Goal-conditioned path-finding
      \item Obstacle avoidance
      \item Topological exploration
    \end{tabitem} &
    \begin{tabitem}
      \item Fine-motor control
      \item Grasping actions
      \item Object relocation
    \end{tabitem} &
    \begin{tabitem}
      \item Context-aware communication
      \item Social collaboration
      \item Command interpretation
    \end{tabitem} \\
    \midrule

    FM type &
    \begin{tabitem}
      \item VFMs
      \item VLMs
    \end{tabitem} &
    \begin{tabitem}
      \item LLMs
      \item VLMs
    \end{tabitem} &
    \begin{tabitem}
      \item LLMs
      \item VFMs
      \item VLMs
      \item VLAs
    \end{tabitem} &
    \begin{tabitem}
      \item VLMs
      \item VLAs
    \end{tabitem} &
    \begin{tabitem}
      \item LLMs
      \item VLMs
      \item VLAs
    \end{tabitem} \\
    \midrule

    Input &
    \begin{tabitem}
      \item RGB-D images
      \item 3D point clouds
      \item Textual prompts
    \end{tabitem} &
    \begin{tabitem}
      \item Natural-language goals
      \item Scene graphs
      \item Symbolic state
      \item Execution feedback
    \end{tabitem} &
    \begin{tabitem}
      \item RGB images
      \item Spatial maps
      \item Spatial memory
      \item Odometry measurements
      \item Language instructions
    \end{tabitem} &
    \begin{tabitem}
      \item Natural-language goals
      \item Proprioception/gripper state
      \item Video demonstrations
      \item Scene descriptions
      \item Tactile/force data
    \end{tabitem} &
    \begin{tabitem}
      \item Human speech
      \item Human gestures
      \item Interaction history
      \item Social context
    \end{tabitem} \\
    \midrule

    Output &
    \begin{tabitem}
      \item Segmentation masks
      \item Scene graphs
      \item Semantic labels
      \item 3D maps
      \item Poses/affordances
    \end{tabitem} &
    \begin{tabitem}
      \item Plans/sub-goals/code
      \item Constraints/precon-ditions
      \item Task decompositions
    \end{tabitem} &
    \begin{tabitem}
      \item Velocity commands
      \item Spatial trajectories
      \item Target locations
      \item Route descriptions
    \end{tabitem} &
    \begin{tabitem}
      \item Action sequences
      \item Grasps/placements
      \item 6-DOF poses
      \item Gripper states
      \item Adapted policies
    \end{tabitem} &
    \begin{tabitem}
      \item Speech output
      \item Socially-aware motion
      \item Goal corrections
      \item Explanations/reports
      \item User preferences
    \end{tabitem} \\
    \midrule

    Strengths &
    \begin{tabitem}
      \item Open-vocabulary recognition
      \item Zero-shot generalization
      \item Multi-modal fusion
      \item Noise robustness
    \end{tabitem} &
    \begin{tabitem}
      \item Interpretability
      \item Generalization
      \item Data efficiency
      \item Safety-constraint integration
    \end{tabitem} &
    \begin{tabitem}
      \item Open-world navigation
      \item Cross-embodiment transfer
      \item Semantic reasoning
    \end{tabitem} &
    \begin{tabitem}
      \item Generalization
      \item Robustness
      \item Embodiment capability
    \end{tabitem} &
    \begin{tabitem}
      \item Rapid generalization
      \item Intuitive control
      \item Safety alignment
      \item Error recovery
    \end{tabitem} \\
    \midrule

    Limita-tions &
    \begin{tabitem}
      \item High latency
      \item Low explainability
      \item Hallucinations
      \item Spatial imprecision
    \end{tabitem} &
    \begin{tabitem}
      \item Logical gaps
      \item Limited grounding
      \item High latency
      \item Closed-loop complexity
    \end{tabitem} &
    \begin{tabitem}
      \item Sim-to-real gap
      \item Data scarcity
      \item High latency
      \item Safety/ethical concerns
    \end{tabitem} &
    \begin{tabitem}
      \item Data scarcity
      \item Safety concerns
      \item Low action precision
    \end{tabitem} &
    \begin{tabitem}
      \item Data scarcity
      \item Human bias
      \item Semantic drift
    \end{tabitem} \\
    \midrule

    Indicative models &
    \begin{tabitem}
      \item CLIP \citep{clip}, SAM \citep{sam}, Grounding DINO \citep{liu2024grounding}, DINOv2 \citep{oquab2023dinov2}, ConceptGraphs \citep{gu2024conceptgraphs}
    \end{tabitem} &
    \begin{tabitem}
      \item SayCan \citep{saycan}, Code-as-Policies \citep{liang2023code}, SayPlan \citep{rana2023sayplan}, AutoTAMP \citep{chen2024autotamp}
    \end{tabitem} &
    \begin{tabitem}
      \item LM-Nav \citep{shah2023lm}, GNM \citep{shah2023gnm}, ViNT \citep{shah2023vint}, DRAGON \citep{liu2024dragon}
    \end{tabitem} &
    \begin{tabitem}
      \item RT-1 \citep{rt1}, OpenVLA \citep{openvla}, GR00T N1 \citep{bjorck2025gr00t}, Diffusion Policy \citep{chi2025diffusion}
    \end{tabitem} &
    \begin{tabitem}
      \item PaLM-E \citep{palme}, Gemini Robotics \citep{team2025gemini}, RoboVQA \citep{sermanet2024robovqa}, TidyBot \citep{wu2023tidybot}, LAMS \citep{tao2025lams}
    \end{tabitem} \\
    \bottomrule
  \end{tabular}%
  }
\end{table}

\begin{figure}[!t]
    \centering

    \begin{minipage}[t]{0.46\textwidth}
        \centering
        \includegraphics[width=\linewidth]{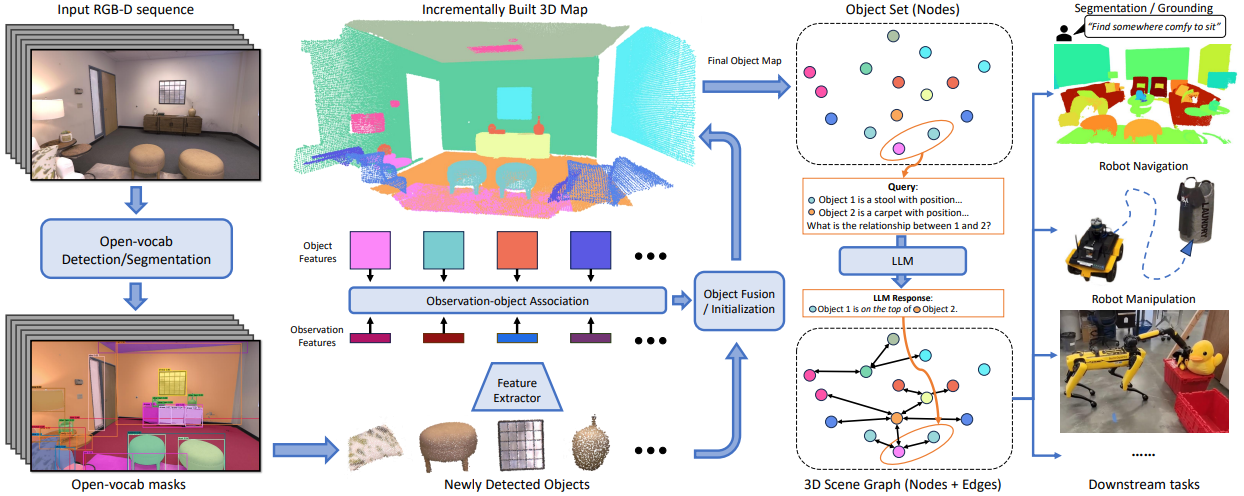}\\[0.3em]
        (a)

        \vspace{1.0em}

        \includegraphics[width=\linewidth]{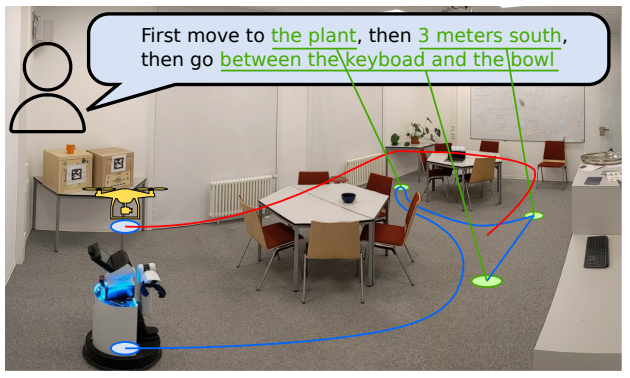}\\[0.3em]
        (c)

        \vspace{1.0em}

        \includegraphics[width=\linewidth]{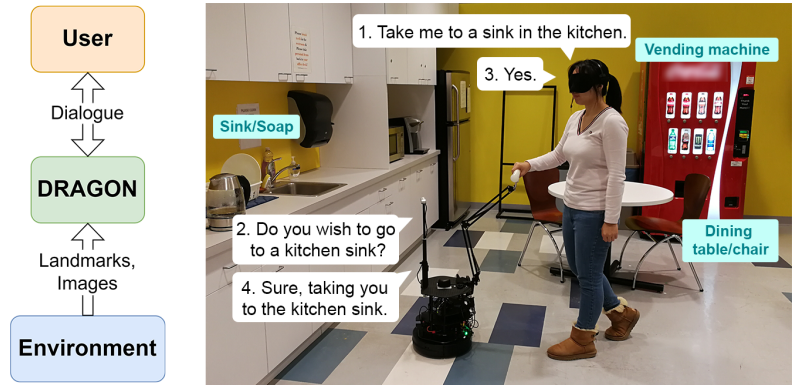}\\[0.3em]
        (e)
    \end{minipage}
    \hspace{0.04\textwidth}
    \begin{minipage}[t]{0.46\textwidth}
        \vspace*{-2.5cm}
        \centering
        \includegraphics[width=\linewidth]{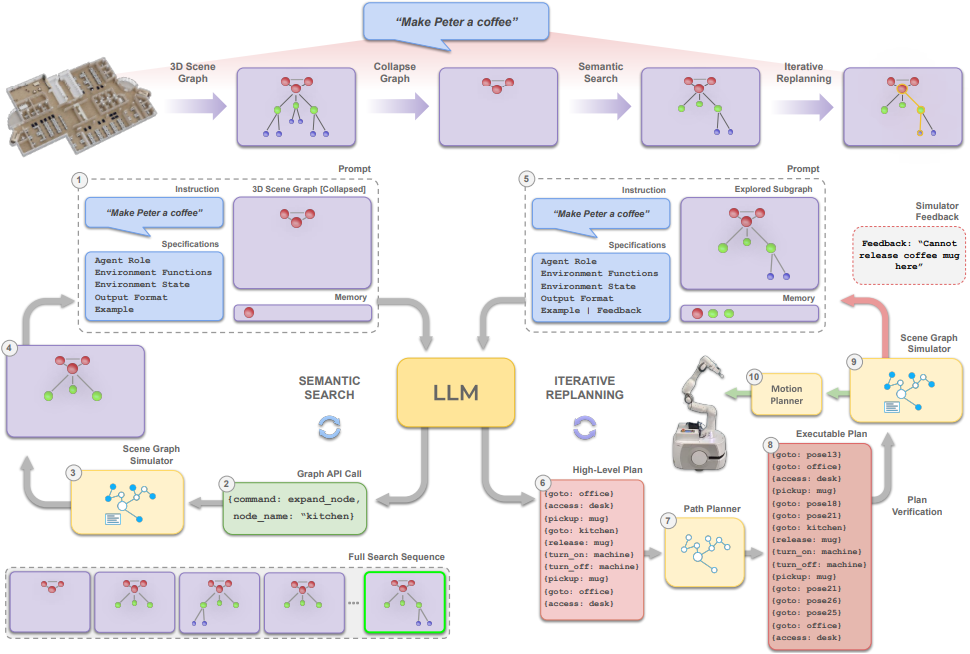}\\[1.5em]
        (b)

        \vspace{1.5em}

        \includegraphics[width=\linewidth]{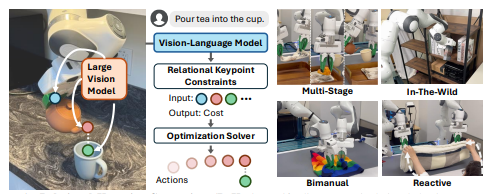}\\[0.3em]
        (d)
    \end{minipage}

    \caption{Representative literature methods per robotic task: (a) Perception (ConceptGraphs \citep{gu2024conceptgraphs}), (b) Planning (SayPlan \citep{rana2023sayplan}), (c) Navigation (VLMaps \citep{huang2023visual}), (d) Manipulation (ReKep \citep{huang2025rekep}), and (e) Human-robot interaction (DRAGON \citep{liu2024dragon}).}
    \label{fig:robotic_tasks_examples}
\end{figure}

\subsection{Comparative analysis and key insights}
\label{ssec:robot-tasks_discussion}

Having discussed in detail the different robotic tasks (Sections \ref{ssec:robotic_tasks-perception}-\ref{ssec:robotic_tasks-human-robot-interaction}), this section systematically examines the literature methods, providing a comparative analysis and critical insights for each task. In this respect, Table \ref{tab:robotic_tasks_summary} summarizes for each robotic task its: a) Primary function, b) FM type, c) Input types, d) Output types, e) Key strengths, f) Critical limitations, and g) Indicative models. Moreover, representative literature methods per robotic task are illustrated in Fig. \ref{fig:robotic_tasks_examples}.

Based on Table \ref{tab:robotic_tasks_summary}, the following main observations can be drawn: a) Across all tasks, a fundamental transition occurs from isolated, task-specific modules to integrated, generalist architectures leveraging internet-scale pretraining, b) Perception achieves robust open-vocabulary recognition and semantic fluency, but suffers spatial imprecision and hallucinations, c) Planning attains interpretable long-horizon reasoning, yet depends on grounding to keep plans physically feasible, d) Navigation and manipulation emphasize embodiment, facing a trade-off between semantic-reasoning latency and real-time motor control, and e) HRI enables intuitive communication, though hindered by performance-latency trade-offs and the need to adhere to safety and social norms.

The full details of the robotic tasks discussed in this section are provided in Appendix~\ref{sec:robotic_tasks_details}. In particular, for each robotic task (i.e., perception, planning, navigation, manipulation, and human-robot interaction), the appendix contains the complete category definition, the corresponding main advantages and limitations, as well as the extensive list of representative methods per sub-category.

\section{Application domains}
\label{sec:app_domains}

\begin{figure}[!t]
    \centering

    \begin{minipage}[t]{0.81\textwidth}
        \centering
        \includegraphics[width=\linewidth,height=0.16\textheight,keepaspectratio]{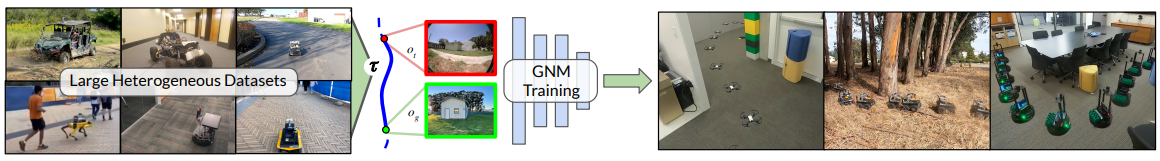}\\[0.2em]
        (a)
    \end{minipage}

    \vspace{0.3em}

    \begin{minipage}[t]{0.39\textwidth}
        \centering
        \includegraphics[width=\linewidth,height=0.15\textheight,keepaspectratio]{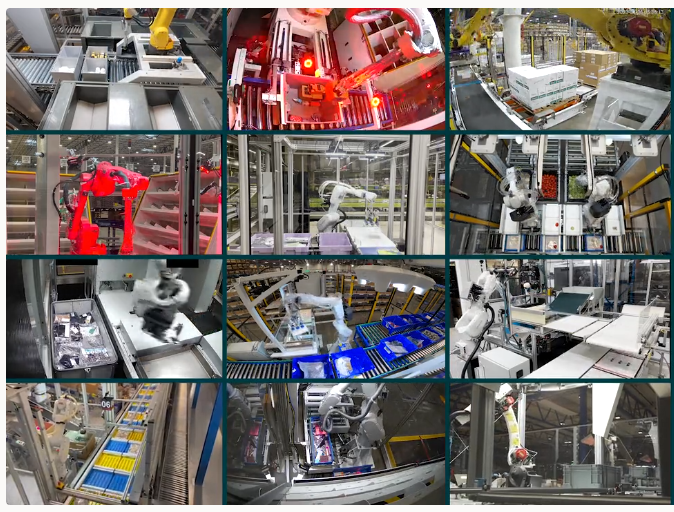}\\[0.2em]
        (b)
    \end{minipage}
    \hspace{0.02\textwidth}
    \begin{minipage}[t]{0.39\textwidth}
        \centering
        \includegraphics[width=\linewidth,height=0.15\textheight,keepaspectratio]{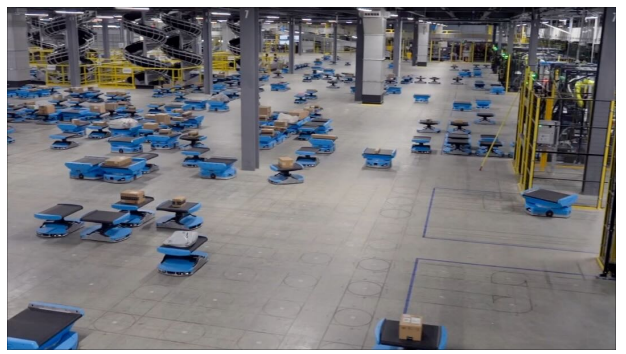}\\[0.2em]
        (c)
    \end{minipage}

    \vspace{0.3em}

    \begin{minipage}[t]{0.39\textwidth}
        \centering
        \includegraphics[width=\linewidth,height=0.15\textheight,keepaspectratio]{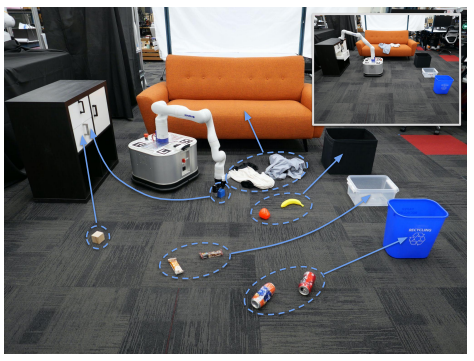}\\[0.2em]
        (d)
    \end{minipage}
    \hspace{0.02\textwidth}
    \begin{minipage}[t]{0.39\textwidth}
        \centering
        \includegraphics[width=\linewidth,height=0.15\textheight,keepaspectratio]{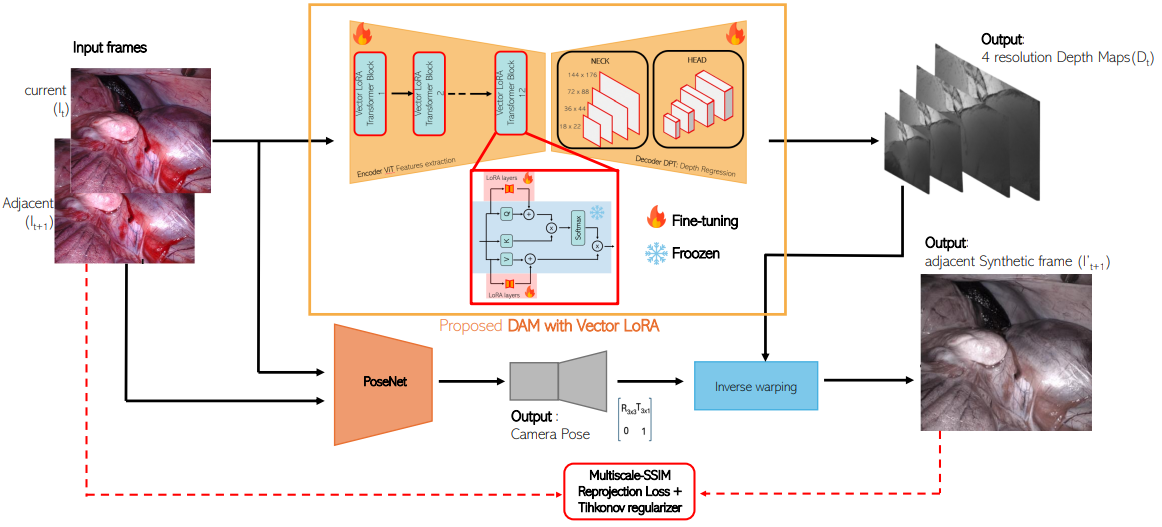}\\[0.2em]
        (e)
    \end{minipage}

    \vspace{0.3em}

    \begin{minipage}[t]{0.39\textwidth}
        \centering
        \includegraphics[width=\linewidth,height=0.15\textheight,keepaspectratio]{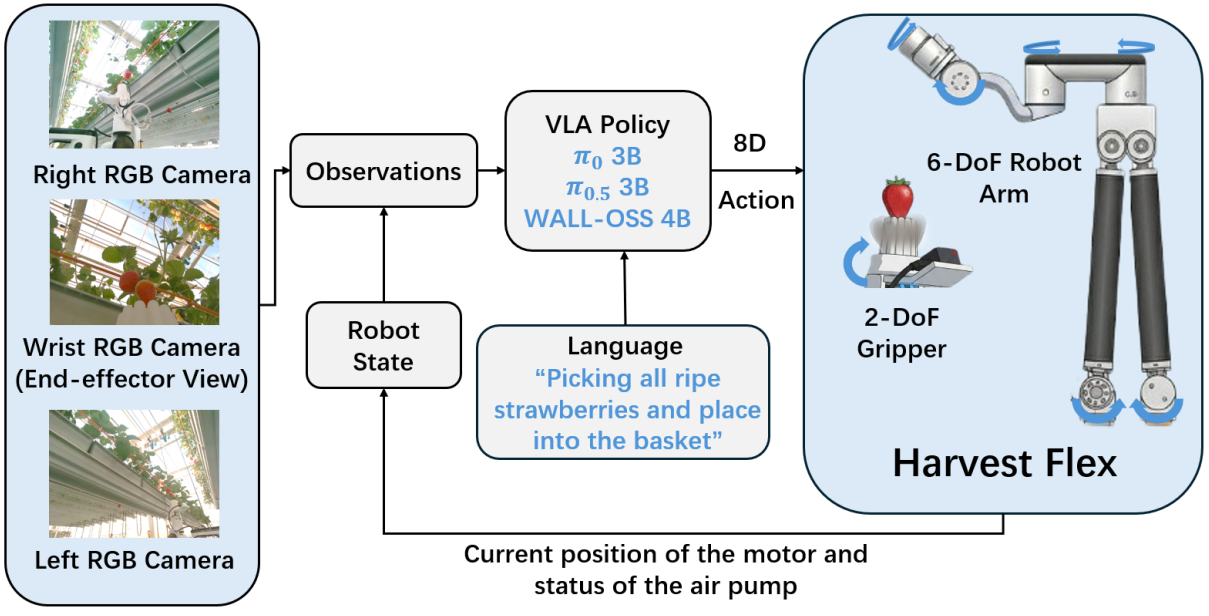}\\[0.2em]
        (f)
    \end{minipage}
    \hspace{0.02\textwidth}
    \begin{minipage}[t]{0.39\textwidth}
        \centering
        \includegraphics[width=\linewidth,height=0.15\textheight,keepaspectratio]{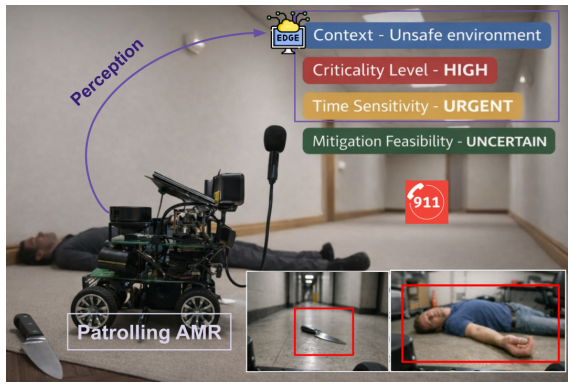}\\[0.2em]
        (g)
    \end{minipage}

    \vspace{0.3em}

    \begin{minipage}[t]{0.39\textwidth}
        \centering
        \includegraphics[width=\linewidth,height=0.15\textheight,keepaspectratio]{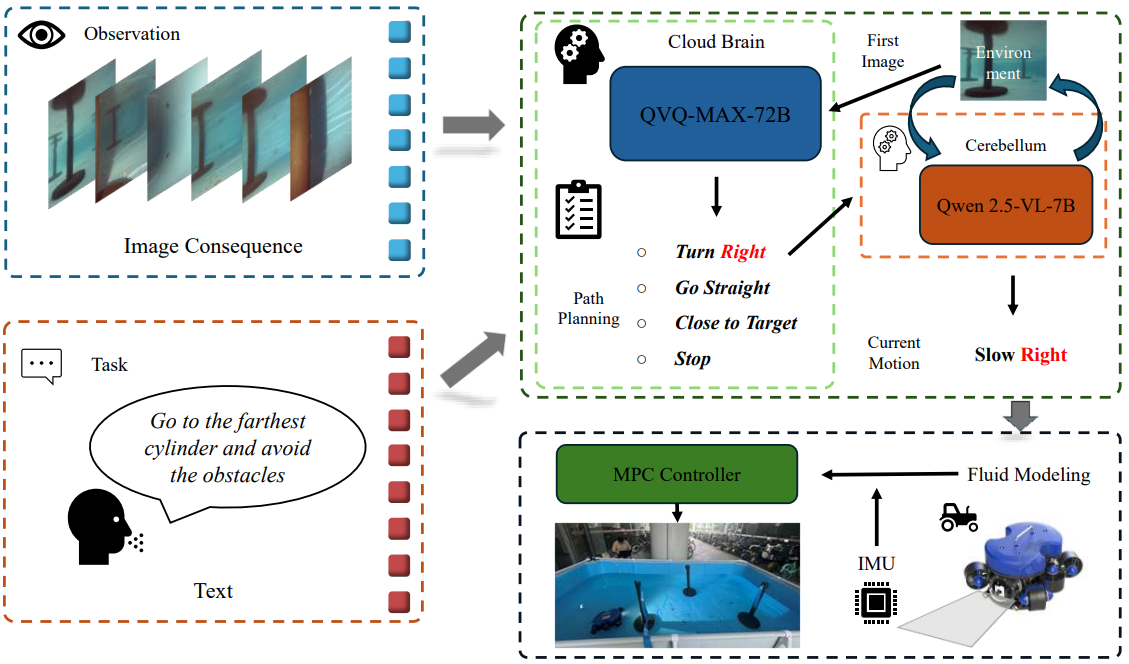}\\[0.2em]
        (h)
    \end{minipage}
    \hspace{0.02\textwidth}
    \begin{minipage}[t]{0.39\textwidth}
        \centering
        \includegraphics[width=\linewidth,height=0.15\textheight,keepaspectratio]{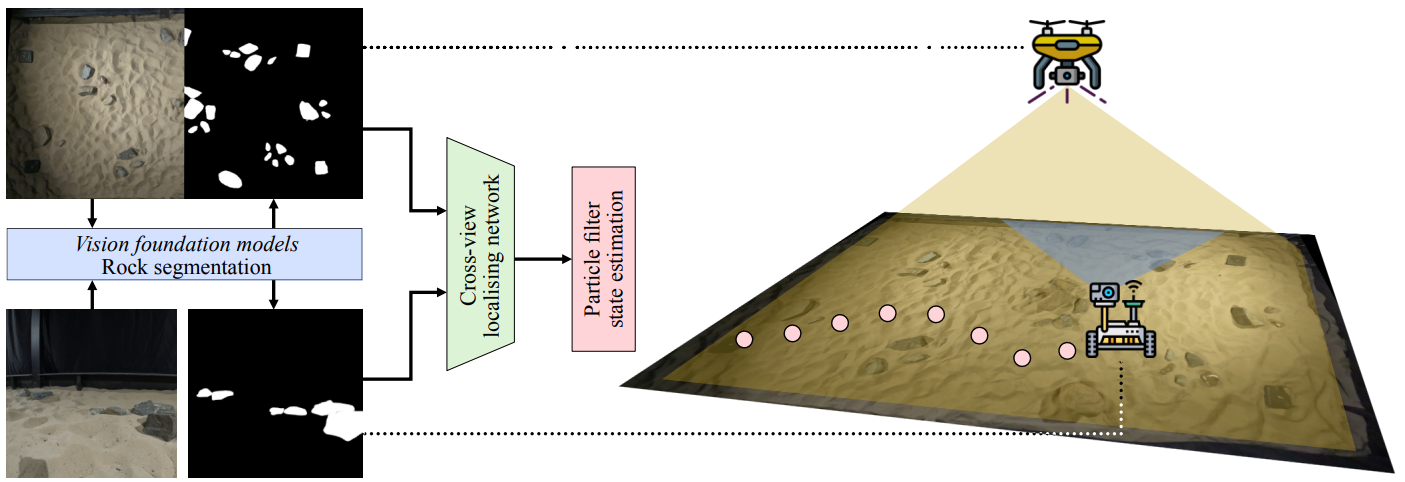}\\[0.2em]
        (i)
    \end{minipage}

    \caption{Representative literature methods per application domain: (a) Agentic mobility (GNM \citep{shah2023gnm}), (b) Industrial manipulation (RFM-1 \citep{covariant2024rfm1}), (c) Supply operations (DeepFleet \citep{agaskar2025deepfleet}), (d) Service robots (TidyBot \citep{wu2023tidybot}), (e) Medical robots (DARES \citep{zeinoddin2024dares}), (f) Cognitive agrisystems (HarvestFlex \citep{zhao2026harvestflex}), (g) Crisis agents (See Something, Say Something \citep{cancelli2026seesomething}), (h) Maritime robotics (UnderwaterVLA \citep{wang2025underwatervla}), and (i) Space robotics (Planetary cross-view localisation \citep{holden2026vision}).}
    \label{fig:application_domains_examples}
\end{figure}

The incorporation of FMs in robotic solutions has greatly boosted multiple aspects of their core technologies (e.g., autonomy, complex decision-making, semantic reasoning, etc.). This has in turn facilitated their widespread use in a wide range of challenging real-world application domains, including: a) Agentic mobility, b) Industrial manipulation, c) Supply operations, d) Service robots, e) Medical robots, f) Cognitive agrisystems, g) Crisis agents, h) Maritime robotics, and i) Space robotics, as discussed in Section \ref{sec:Taxonomy} and further detailed below. Additionally, representative literature methods per application domain are illustrated in Fig. \ref{fig:application_domains_examples}. The full details of the application domains discussed in this section are provided in Appendix~\ref{sec:application_domains_details}. In particular, for each application domain, the appendix contains the complete domain definition, as well as the extensive list of representative methods.

\subsection{Agentic mobility}
\label{ssec:application-domains_mobility}

The integration of FMs has induced a paradigm shift in autonomous movement, transitioning from rigid path-following routines to autonomous agentic solutions, where robots leverage high-level reasoning to understand mission objectives and to adapt to environmental changes. For semantic spatial grounding, VLMaps \citep{huang2023visual} constructs CLIP-indexed spatial memories by ingesting visual-language features into $3$D maps, enabling navigation to specific rooms or objects via natural language queries without task-specific retraining. For cross-embodiment and cross-site generalization, ViNT \citep{shah2023vint} formalizes navigation as a sequence prediction problem over images and poses, enabling a single visual backbone to drive diverse mobile platforms across varied environments. In road environments, DriveGPT-4 \citep{xu2024drivegpt4} conditions planning on linguistic reasoning to produce human-readable rationales and intermediate decisions that are easier to inspect, to debug, and to align with safety standards.

\subsection{Industrial manipulation}
\label{ssec:application-domains_industrial}

FMs are fundamentally redefining industrial automation by replacing rigid, task-specific routines with flexible, general-purpose agents that excel in unstructured manufacturing environments. RFM-$1$ \citep{covariant2024rfm1} is deployed within high-throughput production cells, leveraging visuo-motor backbones trained on millions of real-world pick actions to achieve zero-shot generalization to novel objects. In order to address hardware variability, OpenVLA \citep{openvla} utilizes parameter-efficient fine-tuning to simplify transfer across robotic platforms, reducing per-line retraining, while enabling operators to issue high-level, language-conditioned tasks. Moreover, RT-$2$ \citep{rt2} capitalizes on web-to-robot knowledge transfer to boost robustness against long-tail objects and complex instructions on assembly and kitting lines.

\subsection{Supply operations}
\label{ssec:application-domains_supply}

FMs are drastically evolving autonomous supply operations, by providing robots with the flexibility and intelligence needed to handle unstructured, dynamic, and complex logistics tasks, adapting to new items, facility layouts, and operational disruptions without manual reprogramming. At a broader scale, logistics operations utilize FMs for sustainable planning, multi-agent coordination, and end-to-end decision support, targeting `Logistics 5.0' goals, such as greener routing \citep{nicoletti2024green}. By integrating multi-agent systems with foundation backbones, autonomous supply chains can link demand forecasting, resource allocation, and transport routing through shared representations and agentic reasoning \citep{xu2024multi,nicoletti2025foundation}.

\subsection{Service robots}
\label{ssec:application-domains_service}

The integration of FMs represents a major paradigm shift for service robots, transforming them into adaptable agents capable of navigating unstructured domestic environments and interpreting natural language instructions. TidyBot \citep{wu2023tidybot} uses few-shot LLM reasoning to learn personalized tidying preferences directly from user conversations, grounding abstract `put-away' commands into specific physical actions across novel scenes. Extending these capabilities to multi-step procedures, ELLMER \citep{mon2025embodied} performs tasks, such as coffee brewing, by integrating visual, force, and linguistic feedback with a retrieval-augmented memory mechanism that adapts its plans on the fly under uncertainty.

\subsection{Medical robots}
\label{ssec:application-domains_medical}

FMs in the medical domain serve as a critical bridge between high-level clinical knowledge and low-level robotic execution, increasingly integrated into surgical perception and decision-making pipelines for high-stake, high-variability tasks. A key application is monocular depth estimation: Surgical-DINO \citep{cui2024surgical} adapts DINOv$2$ using a LoRA adapter for endoscopic imagery, while DARES \citep{zeinoddin2024dares} tailors a Depth Anything Model using self-supervised Vector-LoRA to align with surgical scene statistics. Additionally, EndoDDC \citep{lin2026endoddc} addresses sparse-to-dense depth reconstruction for endoscopic robotic navigation through diffusion-based depth completion, supporting reliable $3$D reconstruction and safe instrument guidance.

\subsection{Cognitive agrisystems}
\label{ssec:application-domains_agrisystems}

The integration of FMs into cognitive agrisystems marks a transition towards adaptive, embodied intelligence in field operations, where VLA architectures ingest high-level semantic instructions and translate them into precise motor outputs in real-time. Unlike traditional agricultural robots that rely on hard-coded rules, FMs bridge the semantic gap by reasoning through the variability of biological environments, such as shifting light, overlapping foliage, and irregular crop shapes \citep{yin2025foundation}. For instance, HarvestFlex \citep{zhao2026harvestflex} employs VLA policies for real greenhouse strawberry harvesting, a long-horizon task challenged by occlusion and specular reflections, while FM-based reasoning also supports task planning in crop monitoring and field management \citep{cuaran2026visual}.

\subsection{Crisis agents}
\label{ssec:application-domains_crisis}

FMs are fundamentally reforming disaster response and public safety, by enabling crisis agents to transition from rigid, remote-controlled setups to autonomous systems, capable of high-level reasoning in unpredictable and hazardous environments. In particular, SafeGuard ASF \citep{canh2026safeguard} combines multi-modal hazard perception with agentic reasoning for real-time fire-risk detection and disaster recovery. Additionally, a robotic fire-risk detection system based on dynamic knowledge graphs and LLM-enhanced multi-modal reasoning is presented in \citep{pan2025robotic}, demonstrating how FM-based reasoning can support emergency response in safety-critical settings.

\subsection{Maritime robotics}
\label{ssec:application-domains_maritime}

The incorporation of FM intelligence has significantly bolstered the capabilities of autonomous maritime systems, allowing them to perceive, to reason, and to act more effectively within complex aquatic environments. These models are specifically engineered to navigate typical underwater challenges, such as high turbidity, limited visibility, and severe communication constraints that often degrade traditional robotic sensors. In particular, UnderwaterVLA \citep{wang2025underwatervla} introduces a dual-brain VLA architecture for autonomous underwater navigation, combining multi-modal reasoning with embodied control for improving robustness under degraded visual and communication conditions. Additionally, MarineInst \citep{zheng2024marineinst} and MarineGPT \citep{zheng2023marinegpt} demonstrate FM capabilities in bridging raw marine visual data, semantic understanding, and domain-specific natural-language knowledge, thereby supporting richer perception and reasoning modules for maritime robotic platforms.

\subsection{Space robotics}
\label{ssec:application-domains_space}

The integration of FMs is critically transforming the field of astro-embodied intelligence, enabling robotic agents to reason, to adapt, and to perceive within unstructured, off-world environments, where human intervention is physically impossible. These models provide strong priors and zero-shot generalization capabilities that are crucial for operating under the extreme conditions and data scarcity typical of planetary missions. A primary application involves the usage of SAM for universal crater detection \citep{giannakis2023deep}, which utilizes promptable segmentation to identify features across diverse planetary imagery without requiring domain-specific retraining. Beyond basic detection, FMs are being extended to facilitate autonomous terrain understanding and complex geological analysis \citep{giannakis2023deep, zhao2024crater,holden2026vision}, allowing robots to make high-stake decisions independently in remote and hazardous space settings.

\section{Public datasets}
\label{sec:Datasets}

Having systematically investigated the robotic FM literature using different criteria (Sections \ref{sec:Taxonomy}-\ref{sec:app_domains}), this section outlines the main public datasets that have been introduced so far for developing and evaluating robotic FM methods. Rather than reproducing an exhaustive catalog, the discussion below is organized around the main dataset families that recur in the robotic FM literature; for each family, its scope/definition, its typical uses, a set of representative resources, and its main current gaps are summarized.

\begin{itemize}

    \item \underline{Large-scale cross-embodiment trajectory corpora}: These comprise massive collections of real-world robot interaction trajectories aggregated across many robotic platforms, tasks, and environments, typically pairing synchronized RGB-D observations, natural-language instructions, and low-level action tokens. Their main use lies in pre-training generalist VLA policies that transfer robustly across different kinematic structures and operating conditions, effectively serving as the backbone training data for cross-embodiment models. Representative resources include Open X-Embodiment \citep{openx}, AgiBot World \citep{bu2025agibot}, DROID \citep{khazatsky2024droid}, BridgeData V2 \citep{walke2023bridgedata}, and RoboMIND \citep{wu2024robomind}. Their main current gaps comprise the scarcity of real-world physical interaction signals (particularly tactile and force/torque sensing), the near-absence of failure and recovery recordings, and a strong skew towards table-top manipulation settings.

    \item \underline{Simulation environments and benchmark suites}: These comprise photo-realistic or game-engine simulators and curated task suites that algorithmically generate environments, tasks, and ground-truth supervision for embodied agents. Their main usefulness lies in enabling reproducible, low-cost, large-scale training and standardized evaluation, while also supporting long-horizon, language-conditioned policy learning and the systematic study of knowledge transfer across tasks. Representative resources include AI2-THOR \citep{kolve2017ai2}, RLBench \citep{james2020rlbench}, CALVIN \citep{mees2022calvin}, BEHAVIOR-1K \citep{li2023behavior}, and LIBERO \citep{liu2023libero}. Their main current limitations comprise the persistent sim-to-real gap, the limited fidelity in modeling contact-rich physics, and a narrow object and scene diversity relative to the real world.

    \item \underline{Vision and language navigation datasets}: These comprise benchmarks that pair natural language instructions or goals with embodied navigation episodes in 3D environments. Their main use lies in training and evaluating language-guided navigation, open-vocabulary goal finding, and fine-grained instruction following. Representative resources include REVERIE \citep{qi2020reverie}, RxR \citep{ku2020room}, VLN-CE \citep{krantz2020beyond}, ScaleVLN \citep{wang2023scaling}, and GOAT-Bench \citep{khanna2024goat}. Their main current gaps comprise a heavy reliance on static photorealistic scans, the scarcity of dynamic, human-populated, and outdoor settings, and limited modeling of social/interaction constraints.

    \item \underline{Real-world perception and scene-understanding datasets}: These comprise large, richly annotated real-world sensor collections (RGB-D, LiDAR, and multi-camera) targeting 3D indoor reconstruction and outdoor autonomous-driving perception. Their main use lies in pre-training and benchmarking visual and 3D perception backbones (e.g., semantic and instance segmentation, detection, and mapping) that subsequently feed downstream robotic policies. Representative resources include nuScenes \citep{caesar2020nuscenes}, Waymo Open \citep{sun2020scalability}, Semantic-KITTI \citep{behley2019semantickitti}, ScanNet++ \citep{yeshwanth2023scannet++}, and Matterport3D \citep{chang2017matterport3d}. Their main current gaps comprise their predominantly passive nature (lacking embodiment and action labels), a domain concentration on driving and indoor scenes, and the limited presence of contact or manipulation context.

    \item \underline{Egocentric human-video corpora}: These comprise large-scale first-person (egocentric) and closely related fixed-view videos of humans performing everyday activities, capturing natural manipulation, interaction, and navigation behavior. Their main use lies in providing web-scale human priors for representation pre-training, affordance and activity understanding, and learning from human demonstration to bootstrap robot policies. Representative resources include Ego4D \citep{grauman2022ego4d}, EPIC-KITCHENS \citep{damen2018scaling}, Ego-Exo4D \citep{grauman2024ego}, and COM Kitchens \citep{maeda2024kitchens}. Their main current gaps comprise the still comparatively small number of corpora available at this scale, the absence of explicit robot action/control labels, and the embodiment gap between human and robot morphologies that complicates direct transfer.

    \item \underline{Human-robot interaction and dialogue datasets}: These comprise datasets that capture multimodal human-robot communication, including dialogue, clarification questions, gestures, and speech grounded in embodied tasks. Their main use lies in training and evaluating conversational grounding, interactive task completion, clarification-seeking behavior, and reasoning over interleaved vision-language-action streams. Representative resources include RoboVQA \citep{sermanet2024robovqa}, TEACh \citep{padmakumar2022teach}, CVDN \citep{thomason2020vision}, DialFRED \citep{gao2022dialfred}, and NatSGD \citep{shrestha2024natsgd}. Their main current gaps comprise their limited scale, the largely simulated or scripted nature of the recorded dialogues, and the scarcity of real-world, synchronized multimodal (e.g., speech plus gesture) interaction data.

\end{itemize}

The full details of the public datasets discussed in this section are provided in Appendix~\ref{sec:datasets_details}. In particular, the appendix contains the complete dataset catalog, organized as detailed per-task tables (grouping the various benchmarks by main robotic task and reporting, for each entry, its year, scale, semantic classes, modalities, annotation type, domain, environment type, temporality, embodiment, and a short description), as well as the extensive list of global observations regarding the main current trends across robotic FM datasets.

\section{Current challenges}
\label{sec:challenges}

Despite the large body of works that have recently been introduced in the field of robotic FM-based methods and the tremendous advancements accomplished, significant challenges and open research problems still remain, which if robustly addressed will further increase the efficiency, reliability, acceptance, and adoption of such solutions in real-world deployment scenarios. In the remaining of this section, the main challenges identified in the literature are systematically examined and outlined.

\subsection{Data aspects}
\label{ssec:challenges_data}

Unlike benchmark requirements and availability in fields like NLP and CV, robotic data is physically-grounded, high-dimensional, multi-modal, and typically expensive to collect. In this respect, the following specific challenges related to data aspects in robotic FM research are present:
\begin{itemize}
    \item \underline{Scarcity of physical-world data (C1.1)}: Despite the availability of internet-scale visual/text benchmarks, robotic solution development lacks comparable datasets of high-quality, diverse, real-world robot trajectories \citep{rt1}. The main factor for the latter comprises the data collection cost, which requires physical hardware, human teleoperation/demonstration/instruction, and significant time. Additionally, the added difficulty of collecting `long-tail scenarios' (i.e., rare and unexpected events with huge impact on the system performance) makes the learning of safe recovery behaviors even more challenging \citep{park2025generative}.
    \item \underline{Embodiment heterogeneity (C1.2)}: Robotic data is generally not uniform, since it is produced by different robots with diverse hardware and physical configurations (e.g., degrees of freedom, sensor suites, kinematics, etc.) \citep{openx}. Additionally, there is no single format for continuous action data across diverse embodiments, without losing physical meaning/properties \citep{gato}. The above result into an inherent difficulty to transfer knowledge between different robotic platforms \citep{liu2024rdt}.
    \item \underline{Maintaining high data quality (C1.3)}: Increasing the data scale usually leads to improved performance, conditioned on the fact of maintaining a high-quality in the captured data \citep{rt2}. However, human-captured teleoperation data often includes suboptimal movements, hesitation, or outright failures \citep{hu2023toward}. Additionally, manual inspection of immense amounts of expert demonstration videos is tremendously expensive.
    \item \underline{Domain transfer gap (C1.4)}: Typical approaches to overcome the data scarcity problem is to make use of simulation or data from another (similar) domain. However, simulation engines often fail to accurately reproduce complex physics (e.g., friction, object deformations, etc.), leading to low performance in real-world settings \citep{xiao2024one}. On the other hand, incorporation of data from multiple, diverse environments is shown to have a more crucial impact on training, than simply scaling the available datasets \citep{openx, rt2}.
    \item \underline{Multi-modal and temporal alignment (C1.5)}: Robots need to efficiently integrate vision, language, and proprioception streams to accomplish robust performance in real-world settings. However, the high and different frequency of the various information sequences, along with the inherent difficulty in mapping continuous physical-world actions to discrete information tokens, results into precision degradation (e.g., in dexterous tasks) \citep{zhou2025chatvla}.
\end{itemize}

\subsection{Computation aspects}
\label{ssec:challenges_computation}
In any real-world robotic application case, time performance and resource requirements constitute a cornerstone regarding fundamental safety and stability specifications. In this respect, the following specific challenges related to computation aspects in robotic FM research are present:

\begin{itemize}
    \item \underline{Need for real-time inference (C2.1)}: Robot control loops typically operate at high frequencies (e.g., often more than $50$Hz), in order to achieve stability and safe execution \citep{chi2025diffusion}. However, the inherently extreme scale of FMs naturally introduces significant inference latencies, ranging from some hundreds of milliseconds up to multiple seconds \citep{firoozi2025foundation, rt2}. This is highly likely to result into FM-based solutions not to be applicable in multiple cases or to lead to inaccurate policy executions \citep{ameperosa2025rocoda}.
    \item \underline{Need for safety bound interval (C2.2)}: Apart from the inference time itself, safe robot operation requires the application of certain safety routines or the execution of corrective actions \citep{sinha2024real}. The latter impose additional time constraints during execution, i.e. a strict upper bound for overall end-to-end latency \citep{firoozi2025foundation, sinha2024real}.
    \item \underline{Constrained onboard resources (C2.3)}: Robotic platforms typically pose specific and strict size, weight, and energy specifications, setting particular limitations to the onboard embedded GPUs and their operation. Contrary to FM execution on cloud, resource-abundant environments, edge devices exhibit limitations in terms of available memory, processing time, and energy/thermal tolerance, prohibiting the deployment of full-scale and best-performing versions of robotic FMs \citep{openx, yue2024deer}.
\end{itemize}

\subsection{Safety and security aspects}
\label{ssec:challenges_safety}
The fundamental advantage of FMs in robotic applications relies on their ability to combine high-level semantic reasoning with low-level, physical planning/execution procedures. The latter though raise particular safety and security concerns, which extend beyond conventional/traditional robotic settings. In this respect, the following specific challenges related to safety and security aspects in robotic FM research are present:

\begin{itemize}
    \item \underline{Semantic-physical space mismatch (C3.1)}: It is very common for FMs to generate policies that are linguistically, logically, and semantically sound, but can likely lead to dangerous or even physically impossible scenarios (with the extreme case being that of the presence of hallucinations) \citep{lin2025llm, yin2025towards}. This is mainly due to the typical limitation of most FMs to account for physical-world parameters (e.g., friction, torque limits, material properties, etc.) in their reasoning process \citep{firoozi2025foundation}. To make matters worse, FMs typically lack the ability to estimate the consequences of their (erroneous) actions in the physical environment \citep{rt2, lin2025llm}.
    \item \underline{Adversarial vulnerabilities (C3.2)}: As the number of modality streams and data scale that a single FM can process rises, the respective (cyber) attack surface of the (often cloud-hosted) model itself increases proportionally \citep{radanliev2026threats}. In particular, even relatively minor perturbations in the data can result into significant behavior deviations or incorrect/hazardous actions \citep{wang2025exploring}.
    \item \underline{Inaccurate uncertainty quantification (C3.3)}: Especially in human-robot collaboration settings, the robotic agent needs to constantly maintain a precise estimation of its own state/policy uncertainty \citep{wang2025aleatoric}. Current FMs though do not reliably balance their reactions with respect to aleatoric (environmental noise) and epistemic (model ignorance) uncertainty \citep{Marquesetal2024}. This bottleneck becomes even more evident in onboard deployment settings, where increased inference latency or compressed models are used (due to resource constraints) \citep{rt2}. 
\end{itemize}

\subsection{Embodiment aspects}
\label{ssec:challenges_embodiment}

FMs have equipped robots with unprecedented perception, reasoning, and execution capabilities under real-world, dynamic environments. However, the critical issue that arises comprises that of grounding robot tasks on a physical platform, which poses specific sensor, actuator, and hardware constraints. In this respect, the following specific challenges related to embodiment aspects in robotic FM research are present:
\begin{itemize}
    \item \underline{Heterogeneity of robot action spaces (C4.1)}: Unlike the case of other AI deployment scenarios, robots exhibit a vast variety in terms of physical forms, morphologies, and capabilities. The latter renders difficult to deploy a model trained on a specific hardware setup (e.g., dual-arm, mobile platform, etc.) to another (e.g., quadruped, drone, etc.), due to difference in key robotic specifications (e.g., varying degrees-of-freedom) \citep{openx, team2024octo}. Additionally, the absence of a universally applicable, standardized control interface across different robots, makes the development of robust, generalized, platform-independent FMs particularly difficult \citep{zheng2025universal}.
    \item \underline{Sim-to-real gap (C4.2)}: In an attempt to alleviate from the need for large-scale, high-quality robotic interaction data, simulation engines are often used for data generation. However, employing simulation environments does not always result into the assembly of benchmarks with sufficient diversity in both task execution and environment settings \citep{jonnarth2025sim}. Moreover, simulation engines are typically prone to not model accurately complex physical-world dynamics (e.g., friction, deformation, fluid interactions, etc.), which in turn results into reduced robot performance \citep{makoviychuk2021isaac, ai2025review}.
    \item \underline{Limited physical space grounding (C4.3)}: Despite the unprecedented semantic capabilities of FMs, the mapping of such high-level, semantic representations to low-level, real-world physics is not guaranteed. In particular, current FMs exhibit limitations in precise spatial, geometric, and physical interactions \citep{qi2025beyond}. Additionally, the lack of sufficient contextual, common-sense knowledge in FMs may result into the generation of failure modes in the real-world \citep{chen2024spatialvlm, huang2023voxposer}.
    \item \underline{Constrained haptic capabilities (C4.4)}: The so-called `final frontier' problem of physical intelligence comprises the difficulty of robots to replicate nuanced sensorial feedback, like humans do in the real world \citep{yang2024binding}. The latter constitutes inherently a multi-faceted problem, including perspectives related to the scarcity of large-scale haptic/tactile data, robot hardware heterogeneity, sim-to-real variance, and high-frequency temporal reaction demands \citep{zhang2025unitachand}.
\end{itemize}

\subsection{Reasoning aspects}
\label{ssec:challenges_reasoning}
While inference capabilities of FMs in other AI domains (e.g., CV, NLP, etc.) has successfully achieved the incorporation of rich, abstract logic in most cases, their application to physically-grounded, robot agents exhibits significant obstacles. In this respect, the following specific challenges related to reasoning aspects in robotic FM research are present:

\begin{itemize}
    \item \underline{Lack of physical common sense knowledge (C5.1)}: The usual embodiment-agnostic understanding of the world by FMs (e.g., gravity, object permanence, material properties, etc.) makes their real-world deployment difficult \citep{firoozi2025foundation, kawaharazuka2024real}. Additionally, the lack of sufficient causal reasoning capabilities hinders robots to predict the physical consequences of their actions \citep{toberg2024commonsense}.
    \item \underline{Constrained long-horizon planning (C5.2)}: The reasoning performance of robotic FMs tends to degrade exponentially, as the number of required steps increases \citep{palme}. The latter has a great impact, especially in cases of large-scale operational environments \citep{lisondra2025embodied}. Moreover, the training process itself poses difficulties to FMs to connect long-term goals to specific sequences of decision steps \citep{saycan}.
    \item \underline{Imprecise explanations of robot behaviors (C5.3)}: FMs in robotics comprise by nature massive architectures that integrate multi-modal information streams, while connecting high-level reasoning with low-level, physical control \citep{rt1}. In contrast to traditional modular robotic solutions, where failure modes can be traced to specific components, FMs implement end-to-end generalist policies that make it particularly difficult to specify the factors that have led to the execution of a specific physical action \citep{kawaharazuka2024real}.
\end{itemize}

\subsection{Evaluation aspects}
\label{ssec:challenges_evaluation}
Provided that FMs bridge the entire gap between high-level, semantic reasoning and low-level, physical execution in dynamic environments, the assessment of their performance and robustness exhibits particular characteristics, compared to other conventional AI solutions. In this respect, the following specific challenges related to evaluation aspects in robotic FM research are present:
\begin{itemize}
    \item \underline{Lack of a unified evaluation framework (C6.1)}: Despite the availability of a series of standardized, task-oriented, intuitive performance metrics, no holistic and integrated evaluation protocol is currently present to assess the multi-factored robotic failure cases \citep{firoozi2025foundation}. In particular, the available (and typically binary) metrics are often proven to be coarse, failing to clearly highlight the underlying factors leading to low performance (e.g., inefficient bi-manual coordination, asymmetric arm usage, etc.) \citep{jiang2022vima, openx}. Additionally, the development of independent, domain-specific metrics across different fields makes even more difficult to assess robot performance with respect to different architectures or motion parameters \citep{rt1}. 
    \item \underline{Imprecise generalization ability assessment (C6.2)}: Despite the key driving force of FMs to exhibit increased (especially zero-shot) generalization ability, its comprehensive, accurate, and robust assessment is especially difficult \citep{rt2, liang2023code}. In particular, the introduction of even minimal distribution shifts in task descriptions or observation domains can have a great impact on the resulting robot behavior (or even failure) \citep{kube2026robotic}.
\end{itemize}

\section{Future research directions}
\label{sec:future_directions}

The introduction of FMs in the field of robotics has led to unprecedented accomplishments and advances in all core robotic technologies, as discussed in Sections \ref{sec:Evolution}-\ref{sec:app_domains}. Despite these tremendous developments, several open challenges are still present, which pose restrictions in the wider deployment of robotic solutions in real-world scenarios, as outlined in Section \ref{sec:challenges}. In this respect, this section discusses the main and most promising future research directions towards achieving the goal of developing efficient, robust, and general-purpose robotic agents, in correspondence to the challenges described in Section \ref{sec:challenges}.

\subsection{Architectural evolution}
\label{ssec:research_architectures}

VLAs comprise one of the most promising types of NN architectures, which implement a unified neural function for mapping visual observations, linguistic instructions, and proprioceptive states directly to low-level, robot control policies. Their core characteristic is their end-to-end integration nature, which enables robotic agents to perceive/reason about their surrounding environment and, subsequently, to react upon the input stimuli in a single forward/inference pass. In this respect, the following promising research directions emerge:
\begin{itemize}
    \item \underline{Heterogeneous action spaces}: An ambitious goal of current research efforts focuses on the development of universal robotic FMs, capable of operating across different robotic platforms. This requires the robust addressing of the action heterogeneity (or cross-embodiment generalization, stated alternatively) problem \citep{openx}. In this context, future research could emphasize on the development of embodiment-agnostic action spaces, which would enable FMs to predict desired end-effector trajectories/forces, while their eventual mapping to particular, platform-specific, low-level operations could be controlled by a hardware-specific modulation component \citep{team2024octo}. This direction mainly aims to address the challenges of `Embodiment heterogeneity' (C1.2) and `Heterogeneity of robot action spaces' (C4.1), as outlined in Section~\ref{sec:challenges}.
    \item \underline{Sophisticated action sequence tokenization}: One of the main challenges in the design of VLA transformer architectures comprises the discrete, tokenized representation of continuous space robotic actions. The conventional approach of binning each dimension to a fixed number of values inherently leads to loss of precision, especially for dexterous tasks \citep{rt2, gato}. In this context, future research could emphasize on more sophisticated tokenization methods capable of capturing the detailed dynamics of continuous actions, while maintaining the efficiency of autoregressive decoding. This direction mainly aims to address the challenge of `Multi-modal and temporal alignment' (C1.5) (Section~\ref{sec:challenges}), particularly the precision loss incurred when mapping continuous physical actions to discrete tokens.
    \item \underline{Diffusion and flow-based action modeling}: When a robot attempts to perform an action, there is a set of practically infinite trajectories to select from. Common architectures (e.g. transformers) tend to learn the average of such possible motions, leading to performance degradation or failure \citep{florence2022implicit}. On the contrary, flow- and diffusion-based objectives facilitate the modeling of temporal dynamics in a continuous latent space. In this context, future research could emphasize on the latter aspects, targeting to equip the robots with the ability to predict the consequences of their actions in a continuous space representation \citep{chi2025diffusion}. This direction mainly aims to address the challenges of `Multi-modal and temporal alignment' (C1.5) and `Lack of physical common sense knowledge' (C5.1) (Section~\ref{sec:challenges}), by enabling robots to anticipate the physical consequences of their candidate actions.    
\end{itemize}

\subsection{Multi-modal embodied intelligence}
\label{ssec:research_physical}

The main driving-force of current FMs lies on the incorporation of vision and language information, while true embodied intelligence requires a holistic understanding of the physical world through the integration of touch, sound, and force sensing. The latter is essential especially for tasks requiring dexterity and delicate interaction. In this respect, the following promising research directions emerge:
\begin{itemize}
    \item \underline{Tactile information}: Vision is often insufficient for performing contact-rich tasks and reaching human-level dexterity capability, which relies on incorporating feedback regarding the surface texture, slip, and force \citep{yu2023mimictouch}. For achieving the latter, the robust integration of tactile information is essential, capitalizing on the current/early-works on developing tactile FMs.  In this context, future research could emphasize on the latter aspects, aiming at enhancing robots to efficiently manipulate deformable objects \citep{wu2019learning}. This direction mainly aims to address the challenges of `Constrained haptic capabilities' (C4.4) and `Scarcity of physical-world data' (C1.1) (Section~\ref{sec:challenges}), the latter regarding the under-representation of large-scale tactile data.
    \item \underline{Proprioception and force control}: Beyond tactile sensing, the incorporation of proprioception (i.e., the robot's sense of its own body state) and force control is crucial for realizing contact-rich manipulaton tasks. Current solutions usually focus only on position control, while exploratory works, like the integration of joint torque sensors, emerges as a promising approach \citep{kumar2021rma}. In this context, future research could emphasize on further exploiting dense proprioception and force technologies, in order to boost more fine-grained manipulation and efficient human-robot collaboration \citep{liu2023demonstration}. This direction mainly aims to address the challenges of `Limited physical space grounding' (C4.3) and `Constrained haptic capabilities' (C4.4), as outlined in Section~\ref{sec:challenges}.
    \item \underline{Auditory feedback}: The sound signal can often bear significant information regarding the occurrence of critical events during robot execution, complementing the corresponding visual and force feedback streams \citep{dimiccoli2022recognizing}. Additionally, audio processing requires significantly less resources than the respective visual stream. In this context, future research could emphasize on further investigating the exploitation of auditory feedback, enabling robots to operate more accurately and reliably in dynamic and noisy environments \citep{mejia2024hearing}. This direction mainly aims to address the challenge of `Multi-modal and temporal alignment' (C1.5) (Section~\ref{sec:challenges}), by enriching the set of complementary sensory streams that robots can integrate.
\end{itemize}

\subsection{Reasoning and long-horizon autonomy}
\label{ssec:research_reasoning}

Current VLA-based solutions have demonstrated reliable performance on immediate reactive tasks. However, the implementation of multiple real-world activities requires the robust handling of complex, long-horizon, multi-stage missions. In this respect, the following promising research directions emerge:
\begin{itemize}
    \item \underline{Long-horizon memory frameworks}: Due to constraints in the memory capacity of current FMs, robots often tend to repeat unsuccessful policies, due to the relatively limited context window that they maintain and which might not include information about the previous policy failure \citep{firoozi2025foundation}. To this end, extending the memory capabilities of robotic solutions would have a great impact on long-horizon goal achievement. In this context, future research could emphasize on enhancing long-horizon autonomy, by incorporating structured summaries of past interactions (instead of full-sequence replays) \citep{wu2026reasoning}, latent-graph memory formalisms to maintain experiences over longer contexts \citep{chen2026polarmem}, and stage-aware reward signals for more efficient policy learning in multi-step setups \citep{chen2025sarm}. This direction mainly aims to address the challenge of `Constrained long-horizon planning' (C5.2), as outlined in Section~\ref{sec:challenges}.
    \item \underline{Hierarchical semantic representations}: In case of robot operation in cluttered environments, the usage of hierarchical abstractions (e.g., scene graphs) of the processed semantic information is proven to be efficient in simplifying and strengthening decision-making \citep{gu2024conceptgraphs}. In this context, future research could emphasize on realizing task planning and reasoning on such hierarchical abstractions, so as to both reduce computational cost (i.e., avoid intense computations at the pixel or joint angle level) and to achieve more robust decision making \citep{gao2025vla,hughes2022hydra}. This direction mainly aims to address the challenges of `Constrained long-horizon planning' (C5.2), `Need for real-time inference' (C2.1), and `Constrained onboard resources' (C2.3), as outlined in Section~\ref{sec:challenges}. 
\end{itemize}

\subsection{World foundation models}
\label{ssec:research_world}

The deployment of FMs in robotic applications requires high-quality, diverse, and large-scale embodied interaction data. Given the fact that the collection of the latter datasets is typically expensive and time-consuming, world foundation models are being investigated as an alternative solution for trajectory generation. In this respect, the following promising research directions emerge:
\begin{itemize}
    \item \underline{Physics-informed generative models}: WMs have already been proven to generate photorealistic data of high quality. However, there still exists a gap between simulated physics and real-world dynamics, which would enable robots to learn and to act more reliably in actual operational settings \citep{ai2025review, lee2025dynscene}. In this context, future research could emphasize on explicitly incorporating constraints like gravity, friction, and fluid dynamics into the video generation process \citep{xie2024physgaussian}. Another closely related aspect concerns the integration of sophisticated physics-grounded predictive pipelines, which would serve as value functions for model-based planning, especially in long-horizon tasks \citep{zhou2024robodreamer, yang2023unisim}. This direction mainly aims to address the challenges of `Scarcity of physical-world data' (C1.1), `Domain transfer gap' (C1.4), `Sim-to-real gap' (C4.2), and `Lack of physical common sense knowledge' (C5.1), as outlined in Section~\ref{sec:challenges}.
    \item \underline{Action-conditioned scenario generation}: Research in world models regarding action-conditioned scenario generation has advanced from plain video prediction towards the creation of unified models, which function as physically grounded, interactive cognitive engines \citep{bruce2024genie}. In particular, significant efforts are devoted on integrating differentiable physics and unified geometric representations for ensuring the spatio-temporal consistency and accuracy of the generated predictions over long horizons \citep{lee2025dynscene, tu2025role}. In this context, future research could emphasize on further extending current capabilities, especially focusing on cross-embodiment generalization and `long-tail' scenarios, so as to boost the translation of human-centric video data into physically plausible robotic trajectories across multiple hardware platforms \citep{openx, team2024octo}. This direction mainly aims to address the challenges of `Scarcity of physical-world data' (C1.1), `Embodiment heterogeneity' (C1.2), `Maintaining high data quality' (C1.3), and `Imprecise generalization ability assessment' (C6.2) (Section~\ref{sec:challenges}), since generated long-tail and cross-embodiment scenarios can also serve to systematically probe model generalization.
\end{itemize}

\subsection{Safety and verification}
\label{ssec:research_safety}
Given the increasing deployment of robots in dynamic, unstructured environments, their interaction and collaboration with humans requires the robust addressing of safety risks. For achieving this, FM-based solutions need to incorporate a combined approach, consisting of both high-level semantic understanding and low-level, deterministic safety aspects. In this respect, the following promising research directions emerge:
\begin{itemize}
    \item \underline{Adaptive safety}: The incorporation of adaptive safety measures in robotic FMs is gradually shifting from a reactive post-processing approach towards an integrated one, where constraints are directly projected to the model's training and reasoning loops \citep{ren2023robots}. This enables robots to predict and to mitigate physical risks through grounded reasoning, prior to action execution \citep{zha2024distilling}. In this context, future research could emphasize on further enhancing safety alignment approaches through constrained learning and process reward models for evaluating individual reasoning steps and environmental affordances in real-time \citep{yu2025reward, anand2026adaptive}. This direction mainly aims to address the challenges of `Semantic-physical space mismatch' (C3.1), `Inaccurate uncertainty quantification' (C3.3), and `Need for safety bound interval' (C2.2), as outlined in Section~\ref{sec:challenges}.
    \item \underline{Formal verification}: Research in formal verification of robotic FMs is currently translating from conventional offline proofs towards modular, runtime-based assurance frameworks. In particular, current efforts largely focus on control barrier functions and reachability analysis, in order to intercept model outputs in real-time \citep{miyaoka2025control, wang2025safe}. In this context, future research could emphasize on investigating neuro-symbolic integration approaches for mapping neural outputs to formal logics \citep{cunnington2024role} and automated specification mining pipelines that employ LLMs for translating natural language instructions into precise mathematical formulas \citep{rabiei2025ltlcodegen}. This direction mainly aims to address the challenges of `Semantic-physical space mismatch' (C3.1), `Adversarial vulnerabilities' (C3.2), `Imprecise explanations of robot behaviors' (C5.3), and `Lack of a unified evaluation framework' (C6.1) (Section~\ref{sec:challenges}), the latter through standardized, formally grounded assessment criteria. 
\end{itemize}

\section{Conclusion}
\label{sec:Conclusion}

\textbf{Summary}: The introduction of Foundation Models (FMs) has resulted into transformative effects in the field of robotics, transitioning the current practice from rigid, single-task solutions towards adaptive, multi-sensory, and generalist agents, capable of operating in complex, dynamic, open-world environments. The current review has provided a holistic, thorough, systematic, and in-depth analysis of the  research landscape by delineating five distinct evolution phases, starting from early Natural Language Processing (NLP) and Computer Vision (CV) model integration to the current frontier of multi-sensory generalization and real-world deployment. Through a highly-granular, multi-criteria, taxonomic literature investigation, this work has analyzed the interplay between different foundation model types (LLMs, VFMs, VLMs, and VLAs), underlying neural network architectures, adopted learning paradigms, learning stages for skill acquisition, robotic tasks (perception, planning, navigation,
manipulation, and human-robot interaction), and real-world application domains. The above discussion has been accompanied by a methodical comparative analysis of the various categories of approaches and critical insights per defined criterion. Moreover, an overview of the publicly available datasets used for model training and evaluation was provided, organized around the main recurring dataset families and their respective scope, typical uses, representative resources, and current gaps. Furthermore, a detailed and hierarchical discussion on the current open challenges and promising future research directions in the field was incorporated.

\textbf{Concluding synthesis}: Whereas Section~\ref{sec:challenges} systematically enumerates the open challenges (C1.1-C6.2) and Section~\ref{sec:future_directions} proposes corresponding research directions, the purpose of this concluding discussion is to connect the two more in-depth, by elaborating on the most critical open problems and explicitly relating them to the body of works reviewed throughout this survey. Specifically, for each problem family (Sections~\ref{ssec:challenges_data}-\ref{ssec:challenges_evaluation}), a distinction is made between aspects that existing works already address (at least partially) and those that remain genuinely unsolved, in order to highlight where the field has made tangible progress and where future effort is needed the most.

\underline{Open problems in data}: Although internet-scale pre-training endows robotic FMs with unprecedented perceptual and reasoning priors, the field still lacks corpora of high-quality, diverse, real-world robot trajectories comparable to those available in NLP and CV (C1.1-C1.5). Existing works partially mitigate this gap through large-scale cross-embodiment aggregation \citep{openx, team2024octo, liu2024rdt} and, more recently, through world models that synthesize interaction data and serve as model-based planners \citep{zhou2024robodreamer, yang2023unisim, bruce2024genie}. Nevertheless, the systematic generation of safety-critical `long-tail' scenarios \citep{park2025generative} and the curation of high-quality demonstrations from suboptimal teleoperation data \citep{hu2023toward} remain largely open, keeping data the most fundamental bottleneck of the field.

\underline{Open problems in computation}: A second cluster of unresolved problems concerns the balance between the scale of FMs and the strict real-time, resource, and safety-bound constraints of robotic platforms (C2.1-C2.3). Robot control loops typically require frequencies above $50$Hz \citep{chi2025diffusion}, whereas FM inference latencies range from hundreds of milliseconds to several seconds \citep{firoozi2025foundation, rt2}, and onboard edge devices impose tight memory, energy, and thermal limits \citep{yue2024deer}. Despite progress in model compression and efficient inference, the deployment of full-scale, best-performing FMs under hard latency and onboard-resource constraints remains, to a large extent, an open problem.

\underline{Open problems in safety and security}: The defining strength of robotic FMs, namely the coupling of high-level semantic reasoning with low-level physical execution, is also the source of their most pressing safety and security concerns (C3.1-C3.3). Existing efforts increasingly shift from reactive post-processing towards safety constraints embedded in the reasoning loop \citep{ren2023robots, zha2024distilling} and runtime assurance, via control barrier functions, reachability analysis, and neuro-symbolic verification \citep{miyaoka2025control, wang2025safe, cunnington2024role}. However, these approaches are still at early-stage and reliable uncertainty quantification, together with formal guarantees in high-stake human-robot interaction settings, remains an open challenge.

\underline{Open problems in embodiment}: A further set of open problems stems from the difficulty of grounding generalist policies on heterogeneous physical platforms (C4.1-C4.4). The heterogeneity of robot morphologies and action spaces is partially addressed by embodiment-agnostic action representations and hardware-specific modulation \citep{openx, team2024octo}, while the replication of nuanced sensorial feedback, often referred to as the `final frontier' of physical intelligence \citep{yang2024binding}, is being approached through early tactile FMs \citep{yu2023mimictouch, zhang2025unitachand}. Nonetheless, the scarcity of large-scale haptic data, the absence of standardized control interfaces, and the persistent sim-to-real gap maintain cross-embodiment generalization far from solved.

\underline{Open problems in reasoning}: The application of FM inference to physically-grounded agents exposes open problems in physical common sense and long-horizon autonomy (C5.1-C5.3). Existing works seek to extend reasoning over longer contexts through structured memory summaries, latent-graph memory formalisms, and stage-aware reward signals \citep{wu2026reasoning, chen2026polarmem, chen2025sarm}, as well as through reasoning over hierarchical semantic abstractions, such as scene graphs \citep{gu2024conceptgraphs}. So far, the lack of causal, physically-grounded common-sense knowledge and the exponential degradation of reasoning performance with task horizon remain among the least understood and most impactful open problems.

\underline{Open problems in evaluation}: Acting as an element affecting all the above, the field still lacks a unified protocol for assessing FM-based robotic systems (C6.1-C6.2). Current metrics are typically coarse and binary, failing to expose the underlying factors of multi-factored failure cases \citep{firoozi2025foundation}, while the reliable, comprehensive assessment of (especially zero-shot) generalization under distribution shift is particularly difficult \citep{kube2026robotic}. The absence of holistic, standardized evaluation frameworks therefore constitutes an under-addressed open problem that hampers fair comparison and reproducibility across the works reviewed in this survey.

\textbf{Unifying semantic-physical grounding gap}: It is critical to highlight that all the above problem clusters are indeed not independent, but rather facets of a single overarching open problem, namely the gap between high-level semantic reasoning and low-level physical grounding (most evidently manifested by challenges C3.1, C4.3, and C5.1) \citep{firoozi2025foundation, lin2025llm}. At the same time, these problems are also strongly interdependent: Synthesizing data through world models (C1.1) simultaneously bears on the sim-to-real gap (C4.2) and on physical common sense (C5.1), whereas hard latency constraints (C2.1) limit safety verification (C3.1-C3.3) levels that can be executed online. In terms of prioritization, latency reduction and unified evaluation appear comparatively tractable in the near term, whereas physical common-sense reasoning and human-level haptic intelligence constitute longer-horizon goals; robustly bridging the semantic-physical grounding gap, however, remains the central question on which the convergence of the field ultimately depends.

\textbf{Limitations of current survey}: In line with the comparative positioning summarized in Table~\ref{tab:fm_surveys_rev} and the corresponding discussion in Section~\ref{sec:Intro}, the most significant limitation of this survey is that it provides a qualitative, taxonomic, and comparative analysis rather than a comprehensive quantitative benchmarking of the examined methods; reproducible, head-to-head empirical evaluation is deliberately left outside its scope and the reported insights rely on results as stated by the original works. Additionally, as also acknowledged in Table~\ref{tab:fm_surveys_rev}, the current review inevitably constitutes a snapshot of a rapidly evolving field, so relevant works will continue to appear after its cutoff. Beyond these main limitations, the survey covers partly certain connecting aspects, namely hardware design and low-level control theory, under-represented sensory modalities (such as tactile and auditory feedback), and a number of more specialized topics (e.g., multi-robot/swarm systems and the legal, ethical, and societal dimensions of deployment).

\textbf{Scope for future surveys}: Building upon the above, the identified boundaries directly suggest concrete avenues for subsequent similar reviews. Initially, dedicated empirical/quantitative benchmarking surveys would complement the present qualitative analysis with reproducible, standardized, head-to-head comparisons, particularly for cross-embodiment generalization. Additionally, focused surveys on world foundation models for robotics could consolidate the rapidly growing literature on physics-informed and action-conditioned generation. Moreover, dedicated reviews on the safety and formal verification of embodied FMs would deepen the treatment of the safety-related challenges that this work necessarily covers at a higher level. Moreover, deeper/targeted domain-specific surveys (e.g., on surgical/medical, agricultural, space, or maritime robotics) and horizontal surveys on the under-covered topics noted above (i.e., hardware co-design, under-represented modalities, sustainability/energy efficiency, and the societal and trust dimensions of deployment) would further facilitate developments in the field.

\textbf{Outlook}: Overall, while FMs have already redefined what robotic agents can perceive, reason about, and accomplish, their transition from impressive demonstrations to dependable real-world deployment hinges on the robust addressing of the open problems outlined above. Towards this goal, by explicitly relating these problems to the works reviewed throughout this survey, the present analysis aims to offer a useful map for prioritizing future research towards more autonomous, reliable, and trustworthy robotic systems.

\section*{Acknowledgments}
This work has received funding from the European Union's Horizon Europe research and innovation programme under Grant Agreement No. 101189557 project TORNADO (foundaTion mOdels for Robots that haNdle smAll, soft and Deformable Objects) and No. 101168042 project TRIFFID (auTonomous Robotic aId For increasing First responders Efficiency). The views and opinions expressed in this paper are those of the authors only and do not necessarily reflect those of the European Union or the European Commission.

\bibliographystyle{tmlr}
\bibliography{references}
\clearpage
\appendix

The following appendices provide additional technical details and analyses that complement the main text.

\section{Comparative analysis with recent surveys}
\label{sec:comparative_surveys_details}

This section provides the detailed comparison of the present survey with recent ones in the field, complementing the condensed summary presented in Table~\ref{tab:fm_surveys_rev} of the main text. In particular, for each surveyed work, Table~\ref{tab:fm_surveys_full} reports the following aspects: a) Survey scope, b) Review methodology, c) Primary contributions, and d) Main limitations.

\begin{table}[!ht]
  \caption{Detailed comparative analysis of recent surveys in the field of foundation models in robotics.}
  \label{tab:fm_surveys_full}
  \centering
  \scriptsize

  \setlength{\aboverulesep}{0pt}
  \setlength{\belowrulesep}{0pt}
  \setlength{\tabcolsep}{2pt}
  \renewcommand{\arraystretch}{0.9}

  \setlist*[tabitem]{before=\vspace{2.2pt}, after=\vspace{2.2pt}}

  \rowcolors{2}{gray!25}{white}

  \resizebox{0.93\textwidth}{!}{%
  \begin{tabular}{@{}|
    >{\justifying\arraybackslash}m{1.2cm}|
    >{\justifying\arraybackslash}m{1.85cm}|
    >{\justifying\arraybackslash}m{2.9cm}|
    >{\justifying\arraybackslash}m{6.75cm}|
    >{\justifying\arraybackslash}m{6.75cm}|
  @{}}
    \toprule
    \rowcolor{gray!40}
    \headerbreak{Article} &
    \headerbreak{Scope} &
    \headerbreak{Methodology} &
    \headerbreak{Primary contributions} &
    \headerbreak{Limitations} \\
    \midrule

    \citet{hu2023toward} (arXiv) &
    Broad survey and meta-analysis on the use of FMs towards general-purpose robots &
    \begin{tabitem}
      \item Literature analysis per robotic task
      \item Performance analysis per robotic task
      \item Analysis of 301 papers
    \end{tabitem} &
    \begin{tabitem}
      \item Separation between CV/NLP and native robotic FMs
      \item Identification of trends based on experimental results
      \item Discussion on open challenges and future research directions
    \end{tabitem} &
    \begin{tabitem}
      \item Discussion of early developments in robotic FMs
      \item No systematic and theoretical comparison between different approaches
      \item No discussion on recent developments (e.g., world models, diffusion policies, etc.)
      \item No discussion on application domains
    \end{tabitem} \\
    \midrule

    \citet{xu2024survey} (arXiv) &
    Task-oriented survey focusing on robotic manipulation &
    \begin{tabitem}
      \item Literature categorization to high-level planning and low-level control approaches
      \item Analysis of 64 papers
    \end{tabitem} &
    \begin{tabitem}
      \item Discussion on form and assistant perspectives of planning
      \item Analysis of focused components in the learning process
      \item Discussion on open challenges and future research directions
    \end{tabitem} &
    \begin{tabitem}
      \item Investigation of only robotic manipulation approaches
      \item Limited literature coverage
      \item No systematic and theoretical comparison between different approaches
      \item No empirical/experimental benchmarking of the surveyed methods
      \item No discussion on application domains
    \end{tabitem} \\
    \midrule

    \citet{ma2024survey} (arXiv) &
    VLA-oriented survey emphasizing on embodied AI aspects &
    \begin{tabitem}
      \item Taxonomic analysis of VLAs based on individual components, control policies, and high-level task planning
      \item Analysis of 785 papers
    \end{tabitem} &
    \begin{tabitem}
      \item Systematic comparison of methods
      \item Analysis of strengths and limitations per category
      \item Discussion on open challenges and future research directions
    \end{tabitem} &
    \begin{tabitem}
      \item No explicit discussion on other FM types (i.e., LLMs, VFMs, VLMs)
      \item No systematic analysis and comparisons across multiple criteria (i.e., NN architecture, learning paradigm, learning stage, robotic task)
      \item No empirical/experimental benchmarking of the surveyed methods
      \item No discussion on application domains
    \end{tabitem} \\
    \midrule

    \citet{jang2024unlocking} (IJCAS) &
    Application-oriented survey focusing on robotic autonomy &
    \begin{tabitem}
      \item Literature categorization based on perception, task planning, and control
      \item Analysis of environmental setups
      \item Analysis of 255 papers
    \end{tabitem} &
    \begin{tabitem}
      \item Discussion of impact of individual FM components on robot autonomy
      \item Analysis of robotic platforms and simulation environments
      \item Discussion on future research directions
    \end{tabitem} &
    \begin{tabitem}
      \item No systematic analysis across multiple criteria (i.e., NN architecture, learning paradigm, learning stage)
      \item No theoretical comparison between different approaches
      \item No empirical/experimental benchmarking of the surveyed methods
      \item No discussion on application domains
    \end{tabitem} \\
    \midrule

    \citet{kawaharazuka2024real} (AR) &
    Application-oriented survey focusing on component replacement with FMs &
    \begin{tabitem}
      \item Literature categorization based on perception, planning, and data augmentation
      \item Analysis of 225 papers
    \end{tabitem} &
    \begin{tabitem}
      \item Analysis of input-output relationships in FMs
      \item Discussion on the role of FMs in perception, motion planning, and control
      \item Discussion on future research directions
    \end{tabitem} &
    \begin{tabitem}
      \item No systematic analysis across multiple criteria (i.e., FM type, NN architecture, learning paradigm, learning stage)
      \item No theoretical comparison between different approaches
      \item No empirical/experimental benchmarking of the surveyed methods
      \item No discussion on application domains
    \end{tabitem} \\
    \midrule

    \citet{firoozi2025foundation} (IJRR) &
    Broad survey emphasizing on the use of FMs in robotics and embodied AI &
    \begin{tabitem}
      \item Literature analysis regarding decision-making, planning, and control
      \item Analysis of 233 papers
    \end{tabitem} &
    \begin{tabitem}
      \item Literature investigation regarding FMs that are native and relevant to robotics
      \item Analysis of embodied AI aspects
      \item Discussion on open challenges and future research directions
    \end{tabitem} &
    \begin{tabitem}
      \item No systematic analysis across multiple criteria (i.e., FM type, NN architecture, learning paradigm, learning stage)
      \item No theoretical comparison between different approaches
      \item No empirical/experimental benchmarking of the surveyed methods
      \item No discussion on application domains
    \end{tabitem} \\
    \midrule

    \citet{xiao2025robot} (Neurocomputing) &
    Robot learning-oriented survey for generalist robots &
    \begin{tabitem}
      \item Analysis of robot learning in manipulation, navigation, task planning, and reasoning
      \item Analysis of 464 papers
    \end{tabitem} &
    \begin{tabitem}
      \item Systematic and hierarchical analysis of robot learning techniques
      \item Discussion on open challenges and future research directions
    \end{tabitem} &
    \begin{tabitem}
      \item No systematic analysis across multiple criteria (i.e., FM type, NN architecture, learning stage)
      \item No theoretical comparison between different approaches
      \item No empirical/experimental benchmarking of the surveyed methods
      \item No discussion on application domains
    \end{tabitem} \\
    \midrule

    \citet{tayyab2025foundation} (arXiv) &
    Integration-oriented survey focusing on real-world deployment &
    \begin{tabitem}
      \item Literature analysis regarding integrated, system-level strategies
      \item Feasibility analysis in real-world environments
      \item Analysis of 175 papers
    \end{tabitem} &
    \begin{tabitem}
      \item Literature categorization across simulation-driven design, open-world execution, sim-to-real transfer, and adaptable robotics
      \item Discussion on open challenges and future research directions
    \end{tabitem} &
    \begin{tabitem}
      \item No systematic analysis across multiple criteria (i.e., FM type, NN architecture, learning paradigm, learning stage, robotic task)
      \item No theoretical comparison between different approaches
      \item No empirical/experimental benchmarking of the surveyed methods
      \item No discussion on application domains
    \end{tabitem} \\
    \midrule

    \citet{kawaharazuka2025vision} (IEEE Access) &
    VLA-oriented survey emphasizing on full-stack aspects &
    \begin{tabitem}
      \item VLA literature analysis regarding both software and hardware perspectives
      \item Analysis of 427 papers
    \end{tabitem} &
    \begin{tabitem}
      \item Systematic analysis of architectural designs and learning paradigms
      \item Discussion on practical implementation aspects
      \item Discussion on future research directions
    \end{tabitem} &
    \begin{tabitem}
      \item No explicit discussion on other FM types (i.e., LLMs, VFMs, VLMs)
      \item Coarse categorization of all learning paradigms and robotic tasks
      \item No theoretical comparison between different approaches
      \item No empirical/experimental benchmarking of the surveyed methods
      \item No discussion on application domains
    \end{tabitem} \\
    \midrule

    \citet{sapkota2025vision} (arXiv) &
    VLA-oriented survey detailing research evolution aspects &
    \begin{tabitem}
      \item Literature analysis covering FM evolution, key progress areas, and application domains
      \item Analysis of 299 papers
    \end{tabitem} &
    \begin{tabitem}
      \item Analysis of research evolution
      \item Discussion on architectural innovations, training strategies, and real-time inference
      \item Analysis of application domains
      \item Discussion on open challenges and future research directions
    \end{tabitem} &
    \begin{tabitem}
      \item No explicit discussion on other FM types (i.e., LLMs, VFMs, VLMs)
      \item No systematic analysis across multiple criteria (i.e., NN architecture, learning paradigm, learning stage)
      \item No theoretical comparison between different approaches
      \item No empirical/experimental benchmarking of the surveyed methods
    \end{tabitem} \\
    \midrule

    \textbf{Current survey} &
    Thorough and systematic analysis of the landscape of FMs in robotics &
    \begin{tabitem}
      \item Structured and systematic literature analysis, querying 6 major databases, using specific inclusion/exclusion criteria, and applying iterative screening
      \item Analysis of 438 papers
    \end{tabitem} &
    \begin{tabitem}
      \item Investigation of research evolution in 5 distinct phases
      \item Highly-granular multi-criteria (6) taxonomic analysis of literature, examining FM types, NN architectures, learning paradigms, learning stages, robotic tasks, and application domains
      \item Per criterion methodical comparative analysis of different approaches and insights reporting
      \item Comprehensive and hierarchical discussion on current challenges and future research directions
    \end{tabitem} &
    \begin{tabitem}
      \item No empirical/quantitative benchmarking of the surveyed methods
      \item Constitutes a snapshot of a rapidly evolving field
    \end{tabitem} \\

    \bottomrule
  \end{tabular}%
  }
\end{table}

\clearpage

\section{Details of literature review methodology}
\label{sec:sup_methodology-details}

This section provides the full details of the structured literature review methodology that was followed in order to efficiently and thoroughly identify/map the robotic FM literature, while at the same time detecting key concepts and trends (complementing the condensed presentation of Section \ref{sec:literature} of the main text). The adopted systematic approach, which ensures comprehensiveness and relevance of the selected research works, consists of the iterative main steps described below.

\subsection{Scope and objectives formulation}
The fundamental goal of the performed survey study was to review the robotic FM literature, i.e., approaches for various robotic tasks (namely, perception, planning, navigation, manipulation, and human-robot interaction) whose execution relies on the use of an underlying FM. In particular, the focus was on identifying the most recent advancements and works with substantial contribution to the field, emphasizing the following objectives: a) The main categories of methods based on multiple criteria, b) The utilized datasets, c) Current challenges, and d) Future research directions.

\subsection{Literature search}
In order to ensure broad and thorough coverage of the relevant literature, the search strategy involved querying several major scientific databases, including IEEE Xplore, Google Scholar, Scopus, DBLP, arXiv, and Web of Science. The actual search was performed by combining targeted keywords/terms (e.g., ``foundation model'', ``robotics'', ``vision-language-action'', ``large language model'', etc.) with Boolean operators (i.e., ``AND'', ``OR'', ``NOT''). To guarantee contemporary relevance, the search primarily focused on research works published within the last five years; however, certain earlier seminal studies were also included. An example of the query used in the Scopus database is provided below. It needs to be emphasized that this Scopus-style string comprises only an illustrative example of one search instance and does not constitute the complete literature search protocol.\\
\noindent\textcolor{violet}{TITLE-ABS-KEY}(\textcolor{teal}{``foundation model''} \textcolor{cyan}{OR} \textcolor{teal}{``vision-language model''} \textcolor{cyan}{OR} \textcolor{teal}{``visual foundation model''} \textcolor{cyan}{OR} \textcolor{teal}{``large language model''} \textcolor{cyan}{OR} \textcolor{teal}{``vision-language-action''} \textcolor{cyan}{OR} \textcolor{teal}{``robotic foundation model''})\\
\textcolor{blue}{AND} \textcolor{violet}{TITLE-ABS-KEY}(\textcolor{teal}{``robot''} \textcolor{cyan}{OR} \textcolor{teal}{``robotics''} \textcolor{cyan}{OR} \textcolor{teal}{``manipulation''} \textcolor{cyan}{OR} \textcolor{teal}{``navigation''} \textcolor{cyan}{OR} \textcolor{teal}{``human-robot interaction''} \textcolor{cyan}{OR} \textcolor{teal}{``embodied''})\\
\textcolor{blue}{AND} PUBYEAR $>$ $2020$\\
\textcolor{blue}{AND} PUBYEAR $<=$ $2026$\\
\textcolor{blue}{AND} (\textcolor{violet}{SRCTYPE}(j) \textcolor{cyan}{OR} \textcolor{violet}{SRCTYPE}(p))\\
\textcolor{blue}{AND} \textcolor{violet}{EXCLUDE}(DOCTYPE, ``bk'')\\
Multiple search steps were performed iteratively, involving refinements to the employed keywords so as to retrieve more relevant works. Moreover, the list of references of each research article was also analyzed in order to identify additional relevant studies. The combination of the above elements (i.e., the use of six complementary databases, the iterative keyword refinement, and the extensive backward reference-checking/chaining procedure) renders the overall literature review process relatively robust against missing critical/important works in the field. In particular, this design is capable of capturing seminal studies whose original terminology predates the currently established FM one and which, therefore, would be missed by a single keyword-based query.

Regarding the choice of the search mechanism, the adopted keyword plus reference search strategy was deliberately favored over a pure embedding-based semantic retrieval over paper embeddings. The main rationale is that the keyword-based approach offers improved reproducibility (the exact query strings, databases, and date bounds can be reported and re-executed), transparency (the inclusion of each work can be traced to explicit criteria rather than to an opaque similarity score), and precision/recall control (the Boolean structure allows for the systematic tuning of the retrieved set). In contrast, embedding-based semantic retrieval depends on a particular embedding model, a seed corpus, and a similarity threshold, none of which are easily auditable or reproducible across the six heterogeneous databases considered, most of which do not natively expose such a functionality. Nevertheless, it is acknowledged that embedding-based semantic search can serve as a useful complementary tool, particularly for further improving recall under terminology drift, a role that in the current study was effectively fulfilled by the iterative refinement and the reference-chaining steps described above.

\begin{figure}[!t]
    \centering
    \begin{minipage}[c]{0.47\textwidth}
        \centering
        \includegraphics[width=\linewidth]{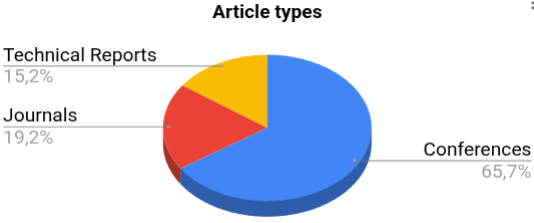}\\[0.3em]
        (a)
    \end{minipage}
    \hfill
    \begin{minipage}[c]{0.47\textwidth}
        \centering
        \includegraphics[width=\linewidth]{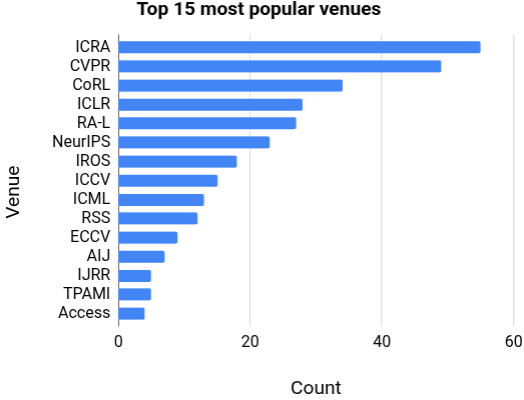}\\[0.3em]
        (b)
    \end{minipage}
\caption{Key bibliometric analytics regarding robotic FM literature: (a) Article types, and (b) Top-15 most popular venues.}
\label{fig:bibliometric-analytics}
\end{figure}

\subsection{Screening}
Initial screening relied on excluding duplicate records, non-English papers, and articles without full-text access, in order to maintain the integrity of the review study. Then, article selection was performed taking into account title and abstract information, so as to eliminate irrelevant works. Subsequently, in-depth full-text review was performed for considering only research studies that: a) Focus on demonstrating approaches for implementing various robotic tasks (i.e., perception, planning, navigation, manipulation, and human-robot interaction), where a FM comprises a key algorithmic component, and b) Exhibit substantial theoretical and/or experimental contributions. Additionally, priority was given to research works originating from prominent robotics and AI/ML publication venues. Eventually, a total of $438$ articles were selected for analysis and were included as references in the  manuscript.

Key bibliometric analytics regarding the performed literature review study are illustrated in Fig.~\ref{fig:bibliometric-analytics}.

\subsection{LLM assistance} Regarding the use of automated tools during the preparation process of the manuscript, LLM assistance was employed only on specific occasions and solely as a complementary aid for cross-checking potential gaps in the literature search described above (i.e., as a supplementary means of identifying relevant works that might have been missed by the database queries). However, all substantive scholarly work was carried out manually by the authors; in particular, the taxonomy design, the paper inclusion/exclusion decisions, all comparative methodical analyses, the study of individual works, and the inclusion and verification of all references were carefully performed and validated by the authors. The authors take full responsibility for the entire content of the manuscript.

\section{Foundation model types}
\label{sec:fm_types_details}

This section provides the full details of the different types of FMs that are utilized in robotics, complementing the condensed presentation of Section~\ref{sec:fm_types} of the main text. For each FM type (namely, LLMs, VFMs, VLMs, and VLAs), the complete category definition, the corresponding main advantages and limitations, as well as the extensive list of representative methods per sub-category of supported functionality are provided below.

\subsection{Large Language Models (LLMs)}
\label{sup:ssec:llm_in_robotics}

In the context of robotics, LLMs are primarily used as high-level, cognitive task planners and reasoning engines that generate the sequence of operations that are necessary for accomplishing a stated goal \citep{saycan,liang2023code,chen2024autotamp}. So far, they have been shown to be particularly effective for tasks that require sophisticated, logical reasoning and complex decision-making. In practice, LLMs translate high-level natural language inputs to low-level robot behaviors, by turning free-form instructions and short state summaries into typed goals, multi-step plans, executable code, constraints, and run-time feedback. Their main advantages are \citep{li2025large}: a) Robustness in instruction translation, where free-form, flexible, human-like natural language instructions are transformed into formal, actionable representations, b) Efficiency in task decomposition and sequencing, where complex, long-horizon natural language instructions are mapped to concrete sequences of robot sub-tasks, and c) Increased generalization ability, where the large amount of general-purpose knowledge incorporated in an LLM can boost the generation of robust action plans in diverse, complex, real-world environments. On the contrary, LLMs  exhibit the following critical limitations \citep{tayyab2025foundation}: a) Lack of embodiment and grounding, where LLMs operate in a semantic space that does not incorporate connections to the physical environment; hence, leading to action plans that may not be feasible, b) Presence of hallucinations, where the LLM inference procedure may result into semantically/syntactically correct outcomes that, however, correspond to irrational actions, c) Increased latency, where the inherently high computational complexity of LLMs may not be suitable for real-time operational settings, and d) Input bias and sensitivity, where obscure or ambiguous input instructions may lead to unpredictable or biased action plans.

In terms of supported functionality/operation, LLM-based systems can be classified into the following main categories:
\begin{itemize}
    \item \underline{Goal/constraint grounding and context}: LLMs serve as a reasoning and translation interface that maps high-level, abstract human instructions (i.e., goal and context) to low-level, physical robot actions (i.e., grounding). In particular, SayCan \citep{saycan} employs a base LLM mechanism (PaLM \citep{palm}) for filtering using value-based affordances, so that the selected skills to be maintained within capabilities and context. Additionally, LM-Nav \citep{shah2023lm} converts instructions into visually grounded way-points that a planner can subsequently follow in a navigation setting. For larger spaces, SayPlan \citep{rana2023sayplan} associates language with $3$D scene graphs, so that plans can remain consistent across different rooms and floors.
    \item \underline{Command interpretation and code synthesis}: LLMs in principle act as natural language to code translators, enabling robots to receive high-level, human-like flexible instructions, instead of requiring the writing of conventional code. Specifically, Code-as-Policies \citep{liang2023code} compiles instructions into robot Application Programming Interface (API) code using code-generating LLMs, which renders robot behavior easy to inspect and to reuse. Similarly, AutoTAMP \citep{chen2024autotamp} translates requests into task-and-motion planning (TAMP) algorithmic specs that a symbolic planner checks for geometric and kinematic feasibility. Additional works focus on generating behavioral trees or decomposing tasks into formalized subproblems \citep{ao2025llm,kwon2025fast,liu2025delta}, synthesizing policy code from video-plus-text prompts \citep{xie2025robotic}, and constructing language-conditioned $3$D value maps to guide placement and grasping \citep{huang2023voxposer}. 
    \item \underline{Task planning and long-horizon reasoning}: LLMs implement a high-level, cognitive mechanism that decomposes abstract human goals into sequential, grounded actions and maintains contextual awareness over multiple subsequent steps. In particular, SELP \citep{wu2025selp} performs the mapping of language instructions to temporal logic representations, while combining constrained decoding with domain tuning, so that the generated plans to eventually satisfy safety and efficiency constraints. Similarly, LLM-GROP \citep{zhang2025llm} cross-checks language-driven task instructions with motion feasibility and perceptual information for improving execution in cluttered settings. Moreover, hierarchical goal decomposition can speed up planning and robustness of long-horizon task execution \citep{kwon2025fast,liu2025delta,liu2024enhancing}.
    \item \underline{Perception-aware and multimodal integration}: The LLM core language processing functionality is enhanced by taking into account the robot's surrounding visual and physical world. In this context, PaLM-E \citep{palme} incorporates visual and proprioceptive information streams into the inference process, so that the model decisions to better correspond to real-world observations and actions. In a similar way, Chain-of-Modality \citep{wang2025chain} employs prompts originating from both human-performing videos and auxiliary, associated signals (i.e., muscle or audio), in order to define both a task plan and corresponding control parameters from a single demonstration. Moreover, associating language inputs with visual grounding and $3$D value maps can also facilitate more accurate robot manipulation \citep{huang2023voxposer}.
    \item \underline{Navigation and spatial understanding}: LLMs translate abstract navigation commands into physically grounded, actionable paths within a map, primarily through the estimation of structured textual representations of the surrounding environment. For instance, LM-Nav \citep{shah2023lm} links human instructions with spatial landmarks and routes, via a CLIP-based grounding mechanism, and passes way-points to a low-level navigation module. Additionally, SayPlan \citep{rana2023sayplan} partitions a plan over a $3$D scene graph into layers and, subsequently, performs re-planning when an action step is infeasible, keeping in this way long-horizon missions on track in large-scale environments.
    \item \underline{Conversational interfaces and teleoperation}: LLMs create intuitive, conversational interfaces and enable shared-control teleoperation, making robots accessible to non-expert users. In particular, TidyBot \citep{wu2023tidybot} learns user-specific tidying conventions through conversation and transfers them to novel domestic scenarios. Additionally, LAMS \citep{tao2025lams} predicts human intent and automatically switches teleoperation modes in assistive settings, aiming at lowering cognitive load during extended task executions.  
    \item \underline{Execution-time validation, error handling, and recovery}: LLMs translate technical run-time failures into semantic problems, enabling the definition of dynamic, common-sense solutions, instead of relying solely on pre-defined, baseline, brittle recovery routines. Specifically, STATLER \citep{yoneda2024statler} collects and interprets robot state and tool feedback information, allowing the suggestion of targeted repairs without restarting the overall execution pipeline. Similarly, CAPE \citep{raman2024cape} performs re-prompting and proposes concrete fixes in case of preconditioned failures, while CoPAL \citep{joublin2024copal} realizes re-planning when divergence incidents are detected. Moreover, uncertainty estimation over planning proposals can provide additional mitigation measures, prior to acting on hardware \citep{yin2025towards}.
    \item \underline{Adaptation, efficiency, and safety}: LLMs facilitate robotic systems to be adaptive, efficient, and safe during task execution, by robustly handling novel situations and operational disruptions. In particular, Eureka \citep{ma2023eureka} writes and refines reward code using GPT-$4$ \citep{gpt4}, accelerating in this way skill acquisition across diverse robot platforms. Additionally, DrEureka \citep{ma2024dreureka} co-designs rewards and domain randomization for efficient sim-to-real transfer. AutoRT \citep{ahn2024autort} implements simultaneously instruction handling, task assignment, and safety checking across multiple robots. Moreover, planners can encode safety rules directly (e.g., SELP’s use of temporal logic) \citep{wu2025selp}, while recent works study jailbreak-style exploits and propose corresponding defense measures \citep{ravichandran2025safety}.
    \item \underline{Knowledge retrieval and memory}: LLMs leverage their ability to access, synthesize, and store massive amounts of information from external knowledge and past execution episodes, allowing systems to overcome limitations of their immediate sensorial data or within a single planning session. For instance, ELLMER \citep{mon2025embodied} combines GPT-$4$ with a retrieval-augmented memory mechanism, so that a mobile manipulator can incorporate context, adapt plans on the fly, and complete multi-step household tasks as conditions change. Similar architectures are capable of performing inference over planning route and tool knowledge for on-demand lookup \citep{temiraliev2026retrieval,anwar2025remembr}.
\end{itemize}

\subsection{Vision Foundation Models (VFMs)}
\label{sup:ssec:vfm_in_robotics}

The ultimate goal of VFMs is to satisfy the perceptual requirements of embodied AI systems, by providing generalized, high-quality visual representations necessary for interaction with the physical world \citep{sam,oquab2023dinov2}. In the robotics setting, VFMs aim at distilling raw pixel information into rich, transferable visual features or embeddings. The latter enables a robust and generalized visual understanding that serves as the input information stream for modulating downstream policies, enabling the direct mapping of visual observations to specific control actions \citep{shangtheia}. Their main advantages are \citep{tayyab2025foundation}: a) Increased transfer learning capabilities, where visuomotor robot policies can efficiently generalize and acquire new skills, without requiring task-specific perception module learning from scratch, b) Open-world perception, where robots are enabled to recognize and to process previously unseen object and scene categories, c) Improved spatial awareness, where especially the incorporation of depth perception is crucial for enabling the execution of intricate actions with precision, and d) Increased robustness, where VFMs are shown to be less susceptible to visual distortions (e.g., presence of noise, variations in lighting conditions, etc.), compared to previous visual information processing modules. On the other hand, the main limitations of VFMs are \citep{awais2025foundation}: a) Domain specificity, where VFMs can often perform sufficiently well in a relatively narrow range of domains, b) Incomplete physical world dynamics modeling, where VFMs often exhibit limitations to generalize robustly to subtle physical dynamics, long-range temporal correlations, causal coherence, and geometric properties, and c) Increased computational cost, which may result into critical constraints regarding real-time, uninterrupted operation.

In terms of supported functionality/operation, VFM-based
systems can be classified into the following main categories:
\begin{itemize}
    \item \underline{Object recognition}: VFMs enable generalized visual recognition, by providing transferable representations that reduce the need for task-specific perception training. In particular, SAM \citep{sam} provides class-agnostic, promptable segmentation masks that facilitate isolating objects and parts, while DINOv2 \citep{oquab2023dinov2} provides robust dense visual features that transfer well across scenes and support recognition and retrieval under domain shift \citep{sam,oquab2023dinov2}. Additionally, $3$D-MVP \citep{qian20253d} uses a multi-view encoder to learn object and part-level representations that are useful for recognition and downstream manipulation. Moreover, ZeroGrasp \citep{iwase2025zerograsp} couples recognition with reconstruction, estimating object geometry and predicting grasp poses from a single RGB-D observation in near real-time.

    \item \underline{Localization}: VFMs enhance robot localization by estimating robust, semantic, and globally consistent visual representations, outperforming conventional geometric methods. Specifically, DINO-VO \citep{azhari2025dino} utilizes DINOv2-generated features and a ViT-based keypoint estimation scheme for improving robustness and generalization in monocular visual odometry at high throughput rates. Additionally, LiteVLoc \citep{jiao2025litevloc} and ZeroVO \citep{lai2025zerovo} support long-range re-localization for image-goal navigation and zero-shot cross-camera visual odometry, respectively.

    \item \underline{Object tracking}: VFMs provide trackers with rich, semantic understanding and robust long-term memory, increasing robustness with respect to occlusion, viewpoint change, and category drift. In particular, \citet{zhong2024empowering} use a pre-trained VFM to extract semantic segmentation masks with text prompts, while a recurrent policy network with offline reinforcement learning is subsequently trained from the collected demonstrations. Additional approaches make use of open-vocabulary or frozen-backbone cues for instance tracking and segmentation, which is particularly useful for handling novel objects over time \citep{guo2025openvis,fang2025decoupled}.
    
    \item \underline{Depth perception}: VFMs enable robust, generalized, and high-fidelity depth estimation, especially in scenarios where specialized sensors or traditional methods tend to under-perform. Specifically, Metric$3$D v$2$ \citep{hu2024metric3d} employs geometric priors (namely, canonical camera transformations and joint depth-normal optimization) and supports zero-shot metric depth estimation across diverse camera settings. Additionally, Prompt-Depth-Anything \citep{lin2025prompting} demonstrates that a small LiDAR `metric prompt' can steer a FM to accurate, high-resolution metric depth estimation. Similarly, DepthCrafter \citep{hu2025depthcrafter} generates temporally consistent long depth sequences with intricate details for open-world videos.    
    
    \item \underline{Semantic map creation}: VFMs provide robust, semantic-aware features that enhance accuracy, robustness, and informativeness of the generated maps. In particular,  \citet{busch2025one} create reusable open-vocabulary feature maps, capable of supporting probabilistic-semantic updating for informed multi-object exploration. In a similar way, VFM feature representations can be combined with Gaussian-splatting or factor-graph mechanisms for robust long-horizon missions in dynamic environments \citep{zheng2025wildgs,yugay2025magic}. 
    
    \item \underline{Visual-inertial fusion}: VFMs enhance the visual part of visual-inertial odometry (VIO) systems, which is crucial for drift correction and estimation of metric scale. Specifically, features that improve VO \citep{azhari2025dino} or metric priors from depth FMs \citep{hu2024metric3d} are combined with Inertial Measurement Unit (IMU) data in standard VIO estimators. Additionally, depth-foundation priors are injected to stereo/VIO pipelines for stabilizing scale and short-horizon pose, while being complemented by IMU FMs for cross-platform generalization \citep{jiang2025defom,zhao2025tartan}.

    \item \underline{Environment mapping}: VFMs enable map building by producing dense visual embeddings that can be fused into persistent, semantically enriched scene representations. FMGS \citep{zuo2025fmgs} combines FM features with $3$D Gaussian splatting for semantic 3D scene reconstruction and open-vocabulary scene understanding. Similarly, OpenGS-SLAM \citep{yang2025opengs} and FeatureSLAM \citep{thirgood2026featureslam} extend Gaussian-splatting-based mapping with FM-derived semantic features, enabling open-set scene understanding and improving the robustness of real-time tracking and mapping.
\end{itemize}

\subsection{Vision-Language Models (VLMs)}
\label{sup:ssec:vlms_in_robotics}

VLMs combine computer vision with natural language processing capabilities for establishing a coherent, concrete semantic understanding of the world, enabling interpretation and generation of language descriptions of the observed visual entities \citep{clip}. In the context of robotics, VLMs enable robots to simultaneously interpret visual data and natural language commands, allowing for intuitive human-robot interaction and robust task execution in unstructured environments \citep{zhou2025physvlm}. Their main advantages are \citep{tayyab2025foundation}: a) Richer semantic information and environment understanding, where VLMs generate detailed, interpretable semantic outputs that boost robots to handle complex and rare/novel scenarios more efficiently, b) Open-vocabulary object recognition, where the use of the VLMs' shared visual-language embedding space significantly facilitates open-world perception, c) Flexible perception interface, where embodied agents are capable of supporting nuanced queries and generalized reasoning, and d) Enhanced generalization capability, where VLMs allow for efficient handling of novel visual and linguistic combinations. On the contrary, the main limitations of VLMs are \citep{tayyab2025foundation}: a) Inability to define precise actions, where VLMs are capable of interpreting instructions and identifying visual entities, but they lack the intrinsic capability to translate this semantic understanding into precise, executable motor control commands, b) Incomplete semantic grounding, where VLMs exhibit difficulties in connecting abstract, human-like commands to concrete, actionable physical locations and poses required for robot manipulation tasks in real-world environments, and c) Dependency on external policy generators, where VLMs inherently require a dedicated external policy or low-level planner to provide their semantic output for implementing robot actions.

With respect to supported functionalities/operations, VLM-based systems can be classified into the following main categories:
\begin{itemize}
    \item \underline{Manipulation grounding and control signals}: VLMs map high-level, semantic intents into concrete, physically-actionable constraints in the proximity of the robot's operating environment. In particular, OmniManip \citep{pan2025omnimanip} translates VLM reasoning outcomes to object-centric interaction primitives and, subsequently, implements dual closed-loop procedures (namely, planning and execution) to produce precise $3$D spatial constraints. Additionally, RoboGround \citep{huang2025roboground} provides grounded masks for both targets and placement regions into a low-level policy for improving generalization across different types of scenes. KUDA \citep{liu2025kuda} poses queries to a VLM for task keypoints and, subsequently, converts them into optimization costs for model-based planning, enabling open-vocabulary manipulation over rigid, deformable, and granular objects. Moreover, chain-of-modality-style schemes prompt a VLM on human videos and auxiliary signals for extracting step-wise plans and control parameters \citep{wang2025chain}. In a similar way, IKER \citep{patelreal} and ReWiND \citep{zhang2025rewind} refine visually-grounded rewards or costs over long-horizon tasks for stabilizing execution, respectively.
    
    \item \underline{Semantic mapping, referring expressions, and navigation}: VLMs are capable of creating human-readable maps, understanding complex spatial language, and grounding navigation goals in the physical world, materializing the transition from purely geometric to semantic navigation. In particular, One-Map-to-Find-Them-All \citep{busch2025one} creates a reusable open-vocabulary feature map for real-time, zero-shot, multi-object-oriented navigation, while supporting probabilistic semantic updates. Additionally, functional and relational reasoning at scene level is enabled by the use of open-vocabulary functional $3$D scene graphs \citep{zhang2025open} and open-scene graphs \citep{loo2025open} for open-world object-based navigation. OpenVIS \citep{guo2025openvis} incorporates open-vocabulary video instance segmentation and tracking for pursuing tools or novel objects. Moreover, Vision-Language Fly (VLFly) \citep{zhang2025grounded} supports grounded vision-language navigation for Unmanned Aerial Vehicles (UAVs) with open-vocabulary goal understanding, without requiring localization or active ranging sensors.
    
    \item \underline{Execution-time check and progress verification}: VLMs enable robot, semantic, self-monitoring within a closed-loop control framework, materializing the translation of low-level, sensor feedback to high-level, human-understandable checks. Specifically, ExploreVLM \citep{lou2025explorevlm} deploys a closed-loop task planning framework for real-time integration of perception, planning, and execution validation. Additionally, \citet{ahmad2025unified} introduce a unified framework for real-time failure recovery, where a VLM acts as a monitoring tool for verifying pre- and post-conditions of individual skills, inferring missing pre-conditions, and suggesting new skills for recovery. Moreover, Guardian \citep{pacaud2025guardian} enhances the robot abilities to detect manipulation planning and execution errors, by identifying fine-grained failure modes (e.g., object slippage, incorrect action sequencing, etc.).
    
    \item \underline{Closed-loop mobile manipulation}: VLMs provide the necessary cognitive capabilities regarding continuous feedback and adaptation, so as to directly address the inherent long-horizon execution challenge in unstructured, large-scale, and dynamic environments. For instance, COME-robot \citep{zhi2025closed} uses GPT-$4$ for situated reasoning and iterative feedback, in order to recover from failures. Additionally, HomeRobot \citep{yenamandra2023homerobot} relies on an agent that is capable of navigating through household environments for grasping novel objects and placing them on target receptacles.
\end{itemize}

\subsection{Vision-Language-Action models (VLAs)}
\label{sup:ssec:vla_in_robotics}
VLAs aim at integrating multi-modal understanding with direct physical execution, targeting to serve as the basis for autonomous embodied task execution \citep{ma2024survey}. In particular, a VLA model receives multi-modal inputs (typically, vision, language, and robot state) and generates real-world physical actions or control policies in real-time, often designed in an end-to-end way. Their main advantages are \citep{sapkota2025vision}: a) End-to-end implementation and operational simplicity, where VLAs perform the direct mapping from perception signals to control actions, eliminating the need for complex interconnection of multiple, distinct planning modules, b) Robust generalization ability, where VLAs are shown to demonstrate increased generalization performance across different operating environments and robotic platforms, and c) Integration of action and reasoning, where VLAs by design address the multi-modal, semantic grounding problem internally. On the other hand, the main limitations of VLAs are \citep{sapkota2025vision}: a) High demand for large-scale training data, where VLAs require outstanding amounts of high-quality, heterogeneous action data, b) Decreased efficiency across multiple robotic platforms, where unavailability of not always sufficiently-broad robotic training datasets leads to suboptimal task performance across different robotic setups, c) Increased operational latency, where the large-scale nature of the underlying VLA model architectures leads to critical challenges in real-time control settings, and d) Increased complexity in failure management, where errors in end-to-end VLA systems may affect significantly the model's overall behavior.

In terms of supported functionality/operation, VLA-based
systems can be classified into the following main categories:
\begin{itemize}
    \item \underline{Scaling and web-to-robot transfer}: VLAs target the transferring of vast semantic, visual, and common sense knowledge acquired from the internet to the robots' physical-world control policies, while reinforcing generalization across different tasks and embodiments. In particular, RT-$1$ \citep{rt1} scales-up imitation learning capabilities using tokenized action sequences corresponding to language formalized goals, improving robustness in long-tail household task execution. Its successor RT-$2$ \citep{rt2} incorporates web-scale vision-language pretraining in the control pipeline, so that open-vocabulary knowledge transfer to be performed to real-world robots, through the use of a language-aligned visual encoder. Additionally, OpenVLA \citep{openvla} integrates a vision-language encoder and a relatively small action head, while supporting different robotic platforms and maintaining increased zero/few-shot generalization capabilities. Similar approaches incorporate stronger inductive biases and specific $3$D spatial priors \citep{li20253ds}, object-centric adapters for few-shot tuning \citep{li2025controlvla}, standardized cross-embodiment corpora \citep{openx}, multiple low-rank adaptation modules \citep{zhao2025more}, and policy distillation schemes \citep{xu2024rldg}.
    \item \underline{Fusion and action parameterization}: VLAs aim at unifying perception, reasoning, and control by combining a VLM with an action decoder. In particular, GR$00$T N$1$ \citep{bjorck2025gr00t} tightly couples a VLM for interpreting the environment, through vision and language instructions, with a subsequent diffusion transformer for generating motor actions in real-time. Additionally, $\pi_{0.5}$ \citep{intelligence2025pi_} formalizes action generation as a flow matching process on top of a VLM, supporting stable, continuous control and increased generalization at moderate computing requirements. Moreover, similar systems further improve the interaction between the VLM and the action decoder through several complementary design choices, including diffusion models for enhanced dexterity \citep{wen2025dexvla}, unified autoregressive-diffusion heads \citep{liu2025hybridvla}, rectified-flow policies \citep{reuss2025flower}, state-space formulations \citep{liu2024robomamba}, short-horizon video prediction \citep{huvideo}, and embodied reasoning mechanisms \citep{team2025gemini}.
    
    \item \underline{Specialization, adapters, and mixture-of-experts}: VLAs accomplish to leverage massive pre-trained backbones without the need/cost of parsing/running the entire network for every predicted action. Specifically, MoRE \citep{zhao2025more} makes use of sparse Low-Rank Adaptation (LoRA) experts, selecting only a few of them per step for expanding the model's capacity without increased inference cost. Additionally, OpenVLA \citep{openvla} combines a Llama $2$ language model with a visual encoder and implements efficient fine-tuning for new tasks, boosting robust, generalizable policies for visuo-motor control. Moreover, RLDG \citep{xu2024rldg} combines task-specific Reinforcement Learning (RL) with generalist policy distillation for more efficient robotic manipulation.
    
    \item \underline{Navigation and locomotion}: VLAs serve as semantic, navigation planners that translate abstract, natural-language instructions into physically executable movements by mobile and legged robots, often adopting a hierarchical design. In particular, NaVILA \citep{ChengA-RSS-25} initially generates mid-level actions with spatial information in the form of language and, subsequently, utilizes this as input to a visual locomotion RL policy generator for execution. Additionally, VAMOS \citep{castro2025vamos} comprises a hierarchical VLA that decouples semantic planning from embodiment grounding, where a generalist planner learns from diverse, open-world data and a specialist affordance model encodes the robot's physical constraints and capabilities. Moreover, Humanoid-VLA \citep{ding2025humanoid} integrates language-motion pre-alignment (using non-egocentric human motion data paired with textual descriptions), and egocentric visual context (through parameter efficient video-conditioned fine-tuning).
    \item \underline{Operations, deployment, and safety}: VLAs aim at combining their high-level, flexible, and generalized reasoning capabilities with conventional safety guarantees for real-world deployment. Specifically, SafeVLA \citep{zhang2025safevla} formalizes safety alignment as a constraint learning problem, targeting the VLA operation to respect task rules and safety measures, and not relying only on post-hoc filtering. VLATest \citep{wang2025vlatest} comprises a framework designed to generate robotic manipulation scenes for testing VLAs. Moreover, fleet orchestration frameworks combine language interfaces with guardrails and human-in-the-loop supervision, while mechanistic steering implements zero-shot behavior shaping \citep{ahn2024autort,haon2025mechanistic}.
\end{itemize}

\section{Neural network architectures}
\label{sec:nn_architectures_details}

This section provides the full details of the different types of neural network (NN) architectures that are utilized in robotic FM methods, complementing the condensed presentation of Section~\ref{sec:architectures} of the main text. For each architecture type (namely, transformers, state-space models, diffusion models, convolutional and hybrid encoders, and graphical models), the complete category definition, the corresponding main advantages and limitations, as well as the extensive list of representative methods per sub-category are provided below.

\subsection{Transformers}
\label{sup:ssec:architectures_transformers}
In the context of robotics, transformers are widely used for different/diverse tasks (e.g., high-level task planning, low-level policy learning, perception, human-robot interaction, etc.), relying on the fundamental principle of formalizing them as a sequence modeling problem \citep{firoozi2025foundation}. The latter is grounded on converting input data (e.g., states, actions, images, etc.) to numerical tokens, which are then processed as a sequence for estimating a prediction. Their main advantages are \citep{sanghai2024advances}: a) Increased long-range dependency modeling, where transformers enable policies to capture complex, long-range dependencies between states, actions, and rewards over long time horizons, b) Architectural homogenization, where a single, well-understood transformer architecture can be used as the backbone for diverse robotic tasks and input modalities, and c) Increased efficiency in high-level planning and reasoning, where the primary origin of the transformer architecture in language modeling also makes it suitable for high-level, language-conditioned task planning. On the other hand, the main limitations of transformers are \citep{firoozi2025foundation}: a) Increased computational complexity, where the transformer self-attention mechanism compute and memory requirements scale quadratically with the input sequence length; hence, often making pure transformers unsuitable for low-latency, real-time control loops, b) Requirement for discrete tokenization, where transformers inherently operate on discrete tokens, which in turn introduces quantization error and is a fundamentally unnatural representation of physical dynamics, and c) Infinite context contradiction, where a pure-form transformer exhibits outstanding reasoning performance over a finite context, but it is architecturally unsuited for  infinite-horizon, continuous processing required by embodied agents.

With respect to the input data modality, transformer-based
systems can be classified into the following main categories:
\begin{itemize}

    \item \underline{Vision Transformers (ViTs)}:
    ViTs comprise the fundamental architecture of multiple VFMs in robotics. They use self-attention to capture global image relationships, unlike traditional local convolutional methods \citep{vit}. Models like DINOv$2$ use self-supervised learning on web-scale datasets to estimate general-purpose visual features that transfer across tasks without extensive fine-tuning \citep{oquab2023dinov2}. As a consequence, ViTs have a central role in several perception pipelines in robotics. One such key area is scene understanding, which focuses on the immediate, local environment (including $3$D scene understanding) for realizing manipulation, localization, and visual odometry \citep{azhari2025dino,martins2024ovo}.
    Additionally, in the context of semantic mapping, ViT features facilitate the construction/maintenance of long-term, global, semantic maps, supporting open-vocabulary representations for multi-object navigation, SLAM, and targeted exploration \citep{busch2025one,laina2025findanything,martins2025open,jiang2025dualmap,deng2025openvox}. Moreover, ViTs also provide essential geometric cues for low-level control, by generating dense spatial signals (such as zero-shot metric depth and surface normals) from a single image \citep{hu2024metric3d,yang2024depth,guo2025depth}; these signals, along with grounded spatial constraints, support closed-loop execution and grasp planning in manipulation systems \citep{huang2025roboground}.

    \item \underline{Text transformers}: Text transformers act as the language interface, planner, programmer, memory, and supervisor across robotic pipelines, providing a common semantic layer that connects high-level intent to grounded perception and control. In particular, bidirectional encoders (e.g., BERT) learn transferable semantics for downstream modules \citep{devlin2019bert} and decoder-only models (e.g., PaLM \citep{palm} and GPT-$4$ \citep{gpt4}) benefit from scale, improving robotic reasoning and planning performance. Also open-weight LLaMA backbones enable practical on-device deployment \citep{touvron2023llama}. In terms of exhibited functionality, text transformers can translate free-form language into structured plans and formal artefacts, by decomposing long-term goals, filling behavioural trees, and generating PDDL templates that pass feasibility checks \citep{ao2025llm,liu2025delta,zhang2025llm}. Similarly, the same models can generate or adapt policy code from language or video inputs, enabling low-level controllers to remain auditable and closing the gap between high-level intent and executable actions \citep{liang2023code,xie2025robotic,ji2026genswarm}. Additionally, retrieval-augmented setups can ground the planner in external memory and past context, in order to improve long-horizon behaviours, to sustain states across sub-goals, and to reduce drift during environmental changes \citep{mon2025embodied,gu2024conceptgraphs,anwar2025remembr}. In case of action execution divergence from the planned one, text transformers are able to propose corrections, to reason about failed preconditions, and to maintain explicit state estimations to prevent cascading errors/effects \citep{yoneda2024statler,raman2024cape,joublin2024copal}. Moreover, even in generalist vision-language-action frameworks, the textual component remains on top of the formed reasoning layer, efficiently modulating the respective perception and control ones \citep{rt2,team2024octo,bjorck2025gr00t,openvla,team2025gemini}.

    \item \underline{Multi-modal transformers}: These models transform each individual sensor stream into a sequence of tokens, project them to a common/shared latent space, and subsequently align them using a cross-attention or a gated fusion mechanism. The training process combines alignment and generative objectives, so that a single formed backbone network to be capable of understanding scenes, following instructions, and selecting actions \citep{openvla,rt2,palme,team2025gemini, team2024octo,bjorck2025gr00t}. Multi-modal transformers are also capable of combining vision with proprioception information, where the fusion of visual and robot state tokens allows the generation of embodiment-aware policies that can scale across multiple/diverse robots and tasks \citep{wang2024scaling,team2024octo,rt2}. An additional capability comprises the enrichment of vision with geometric cues, where the integration of metric depth and surface normals can strengthen mapping and grasp planning with dense geometry that can generalize to novel scenes without requiring task-specific labels \citep{hu2024metric3d,yang2024depth,huang2025roboground}. Moreover, multi-modal transformers enable the generation of unified tactile-vision embeddings that can boost reasoning capabilities regarding contact, texture, and stability; hence, improving manipulation under uncertainty and occlusion \citep{yang2024binding,feng2025anytouch}. On another direction, joint audio-visual tokens can support navigation in noisy, multi-source settings and improve robustness in the presence of distractions \citep{shi2025towards,parkentropy}. Furthermore, transformer-based fusion of thermal and RGB information streams can improve perception and localization, especially under low light and adverse weather conditions \citep{puttagunta2024multi,skorokhodov2026sear}. More recently, lightweight adapter modules are widely used for enabling the incorporation of additional sensors without retraining the full model, which comprises a common pattern shared across VLA and heterogeneous pre-training schemes \citep{openvla,wang2024scaling,team2025gemini}.
\end{itemize}

\subsection{State-space models}
\label{sup:ssec:architectures_ssm}
SSMs are increasingly adopted in robotics for realizing real-time control and long-horizon reasoning, by treating the sensorial input streams as latent states and subsequently producing output predictions in a step-by-step way \citep{liu2024robomamba}. The latter is in practice performed by learning end-to-end system matrices with diagonal-plus-low-rank parameterizations and hardware-aware modeling \citep{gu2023mamba,daotransformers}. Their main advantages are \citep{gu2021efficiently,smith2023simplified}: a) Linear scaling of complexity, where computation and memory requirements grow linearly with sequence length, b) Stable long-horizon memory, where the latent state maintains temporal context without large activation caches, and c) Deployment efficiency, where low latency and steady throughput are suitable for embedded robotic solutions. On the contrary, the main limitations of SSMs are \citep{lieber2024jamba}: a) Reduced cross-token check, since explicit/direct correlation between tokens is not inherently supported, b) Reduced global context modeling, compared to full attention-based counterparts, and c) Frequent need for hybrid designs, which typically integrate attention or retrieval blocks for implementing tool use and long-range planning.

With respect to the input modalities, SSM-based systems can be classified into the following main categories:
\begin{itemize}
    \item \underline{Visual SSMs}: Visual SSMs can replace attention mechanisms with corresponding selection-based ones, resulting into linear-complexity encoders that can be plugged into recognition, dense prediction, and tracking heads \citep{liu2024vmamba,xiao2025spatial}. Additionally, scene understanding and object tracking can benefit from long temporal context modeling at constant cost, which is particularly suitable for long video streams and multi-camera setups \citep{park2024videomamba}. Moreover, event-driven perception can be boosted to handle irregular sampling and rapid environmental dynamics, which is beneficial for agile navigation and manipulation in low light or high motion settings \citep{zubic2024state}.

    \item \underline{Policy/control SSMs}: SSMs can be used for realizing data-driven nonlinear reduction in complex systems, such as modeling hysteresis and memory effects. In particular, table-top manipulation can adopt region-aware selective-state policies with flow-matching objectives to learn precise, real-world skills from limited demonstrations \citep{wang2025flowram}. Additionally, hybrid selective-state diffusion policies may reduce parameters while maintaining performance, in order to improve sample efficiency under multi-view inputs and long horizons \citep{cao2024mamba}. Moreover, vision-driven locomotion can incorporate depth and proprioception information, through stacked selective-state formalisms and end-to-end reinforcement learning \citep{wang2025locomamba}.

    \item \underline{Multimodal SSMs}: A single SSM trunk can be simultaneously pretrained on long videos, robot logs, and demonstrations, in order to support perception, planning interfaces, and action heads. In particular, SSM-based VLA models can fuse vision, language, and proprioception inputs as token streams and, subsequently, generate actions with lower latency and memory than attention-only implementations \citep{tsuji2025mamba}, which makes the application of long-horizon policies more efficient and practical \citep{liu2024robomamba}. When explicit/direct lookup, tool use, or long-range queries are required, attention or retrieval layers can be integrated on top of the main SSM model \citep{lieber2024jamba,wang2025mamba}.
\end{itemize}

\subsection{Diffusion models}
\label{sup:ssec:architectures_dms}
DMs can generate robot behaviours by reversing a gradual noising process, where a forward pass increasingly adds Gaussian noise to the input data and, subsequently, a learned reverse model denoises back to the original input space (e.g., actions, trajectories, sub-goals, etc.) \citep{ho2020denoising,songdenoising}. In particular, DMs implement control policies as a conditional denoising process and have been shown robust across different manipulation tasks \citep{chi2025diffusion}, while they can also serve as generative heads on top of pretrained vision and multimodal backbones \citep{kapelyukh2024dream2real,zeng2024lvdiffusor}. Their main advantages are \citep{liang2025diffusion}: a) Multi-modal feature modelling, where sampling can produce diverse, uncertainty-aware candidates that can facilitate planning under partial observability and contact variability \citep{janner2022planning,chi2025diffusion}, b) Composite conditioning, where a single denoising process can employ frozen vision, language, depth, and proprioceptive backbones to modulate goals and action chunks without task-specific labels \citep{kapelyukh2024dream2real,zeng2024lvdiffusor,ze20243d}, and c) Trajectory-level decision making, where value or constraint guidance can steer full roll-outs towards safe and feasible plans, while improving long-horizon behavior \citep{janner2022planning}. On the contrary, the main limitations of DMs are \citep{wolf2025diffusion}: a) Sampling latency and energy cost, stemming from the inherent iterative denoising process \citep{dong2024diffuserlite}, b) Lack of built-in safety guarantees, where constraint checks or guided objectives are needed to avoid collisions and dynamics violations \citep{janner2022planning,wolf2025diffusion}, and c) Sensitivity to conditioning drift, where errors/noise in visual or language features can mislead the sampling process \citep{chi2025diffusion,ze20243d}.

With respect to the conditioning type, DM-based
systems can be classified into the following main
categories:
\begin{itemize}

\item \underline{Vision-conditioned DMs}: DMs can translate visual goals into usable, structured information for control purposes. In particular, image-goal generation and rearrangement of priors can estimate object- and scene-level targets that downstream controllers can subsequently follow \citep{kapelyukh2023dall,zeng2024lvdiffusor}. Additionally, pretrained image-editing DMs can generate sub-goal images from language instructions and current camera views, guiding goal-conditioned policies in real-world settings \citep{black2023zero}. Moreover, compact $3$D visual tokens can be employed for improving spatial grounding and robustness across different viewpoints for manipulation planning \citep{chi2025diffusion,ze20243d,kapelyukh2024dream2real}.

\item \underline{Proprioception-, force-, and haptic-conditioned DMs}: Visuomotor diffusion policies treat action sequences as denoised samples conditioned on images and robot states, which facilitates in handling multi-modal actions and improving stability for manipulation tasks \citep{chi2025diffusion}. When force or contact requirements are present, conditioning on haptics and force signals can be incorporated, for example, in visual-tactile slow-fast policies for contact-rich skills \citep{shukla2025learning,xue2025reactive}. In order to maintain control loops short, progressive refinement can increase prediction rates up to real-time performance \citep{dong2024diffuserlite}.

\item \underline{Language-conditioned DMs}: Textual inputs can serve as a guiding signal, where the denoising mechanism can modulate goals, trajectories, or sub-goals, simplifying task setup and execution \citep{bjorck2025gr00t}. Additional works demonstrate the growing use of language prompts for manipulation and planning tasks based on diffusion backbones \citep{wolf2025diffusion,liang2025diffusion}.

\item \underline{Human behaviour-conditioned DMs}: Diffusion objectives can be defined so that early-stage human motion detection can facilitate the accurate prediction of intent, improving intuitiveness and comfort in human-robot interaction without changing the controller structure. In this context, the Legibility Diffuser \citep{bronars2024legibility} demonstrates that a policy trained on offline demonstrations can generate intent-expressive collaborative motions that humans find easier to understand, while still completing a given task more efficiently \citep{ng2023diffusion}.

\end{itemize}

\subsection{Convolutional and hybrid encoders}
\label{sup:ssec:architectures_cnns}
Visual encoders typically comprise the main perception module of any robotic solution, translating raw pixel data into latent representations that a robot can use for subsequent planning procedures \citep{nair2022r3m}. The selection between CNNs and hybrid CNN-transformer implementations essentially comprises a decision on the trade-off between local spatial precision and global context modeling, respectively. Their main advantages are \citep{rt1, tan2019efficientnet}: a) Zero-shot generalization, where due to the fact that the models are trained on internet-scale datasets, previously unseen objects can often be recognized, b) Robustness to noise, exhibiting increased resilience to changes in lighting, shadows, or cluttered backgrounds, and c) Reduced need for training data, where `frozen' pre-trained encoders often exhibit increased performance, without re-training. On the contrary, their main limitations are \citep{mavip}: a) High computational latency, due to the typical high-scale of FMs, b) Loss of fine-grained details, where encoders often divide processing of images in patches for reducing memory requirements, c) Prone to distribution shifts, where there might be a significant discrepancy between the internet-scale training data and the specific application images, and d) Lack of temporal consistency, due to many visual encoders processing video streams in an independent frame-by-frame fashion.

With respect to the encoder and integration type, the following main categories can be identified:
\begin{itemize}

\item \underline{CNN encoders}:
CNNs excel at capturing low-level spatial details like edges, textures, and object boundaries, due to their local receptive fields. In particular, ResNet-series encoders are used for high-fidelity state representations in diffusion-based exploration tasks in \citep{cao2025dare}. Additionally, EfficientNet-B3 is employed as a visual encoder to facilitate real-time, goal-conditioned navigation and exploration in \citep{sridhar2024nomad}. Additionally, R$3$M \citep{nair2022r3m} freezes a ResNet-50 trained on Ego$4$D for improving manipulation for both simulation and real-world scenarios. Moreover, language-reasoning segmentation masks generated by internet-scale trained encoders are leveraged to condition robot manipulation tasks \citep{yang2025transferring}.

\item \underline{CNN-transformer hybrids}: Hybrid architectures often combine a CNN for handling pixel-level information and a transformer one for addressing context and action aspects. In particular, RT-$1$ \citep{rt1} encodes frames with a FiLM-conditioned EfficientNet \citep{tan2019efficientnet}, compresses them with TokenLearner, and then predicts discrete actions using a transformer, enabling real-world task control. Additionally, BC-Z/PaLM-SayCan \citep{jang2022bc,saycan} employ a lightweight ResNet coupled with a shallow attention network for supporting instruction-conditioned policies. In a similar way, diffusion-transformer policies may also adopt convolutional components as image tokenizers prior to respective transformer layers \citep{dasari2025ingredients}.

\item \underline{CNN tokenizers inside generalist agents}: Generalist agents typically convert high-resolution images into compact tokens, prior to sequence modeling. In particular, Gato \citep{gato} employs a small ResNet image tokenizer and feeds visual, text, and proprioception information to a single transformer for controlling multiple skills. RoboCat \citep{bousmalis2023robocat} makes use of a pretrained VQ-GAN image tokenizer and a transformer network, in order to adapt across robots and tasks according to a sequence of self-improvement cycles. Moreover, CNN tokenizers achieve to maintain low information bandwidth and to preserve an efficient interface to large sequence models for various navigation and manipulation tasks \citep{shah2023vint}.

\item \underline{CNN-conditioned diffusion policies}: Diffusion policies often condition a temporal U-Net on CNN features for generating action chunks that are diverse, yet feasible. In particular, Diffusion Policy \citep{chi2025diffusion} employs ResNet-$18$ features for robot manipulation, while preserving low added latency. DiffuserLite \citep{dong2024diffuserlite} demonstrates that progressive refinement with a frozen MobileNet-V$3$ encoder can increase prediction rates towards real-time performance on embedded platforms. Additional works concentrate on similar CNN-based diffusion frameworks for imitation and reinforcement learning purposes \citep{liang2025dreamitate,dasari2025ingredients}.

\end{itemize}

\subsection{Graphical models}
\label{sup:ssec:architectures_graphs}
In the context of robotic FM methods, the use of graphs introduces additional capabilities towards the goal of connecting low-level, raw sensorial data and high-level, structured reasoning \citep{maggio2024clio,booker2024embodiedrag}. Unlike other NN architectures that process input data as matrices (e.g., images) or sequences (e.g., text tokens), graphs enable the modeling and interpretation of the surrounding environment as a set of interconnected entities. For example, scene graphs often decompose the visual world into nodes (e.g., objects, parts, or agents) and edges (e.g., spatial, semantic, or functional relationships), essentially aiming at modeling their inter-dependencies. Their main advantages are \citep{gu2024conceptgraphs}: a) Combinatorial generalization, where the learning of relationships among entities (instead of specific instances) can boost the generalization to previously unseen cases, b) Permutation invariance, due to the inherent ability of graphs to learn the structure (and not the order) of the data, c) Sample efficiency, which derives from the strong inductive bias natively incorporated in a graph model, and d) Increased explainability, due to the improved efficiency in interpreting the reasoning process of a graphical model. On the contrary, their main limitations are \citep{rana2023sayplan}: a) Computational overhead, where the complexity of the message-passing algorithm in large graphs can introduce critical latency for real-time control applications, b) Dynamic topology, which relates to the mathematical difficulty in modeling real-world, dynamic environments in a stable way, and c) Need for integration with latent spaces, where graphs typically need to connect and to operate on precomputed, high-dimensional vector spaces.

With respect to the graph and function type, the
following main categories can be identified:
\begin{itemize}

  \item \underline{Scene graphs}: Open-vocabulary $3$D scene graphs associate vision-language features with real-world entities (often in a hierarchical way), while remaining compact compared to dense map representations. This structure enables robots to query targets, to reason about relationships, and to define sub-goals to planners in large-scale environments \citep{gu2024conceptgraphs,rana2023sayplan,yan2025dynamic}. More recently, graphs are constructed online directly from RGB-D streams, using hierarchical structures for language-grounded navigation and adding functional links to the incorporated entities \citep{werby2024hierarchical,yin2024sg,zhang2025open}.

  \item \underline{Shared graphs}: Compressed-form scene graphs allow bandwidth-limited sharing and map merging, while maintaining open-vocabulary query capabilities. In particular, decentralized visual FMs can estimate peer poses and can produce local Bird’s-Eye View (BEV) maps on embedded hardware, while reducing communication requirements without losing key semantic information \citep{blumenkamp2025covis,gu2025mr}. Such designs make robot-team perception and planning feasible in large-scale environments.

  \item \underline{Graph neural networks}: Graph Neural Networks (GNNs) enable message passing over task, object, and agent graphs for performing allocation, scheduling, and policy conditioning in a data-driven way. Recent hybrid, cognitive pipelines couple GNN-based scene graphs with LLM or symbolic planners, achieving to maintain plans physically feasible, while at the same time still following language goals \citep{tong2026gnn,strader2025language}.

  \item \underline{Embodiment graphs}: Embodiment graphs encode robot joint information, as well as links between them, allowing a single learned policy to adapt across different platforms. In this respect, attention or message passing algorithms follow the learned graph connectivity, which in turn boosts zero-shot transfer to new morphologies and supports reusable controllers across different robots \citep{patel2025get}.

\end{itemize}

\section{Learning paradigms}
\label{sec:learning_paradigms_details}

This section provides the full details of the different learning paradigms that are utilized in robotic FM methods, complementing the condensed presentation of Section~\ref{sec:learningParadigm} of the main text. For each learning paradigm (namely, supervised learning, self-supervised learning, fine-tuning, domain adaptation, imitation learning, reinforcement learning, in-context/prompt learning, world model learning, and generative learning), the complete category definition, the corresponding main advantages and limitations, as well as the extensive list of representative methods per sub-category are provided below.

\subsection{Supervised learning}
\label{sup:ssec:paradigms_supervised}
Supervised learning in robotic FM development trains a model on large-scale, labeled datasets, directly fitting a mapping from inputs (e.g., images, language instructions, and proprioceptive states) to target outputs (e.g., class labels, tokens, or action commands) under explicit human-provided supervision. Its main advantages are \citep{xiao2025robot}: a) High task accuracy, since explicit targets provide a direct and well-defined optimization signal, b) Stable and efficient optimization, where labeled objectives yield reliable convergence, c) Strong semantic transfer, where supervision on internet-scale labeled corpora injects rich semantic priors that transfer to downstream tasks, and d) Straightforward evaluation, where labeled targets make performance directly measurable. On the contrary, its main limitations are \citep{rt2}: a) Dependence on costly human annotation, which is difficult to scale for embodied robot data, b) Limited coverage, where supervised models generalize poorly beyond the distribution of the labeled set, c) High computational cost when paired with internet-scale corpora, and d) Safety concerns, where the presence of hallucinations in the model behavior leads to lacking of formal safety guarantees. In this context, PaLM-E \citep{palme} is jointly trained using web-scale labeled multi-modal data and embodied experiences, while RT-$2$ \citep{rt2} and Gemini Robotics \citep{team2025gemini} couple web-scale vision-language supervision with labeled robot manipulation data for open-vocabulary, instruction-following control.

\subsection{Self-supervised learning}
\label{sup:ssec:paradigms_ssl}
SSL techniques employ a set of `pretext tasks' (e.g., predicting the next video frame, reconstructing a masked image patch, etc.) and large quantities of unlabeled data streams (e.g., visual, depth, force, audio, robot logs, etc.) for enabling FMs to acquire common sense knowledge regarding physics, object permanence, and spatial relationships \citep{he2022masked,tong2022videomae,oquab2023dinov2}. Their main advantages are \citep{nair2022r3m,wu2023daydreamer,assran2025v}: a) Massive scalability, where robots can learn from internet-scale datasets, without requiring human supervision, b) Sample efficiency, where only a handful of training samples is needed for adapting to new tasks, c) Zero-shot generalization, where SSL-trained models often exhibit increased performance in novel settings, and d) Autonomous improvement, where SSL methods can infer reward signals for training directly from the data itself. On the contrary, their main limitations are \citep{nair2022r3m,wu2023daydreamer}: a) Embodiment gap, where most large-scale SSL-employed sources lack proprioceptive data (e.g., joint torques and forces), b) High computational cost, which typically requires massive GPU resources, c) Presence of hallucinations, where the learned models may infer physically impossible actions, and d) Evaluation difficulty, where it is inherently challenging to measure the success of SSL-learned representations, until actual model deployment.

The most common SSL techniques used for developing robotic FM solutions are:

\begin{itemize}
  \item \underline{Masked Autoencoder (MAE)}: The model learns by reconstructing missing or corrupted parts of the input data, aiming at modeling robust spatial and temporal features. Image MAE \citep{he2022masked} and its video counterpart VideoMAE \citep{tong2022videomae} are widely used for producing robust visual backbone networks. Additionally, robotics-specific masked-pretraining approaches, such as $3$D-MVP \citep{qian20253d}, adapt the MAE paradigm to robot learning for manipulation.

  \item \underline{Contrastive learning}: The fundamental principle relies on aligning matched views (and separating mismatched ones) for creating discriminative features and supporting open-vocabulary grounding, so that robots can link language to perception and retrieval skills. CLIP \citep{clip} establishes vision-language alignment at scale, while egocentric robot features, like R$3$M \citep{nair2022r3m}, can improve manipulation sample-efficiency using recorded human-performing videos.

  \item \underline{Autoregressive sequence modeling}: The ultimate goal is grounded on predicting the next token in vision, language, or action streams, in order to capture long-horizon structure and to enable unified perception-to-policy modeling. Generalist agents, such as Gato \citep{gato}, demonstrate how one sequence model can condition on images, text, and proprioception to produce actions across many tasks. Additionally, LLMs, like GPT-$3$ \citep{brown2020gpt3} and GPT-$4$ \citep{gpt4}, showcase the scalability and cross-domain reasoning capabilities of autoregressive transformers in embodied pipelines \citep{turcato2025towards,mon2025embodied}.

  \item \underline{World model learning}: Also often reported as an individual learning paradigm (Section \ref{sup:ssec:worldmodel}), WMs aim at learning to predict how the world will change in response to specific actions, essentially encoding causal relations. Various individual WM-based methods are introduced for varying environmental settings, including Dreamer-style agents on physical robots \citep{wu2023daydreamer}, JEPA-based designs (such as AdaWorld \citep{gao2025adaworld} and ACT-JEPA \citep{vujinovic2025act}), and compositional video world models (like RoboDreamer \citep{zhou2024robodreamer}).

\end{itemize}

\subsection{Fine-tuning}
\label{sup:ssec:paradigms_fine-tuning}
Fine-tuning aims at adapting the internet-scale acquired knowledge to specific physical-world execution settings. In particular, it targets to adjust the rich, common-sense knowledge structures of a pretrained FM to a specific robot, sensor suite, or task, making use of a smaller, in-domain, annotated dataset. Its main advantages are \citep{yu2025survey}: a) Sample efficiency, where only a relatively reduced set of training data is needed for the new/specific robot task, b) Domain adaptation, allowing the pretrained FM to handle different/specific application settings that are not present in the original training set, and c) Improved precision, where the learned policies are enabled to become more accurate and robust. On the contrary, its main limitations are \citep{team2024octo}: a) Catastrophic forgetting, related to the risk of the FM losing part of its general-purpose skills, b) Overfitting, which may occur when the fine-tuning dataset is relatively small and the FM may extensively adapt to specific operational settings, and c) Need for annotated data, where the requirement for high-quality robotic data is still present.

The most common fine-tuning techniques in robotics are:
\begin{itemize}
    \item\underline{Full fine-tuning}: The goal is to update all FM weights using a relatively small, in-domain robot dataset, in order to re-target a pretrained policy to a new robot, sensor setup, or task. Recent generalist policies report fast adaptation to new observation and action spaces on standard GPUs, using full fine-tuning as the baseline learning method \citep{team2024octo,openvla}.

    \item\underline{Low-Rank Adaptation (LoRA)}: The fundamental aim relies on maintaining the original FM weights unchanged and adding two low-rank matrices that learn/incorporate the required model modifications. In particular, OpenVLA \citep{openvla} demonstrates LoRA-based tuning on the large-scale Open-X-Embodiment dataset \citep{openx}. Additionally, more recent approaches (including quantized LoRA versions) target model adaptation in resource-constrained robotic platforms \citep{williams2026litevla,kim2026adaptive}.

    \item\underline{Quantized parameter-efficient fine-tuning}: This extends the original Parameter-Efficient Fine-Tuning (PEFT) approach, by combining low-precision weights with LoRA-style adapters for maintaining latency and memory requirements low on embedded or field hardware, while at the same time retaining most of the full-precision performance. In particular, LiteVLA \citep{williams2026litevla} and similar approaches \citep{williams2025lite} report that an $4$-$8$ bit quantization plus adapters can preserve high recognition rates, while enabling real-time control on smaller platforms.

    \item\underline{Action space remapping}: This relies on integrating lightweight encoders/decoders or tokenizers, so that the core policy can be fine-tuned to new sensors or actuators without retraining the FM from scratch. Generalist policies, such as Octo \citep{team2024octo} and RT-$2$ \citep{rt2}, rely on this technique to re-target a given FM to multiple robots and grippers with modest additional data.

\end{itemize}

\subsection{Domain adaptation}
\label{sup:ssec:paradigms_domain-adaptation}
Domain Adaptation (DA) aims at bridging the critical gap between FMs being pre-trained on massive internet-scale data and their subsequent deployment in real-world environments. In practice, DA targets to equip FMs with the required physical-world grounding or specific environmental awareness capabilities for given application scenarios, which is typically termed as the `sim-to-real' transfer challenge. Its main advantages are \citep{da2025survey}: a) Reduced need for real-world samples, where the use of simulation data accommodates the need for extensive amounts of real-world ones, b) Robustness to sensor shifts, where models are boosted to remain stable in case of robot hardware changes, and c) Safety guarantee, where, since training is performed in simulation environments, the risk of robots damaging their surroundings during the learning phase is reduced. On the contrary, its main limitations are \citep{tayyab2025foundation}: a) Risk of negative transfer, which may occur when the difference between the simulation and the real-world environments is large, b) Training instability, where the process of defining multiple hyper-parameters in the simulation environment renders the whole training process sensitive to their selection, and c) Lack of predictability, where it is difficult to assess the real-world situations where the trained robot might succeed or fail.

The most common DA techniques in robotics are:
\begin{itemize}

  \item \underline{Sim-to-real transfer}: This aims at training a FM model using large corpora of simulated/synthetic data, evaluating their performance in such simulation suites, and eventually calibrating them on real-world hardware (typically employing some real-world data). In particular, humanoid and manipulation solutions demonstrate the efficiency of this approach \citep{luo2026simvla,deng2025graspvla}.

  \item \underline{Real-to-sim-to-real transfer}: This scenario involves the replay of real-world robot trajectories in high-fidelity simulations, the diversification of the depicted scenes and objects, the synthesis of new training data, the evaluation of the developed models in simulation (often employing domain randomization techniques), and the eventual calibration of the FM model to the real-world specifications. Various works demonstrate the validity of this approach in diverse operational settings \citep{zhu2025vr,fang2025rebot}.
\end{itemize}

\subsection{Imitation learning}
\label{sup:ssec:paradigms_imitation-learning}
The fundamental consideration of Imitation Learning (IL) relies on enabling a model to learn directly from a (human) expert via teleoperation or video demonstrations of the desired skills. The main usefulness of IL lies on its efficient multi-modal alignment, which allows the mapping of high-level language instructions and/or visual inputs directly to low-level motor commands. Its main advantages are \citep{zare2024survey}: a) No reward signal definition, where the robot is only required to mimic the expert behavior, b) Reduced training data, where IL is shown to require a reduced number of expert demonstrations to achieve robust performance, and c) Easy training supervision, which is performed directly through teleoperation and/or expert demonstrations. On the contrary, its main limitations are \citep{kawaharazuka2025vision}: a) High dependency on data quality, where the quality of the observed demonstrations has a direct impact on the robot learning process, b) Covariate shift, where the robot is highly likely to fail if its actions deviate from the observed demonstrations, and c) Causal confusion, where the model might incorrectly learn unintended factors being present in the demonstrations.

The most common IL techniques in robotics are:
\begin{itemize}
  \item \underline{Behavioral cloning}: This serves as a particular type of Supervised Fine-Tuning (SFT), by imitating human expert demonstrations. In particular, RT-$1$ \citep{rt1} learns a direct mapping of observations and language goals to actions, using large-scale transformer networks trained on diverse demonstration data. RT-$2$ \citep{rt2} extends this with web-scale vision-language pretraining to open-vocabulary, instruction-following control. Moreover, vision FMs for embodiment- and environment-agnostic scene representation further decouple perception from control to facilitate cross-robot transfer \citep{riou2024vision}.

  \item \underline{Diffusion-based IL}: The fundamental consideration relies on representing a robot's behavior as a conditional denoising process. In this way, diffusion policies treat actions as a data distribution that can be iteratively refined from random noise. In particular, Diffusion Policy \citep{chi2025diffusion} and Diff-Dagger \citep{lee2025diff} generate action sequences that match expert behavior and handle multi-modal inputs, while improving stability for long-horizon manipulation.

  \item \underline{In-context IL}: The main goal is for the robot to learn novel tasks on the fly, i.e., to enable zero- or few-shot task adaptation, but without updating the model weights. ICRT \citep{fu2025icrt} instantiates this idea by using next-token prediction over sensorimotor streams for real-robot in-context imitation, extending a similar previous approach that is based on a sequence modeling formalism. Similarly, prompt demonstrations are augmented using explicit visual reasoning traces \citep{nguyen2026iclr}, allowing the model to infer task intent more reliably in complex and ambiguous environments, while jointly predicting reasoning and low-level actions in an autoregressive way.
  \item \underline{Continual IL}: This focuses on addressing the long-term memory and evolution challenges of robotic FMs. The core functionality is to enable a robot to sequentially acquire new skills over time, without forgetting previously learned ones. In this context, LOTUS \citep{wan2024lotus} introduces a continual imitation learning framework for skill acquisition by a real robot.
\end{itemize}

\subsection{Reinforcement learning}
\label{sup:ssec:paradigms_reinforcement-learning}
Reinforcement Learning (RL) serves as the optimization formalism that aims at bridging the gap between high-level, semantic reasoning (supported by FMs) and low-level, physical robot actions. In practice, the goal of RL is to involve the FM in a continuous, self-improvement cycle of subsequent perception, action, and evaluation steps. Its main advantages are \citep{tang2025deep}: a) Self-improvement capability, where RL allows a robot to explore and to surpass the quality of its training data, b) Increased generalization ability, where RL encourages the model to investigate alternative strategies, making it more robust to novel situations, and c) Efficient sim-to-real implementation, where the RL component can be extensively trained in simulation, prior to be deployed in real operational settings. On the contrary, its main limitations are \citep{ter2025taxonomy}: a) Sample inefficiency, where RL often requires a very large number of training trajectories to converge, b) Careful reward engineering, which may require detailed definition of the reward function for avoiding misleads during training, and c) Difficulty in credit assignment, which corresponds to the inherent challenge of identifying incorrect robot behaviors over long-horizon tasks.

The most common RL techniques in robotics are:
\begin{itemize}
  \item \underline{SFT-to-RL}: This involves a two-stage process, where BC is initially applied for learning a policy and subsequently RL is employed for further improving it. In this context, RT-$1$ \citep{rt1} and RT-$2$ \citep{rt2} demonstrate how large-scale BC may produce powerful priors that can be refined further through interaction. Additionally, ExploRLLM \citep{ma2025explorllm} combines an LLM-guided exploration policy with a residual RL head to improve sample efficiency and robustness.

  \item \underline{LLM-guided reward design}: This leverages the capabilities of LLMs for writing/refining the RL reward code and tuning domain randomization procedures, in order to improve the robustness of learning and knowledge transfer. In particular, Eureka \citep{ma2023eureka} automates reward design and outperforms expert rewards on multiple robotic tasks, while DrEureka \citep{ma2024dreureka} extends this approach to the sim-to-real setting, by jointly optimizing rewards and randomization for locomotion and dexterous manipulation. Moreover, Gen2Sim \citep{katara2024gen2sim} increases the application scope, by using generative models and LLMs to synthesize tasks, scenes, and reward functions for large-scale RL in simulation.

  \item \underline{Preference-based RL}: This relies on replacing detailed/numeric RL rewards with preferences produced by VLMs (or adapted ones, by involving small-scale human intervention). RL-VLM-F \citep{wang2024rl} models rewards from VLM comparisons over image observations and task-related text, in order to improve manipulation without human guidance. Additionally, VARP \citep{singh2025varp} regularizes VLM-derived preferences with the agent's own rollouts for reducing misalignment and hallucinations in vision-language feedback.

  \item \underline{Offline-to-online RL}: This involves a $3$-step process, where a) The model is initially trained offline on massive, heterogeneous datasets, b) It subsequently undergoes an offline RL refinement step using an appropriate reward function, and c) Eventually, it follows online RL for real-world deployment. Indicatively, embodied visual tracking is combined with a text-promptable encoder and offline RL for improved perception in \citep{zhong2024empowering}. Additionally, FLaRe \citep{hu2025flare} applies large-scale RL fine-tuning on a pre-trained VLA for adaptive manipulation across diverse tasks.

  \item \underline{World-model RL}: World-Model Reinforcement Learning (WM-RL) aims at learning a generative world model with language-aware structure and, subsequently, using it for RL-based policy improvement. In this respect, RoboDreamer \citep{zhou2024robodreamer} factors video generation into compositional parts, conditioned by language and visual goals, and exhibits robust performance on long-horizon tasks.

\end{itemize}

\subsection{In-context/prompt learning}
\label{sup:ssec:paradigms_in-context-learning}
ICL and prompt learning enable the adaptation of FMs at inference time by conditioning their behavior on demonstrations, examples, or task-specific instructions, without requiring updates to the underlying model weights. While closely related, the two are not identical: ICL relies on task-relevant examples provided within the context window, whereas prompt learning more broadly focuses on steering a pre-trained model through textual or multimodal instructions. Their main advantages are \citep{fu2025icrt,yin2024context}: a) Generalization to novel settings, where a small number of demonstrations or prompts can enable adaptation to previously unseen tasks and environments, b) High-level task planning, where natural-language guidance can be translated into structured reasoning steps and, subsequently, into low-level physical actions, and c) Multimodal task specification, where context can be expressed through language, visual observations, or sensorimotor demonstrations. On the contrary, their main limitations are \citep{yao2023react}: a) Prompt sensitivity, where even minor changes in the input context may lead to unstable or suboptimal behavior, b) Limited grounding, where inference-time reasoning does not inherently guarantee consistency with real-world constraints, and c) Inference overhead, where long context windows and multi-step prompting strategies may increase latency and computational cost during deployment.

The most common in-context/prompt learning techniques in robotics are:
\begin{itemize}
  \item \underline{Language prompting}: This relies on the use of natural language instructions for guiding a robot's behavior, decision-making, and physical actions. In particular, few-shot language prompts can encode demonstrations or templates, so that an LLM can output low-level actions or to acquire new skills \citep{yin2024context,liang2023code,huang2023instruct2act}.

  \item \underline{Reason-act prompting (ReAct)}: This interleaves natural language reasoning with physical actions, allowing a robot to decompose complex goals, to validate its progress, and to dynamically adjust its plan. In this respect, planning, execution, and re-planning can be performed in a single loop, also involving an LLM-based verification checking step \citep{yao2023react,grigorev2025verifyllm}.

  \item \underline{In-context imitation}: This enables a FM to perform a novel task by observing a few videos or sensorimotor demonstrations, without applying any permanent changes to its internal weights. In this context, a causal policy can parse short teleoperation trajectories as a prompt and, subsequently, to predict the next action for new tasks without the need for fine-tuning \citep{fu2025icrt}.
\end{itemize}

\subsection{World model learning}
\label{sup:ssec:worldmodel}
The fundamental functionality of World Models (WMs) in robotic FM applications is that they allow robots to predict environmental changes in response to their actions. In particular, their primary role is to decouple perception from action, where, instead of directly operating only on pixel values, the robot can learn the underlying physics of the world. Their main advantages are \citep{li2025comprehensive}: a) Incorporation of `imagined' experiences, which reduces the need for physical-world trials, b) Modeling rules of physics, where the model learns the impact of factors like gravity, friction, and collisions in the real environment, and c) Reduced delays in operation, where the robot is able to predict future states and to maintain smooth actions. On the contrary, their main limitations are \citep{zhang2025step}: a) Presence of hallucinations, where the robot may converge to misleading actions in case of slight inaccuracies in the WM, b) Cumulative errors, where small prediction errors can be accumulated in long-horizon tasks, and c) Increased computational cost, where the training of a high-fidelity WM may require excessive GPU resources.

The most common WM learning techniques in robotics are:
\begin{itemize}
  \item \underline{Feature-space WMs}: Instead of performing a prediction for each pixel, which is computationally expensive and often noisy, feature-space WMs map visual inputs to an abstract feature space and perform future predictions there. In particular, future DINOv2 patch embeddings are predicted from offline trajectories and, subsequently, action sequences are optimized in the embedding space for zero-shot planning \citep{zhou2025dino}.

  \item \underline{Latent-action WMs}: Latent-action WMs aim at learning robots to model the underlying physics and intent of actions, instead of focusing on representing specific skills. In this respect, continuous latent actions are discovered from videos, while an auto-regressive WM is trained that conditions on those actions to transfer skills across scenes and embodiments with small-scale finetuning \citep{gao2025adaworld}.

  \item \underline{Compositional video WMs}: These adopt a modular approach, where the model aims at breaking down (factorizing) the surrounding environment into its constituent parts (namely, objects, relationships, and action primitives) and, subsequently, reconnecting them for generating future scenarios. In this context, videos are factorized into objects and relations so that the model can synthesize plans for unseen combinations of goals-scenes and guide long-horizon decisions \citep{zhou2024robodreamer}.

  \item \underline{JEPA-style WMs}: The main focus lies on predicting an abstract meaning of what will happen next, which enables robots to plan complex tasks, without getting distracted by irrelevant noise. In particular, joint predictions of short-horizon actions and abstract observations are realized \citep{vujinovic2025act}, in order to couple imitation with predictive learning and to reduce control error accumulation.
\end{itemize}

\subsection{Generative learning}
\label{sup:ssec:paradigms_generative-learning}
Generative Learning (GL) enables robots to imagine future states, to synthesize training data, and to propose complex action sequences. Its fundamental use lies on the capability of generative FMs of producing large quantities of data samples, alleviating from the need for extensive high-quality robotic interaction samples. Its main advantages are \citep{zhang2025generative}: a) Increased zero-shot generalization, where generative models are suitable for handling objects or environments previously unseen, b) Handling multi-modality, due to the ability of generative models of predicting missing information between different sensorial data, and c) Long-horizon planning, where generative models enable robots to perform accurate predictions for multiple future steps. On the contrary, its main limitations are \citep{liu2024rdt}: a) Sim-to-real gap, where the generated data can miss subtle physical nuances, b) Presence of hallucinations, where generative robots are prone to hallucinating a physical capability, and c) Increased inference latency, where generative models typically require extensive computations.

The most common GL techniques in robotics are:

\begin{itemize}
    \item \underline{Autoregressive sequence modeling}: The main goal lies on predicting the next action or state based on previous observations. In this respect, PACT \citep{bonatti2023pact} trains a causal transformer to predict the next observation-action token, so that a single model can capture long-horizon structure across different tasks. Additionally, long-horizon manipulation is modeled through sequential generation of action tokens \citep{zhang2025autoregressive}.

    \item \underline{Diffusion-based action policies}: This employs a diffusion model for generating a chunk of actions at once, by gradual application of a denoising process. In particular, Diffusion Policy \citep{chi2025diffusion} learns a conditional denoising process that samples action sequences for handling multi-modal behaviors and supporting stable visuomotor control. Similarly, Legibility Diffuser \citep{bronars2024legibility} comprises an intent-expressive variant of the latter.

    \item \underline{Generative video and scene synthesis}: The fundamental aim comprises the creation of a model of the physical reality, which in turn enables robots to imagine, to simulate, and to plan actions prior to their execution in the real world. In this context, RoboDreamer \citep{zhou2024robodreamer} employs compositional video WMs, while ReBot \citep{fang2025rebot} makes use of a real-to-sim-to-real synthesis approach.

\end{itemize}

\section{Learning stages}
\label{sec:learning_stages_details}

This section provides the full details of the different learning stages involved in the development of robotic FM methods, complementing the condensed presentation of Section~\ref{sec:learningStage} of the main text. For each learning stage (namely, pre-training, offline fine-tuning, online adaptation, and continuous learning), the complete category definition, the corresponding main advantages and limitations, as well as the extensive list of representative methods per adopted learning paradigm are provided below.

\subsection{Pre-training}
\label{sup:ssec:stages_pre-training}
The ultimate goal of the pre-training stage is to estimate robust, general-purpose representations of robotic data. In particular, instead of training a robotic agent for a specific task or application, pre-training aims at processing massive (often internet-scale) amounts of diverse data (e.g., human demonstration videos, simulation data, sensorial data streams, etc.) from multiple robotic platforms, in order to acquire knowledge and to model the cross-correlations among vision, language, and action. Its main advantages are \citep{li2024foundation}: a) Increased generalization, where pre-trained models are more likely to adapt to new environments, than undergoing training from scratch, b) Robust zero-shot capability, where pre-trained models can often perform robustly tasks that they weren't explicitly trained for, without the need for extra training samples, and c) Accurate multi-modal mapping, where due to the usual large-scale datasets used, pre-trained models are proven to robustly map across textual words, visual concepts, and physical actions. On the contrary, its main limitations are \citep{kawaharazuka2025vision}: a) Excessive computational cost, where pre-training requires massive GPU resources to be implemented, b) Reduced performance in real-world, where if a model is pre-trained largely on internet or simulation data, it may not always perform robustly in real-world, physical circumstances, and c) Safety concerns, where training using internet resources does not always take into consideration strict safety constraints.

The most common learning paradigms adopted during the pre-training stage are:

\begin{itemize}

   \item \underline{Supervised learning}: Supervised learning aims at equipping a model with a foundational understanding of the world from high-quality, diverse, labeled data. In particular, PaLM-E \citep{palme} is jointly trained using both web-scale multi-modal data and embodied experiences. Similarly, RT-$2$ \citep{rt2} and Gemini Robotics \citep{team2025gemini} are constructed using web and robot manipulation data.

  \item \underline{Self-supervised learning}: SSL targets to enable robots to operate beyond narrow, task-specific programming, aiming at accomplishing generalized intelligence capabilities. In particular, DINOv$2$ \citep{oquab2023dinov2} and VideoMAE \citep{tong2022videomae} can learn perception priors from unlabeled data. Moreover, R$3$M \citep{nair2022r3m} extends this approach, by incorporating egocentric features.

  \item \underline{Imitation learning}: The goal of IL is to learn a prior distribution of successful behaviors directly from expert demonstrations. In particular, RT-$1$ \citep{rt1} treats robot control as a next-token prediction problem over multi-modal streams, so that the learned policies to inherit both semantic and sensorimotor skills. Additionally, RT-$2$ \citep{rt2} is trained using vision-language and robot action tokens, while Octo \citep{team2024octo} makes use of the Open-X-Embodiment trajectories \citep{openx} for deriving a generalist policy. Moreover, Seer \citep{tianpredictive} investigates the scaling laws of multi-task IL, aiming at drastically reducing the required target-domain data.

\end{itemize}

\subsection{Offline fine-tuning}
\label{sup:ssec:stages_fine-tuning}
Offline fine-tuning aims at bridging the knowledge gap between general-purpose representations (learned during the pre-training step) and the specificities of a given physical-world application. In short, the primary purpose of this stage is task and embodiment specialization. Its main advantages are \citep{hu2023toward}: a) Reduced need for training data, where the need for training samples is significantly lower than the pre-training step, b) Increased training stability, where models are likely to converge to robust performance states, provided that sufficient training examples are available, and c) Knowledge distillation, where only the necessary parts of the general-purpose representations learned during the pre-training stage can be used for the given application at hand. On the contrary, its main limitations are \citep{firoozi2025foundation}: a) Distribution shift, where if a robot encounters novel state challenges, it is difficult for it to handle, b) Dependency on data quality, where the presence of suboptimal or erroneous demonstrations can mislead the training process, and c) Lack of online exploration, where the model can only employ skills present in the training dataset, while not being able to adapt to online challenges.

The most common learning paradigms adopted during the offline fine-tuning stage are:
\begin{itemize}

  \item \underline{Imitation learning}: IL aims at equipping pre-trained models with the necessary low-level precision skills for a specific application. In particular, by mimicking expert demonstrations, it targets to learn specific motor commands with respect to a given task or robotic platform. In particular, Octo \citep{team2024octo} and OpenVLA \citep{openvla} employ large-scale Open X Embodiment pretraining and, subsequently, focus on new robotic platforms, making use of LoRA-style adapters. Similarly, LiteVLA \citep{williams2025lite} demonstrates that NF4 quantized LoRA can be tuned on CPU-only hardware.

  \item \underline{Reinforcement learning}: RL enables robots to learn from a reward signal, by focusing on performed actions that lead to successful task executions. In particular, a recurrent tracker is trained, using conservative offline RL, on VFM annotated trajectories, prior to deployment \citep{zhong2024empowering}. Additionally, FLaRe \citep{hu2025flare} applies large-scale RL fine-tuning to transformer-based policies, in order to improve long-horizon mobile manipulation.

  \item \underline{Generative learning}: GL facilitates the specialization of pre-trained models to specific environments, especially in the presence of sparse constraints on the target task. In particular, ReBot \citep{fang2025rebot} replays real trajectories in simulation and, subsequently, composes them into inpainted real backgrounds to adapt to new domains. Additionally, RoboDreamer \citep{zhou2024robodreamer} makes use of compositional WMs to generate imagined video plans, which serve as additional training data.

\end{itemize}

\subsection{Online adaptation}
\label{sup:ssec:stages_adaptation}
Online adaptation targets to equip robots with the appropriate routines for learning in real-time from their own experiences. In practice, at this stage robots aim to handle the distribution shift between the offline training data and the ones encountered during online deployment. Its main advantages are \citep{firoozi2025foundation}: a) High precision performance, where robots typically accomplish superior task execution accuracy for the specific environments to which they are adapted, b) Continuous improvement, where the robot continuously increases its performance as it constantly learns from its experiences, and c) Robustness to distribution shifts, where the models achieve to maintain performance in the presence of environmental changes. On the contrary, its main limitations are \citep{yuan2025survey}: a) Catastrophic forgetting, which denotes the risk that might occur as the robot learns new skills to lose some of its general-purpose ones, b) Computational latency, where the algorithmic operations involved need to be performed in (near) real-time, and c) Prone to noise, where real-time, noisy sensorial data may mislead the adaptation process.

The most common learning paradigms adopted during the online adaptation stage are:
\begin{itemize}

    \item \underline{Domain adaptation}: This aims at handling the particular physical-world constraints and sensorial noise for a specific robot deployment scenario. In practice, this enables the model to re-calibrate its knowledge structures to the perceived environment. In this context, Test-Time Adaptation (TTA)-Nav \citep{piriyajitakonkij2024tta} incorporates a reconstruction decoder on top of a pre-trained policy, so that the agent can denoise corrupted frames without gradient updates, while restoring point-goal navigation under severe corruptions. Additionally, Phys2Real \citep{wang2025phys2real} targets to bridge sim-to-real gaps, by combining FM priors with interaction-based estimations.

    \item \underline{Reinforcement learning}: The aim of RL lies on allowing the robot to perform micro-adjustments to its general-purpose knowledge, based on sensory feedback and exploration. In this context, self-improving embodied FMs refine pre-trained policies, based on reward and success estimation from model predictions, across a robot fleet \citep{ghasemipour2025self}. Additionally, RL-VLM-F \citep{wang2024rl} estimates rewards, using a VLM model that compares trajectory snippets with language goals. Similarly, VARP \citep{singh2025varp} regularizes VLM-derived preferences with the agent's own rollouts for reducing misalignment. Moreover, Eureka \citep{ma2023eureka} and DrEureka \citep{ma2024dreureka} are LLM-guided RL frameworks that automate reward design and, in the case of DrEureka, domain randomization, in order to improve policies for locomotion and dexterous manipulation.

    \item \underline{In-context/prompt learning}: This paradigm is particularly suitable for online adaptation, since it enables zero- or few-shot specialization at deployment time, without requiring updates to the underlying model weights. In particular, ICRT \citep{fu2025icrt} employs in-context imitation policies that condition on a small number of recent demonstration trajectories for adapting to novel manipulation tasks. Additionally, LLM-based control stacks can interleave reasoning and acting through ReAct-style prompting, allowing robots to monitor execution progress and to revise plans when needed \citep{yao2023react}. Similarly, verification-based frameworks can check high-level task plans prior to execution, improving reliability in dynamic deployment settings \citep{grigorev2025verifyllm}.

\end{itemize}

\subsection{Continuous learning}
\label{sup:ssec:stages_continuous_learning}
Continuous Learning (CL) (often termed lifelong learning) aims at enabling robots to acquire new skills or to adapt to new environments incrementally, without requiring to be fully retrained from scratch. In practice, it employs conventional learning paradigms (e.g., IL, RL, etc.) in slower outer loops for improving the robot's performance. Its main advantages are \citep{xiao2025robot}: a) Increased adaptability, where robot capabilities can continuously evolve in response to changes in the surrounding  environment, making them suitable for long-term deployment, b) Improved scalability, where data from a single robot can be used to update the knowledge structures of an entire fleet, and c) Reduced downtime, where robots that continuously update their underlying models are less likely to become non-sufficiently robust/operational for long periods of time. On the contrary, its main limitations are \citep{firoozi2025foundation}: a) Catastrophic forgetting, where the continuous update in the robot's knowledge structures may result into overwriting previously learned skills, b) Stability-plasticity dilemma, which corresponds to a critical trade-off between the ability to integrate new skills and the capability to maintain the old ones, and c) Memory overhead, where the model needs to maintain increased past data for robustly updating its behavior in subsequent steps.

The most common learning paradigms adopted during the continuous learning stage are:

\begin{itemize}

  \item \underline{Domain adaptation}: DA enables a model to continuously adjust its internal knowledge to dynamic, real-world settings. In particular, Action Flow Matching for Lifelong Learning \citep{murillo2025actionflowmatching} develops a lifelong robot-learning framework that incrementally aligns robot dynamics across sequential tasks, supporting efficient and safe continual adaptation. On another direction, VR-Robo \citep{zhu2025vr} utilizes a real-to-sim-to-real framework for constructing photorealistic digital twins from logged data, retraining navigation or locomotion policies in them, and transferring the acquired skills to the real-world.

  \item \underline{Imitation learning}: IL aims at constantly bridging the gap between a model's general knowledge and the specific, high-precision requirements of a real-world application case, on the condition of the availability of few expert demonstrations. In particular, SkillsCrafter \citep{wang2026lifelong} realizes lifelong language-conditioned robot learning across multiple sequential manipulation skills, while reducing catastrophic forgetting through symbolic skill distillation. Additionally, LOTUS \citep{wan2024lotus} models and refines manipulation skills from demonstration streams, supporting long-term expansion of its skill repertoire.


 \item \underline{Reinforcement learning}: RL supports continuous learning by enabling FM robotic policies to improve through repeated interaction, autonomous practice, and reward-driven post-training over extended deployment horizons. In particular, self-improving embodied FMs refine pretrained policies through autonomous practice based on self-predicted rewards and success signals across a robot fleet, thereby enabling downstream skill acquisition with minimal human supervision \citep{ghasemipour2025self}. Similarly, LiReN \citep{stachowicz2024lifelong} demonstrates that navigation FMs can improve lifelong learning, through the employment of online RL pipelines, in open-world settings. More recently, VLA models have been adapted through reinforcement fine-tuning, showing improved retention and adaptation to new tasks, while mitigating catastrophic forgetting in sequential manipulation tasks \citep{liu2026towards}.

\end{itemize}

\section{Robotic tasks}
\label{sec:robotic_tasks_details}

This section provides the full details of the different robotic tasks for which FM-based solutions have been developed, complementing the condensed presentation of Section~\ref{sec:tasks} of the main text. For each robotic task (namely, perception, planning, navigation, manipulation, and human-robot interaction), the complete category definition, the corresponding main advantages and limitations, as well as the extensive list of representative methods per sub-category are provided below.

\subsection{Perception}
\label{sup:ssec:robotic_tasks-perception}
Perception aims at creating rich, semantic maps of the surrounding environment, which subsequently enable robots to execute individual actions. In particular, robot perception enables the realization of semantic grounding (i.e., the connection of visual stimuli with real-world entities), discovery of object affordances (i.e., the tasks that can be performed with different objects), and contextual awareness (i.e., the identification of the different types of semantic entities and their location). The main advantages of the incorporation of FMs in robot perception are \citep{kawaharazuka2024real}: a) Open-vocabulary recognition, where models can identify entities for which they are not specifically trained for, b) Zero-shot generalization, where robots can handle novel environments or object types, c) Multi-modal fusion, where robotic agents can efficiently combine multiple information streams (e.g., visual, language, proprioception, etc.), and d) Robustness to noise, where FMs are shown to be reliable in the presence of noisy data. On the contrary, the main limitations of the integration of FMs in robot perception are \citep{hu2023toward}: a) High latency, due to the typical extreme scale of the underlying models employed, b) Decreased explainability, where the main factors leading to a particular robot decision is difficult to be precisely defined, c) Presence of hallucinations, where model predictions can be misled and to result in failures in the physical world, and d) Spatial imprecision, which relates to inaccurate localization of (even correct) entity predictions.

The main categories of perception methods are:
\begin{itemize}
  \item \underline{Language-grounded detection and segmentation}: This aims at identifying (detection) and precisely outlining (segmentation) the objects present in the robot's surrounding environment based on natural language prompts. In particular, GLIP \citep{li2022grounded} and Grounding DINO \citep{liu2024grounding} provide phrase-based detections that remain robust in the presence of clutter as well as in the zero-shot setting; such detections can subsequently feed language-conditioned manipulation and navigation pipelines, such as CLIPort \citep{shridhar2022cliport} and similar solutions \citep{unlu2024reliable,10977668}. Additionally, SAM-style promptable segmenters \citep{sam,ravisam,carion2025sam} allow the incorporation of box, click, or text prompt information into control pipelines in an interactive way, like SAM-$6$D \citep{lin2024sam} for $6$D pose estimation and RoG-SAM \citep{mei2025rog} for instance-level robotic grasping detection.

  \item \underline{Open-vocabulary $3$D semantic mapping}: This allows robots to perceive and to localize objects of previously unseen categories in the $3$D space, making use of natural language inputs. In this respect, ConceptFusion \citep{jatavallabhulaconceptfusion} builds open-set, language-searchable maps that support uncommon and previously unseen concepts, and multi-modal queries. Additionally, ConceptGraphs \citep{gu2024conceptgraphs} estimates object nodes and their relations, so that task planners can subsequently operate on top of a semantic scene graph (instead of raw pixels). Moreover, radiance-field (e.g., LERF \citep{kerr2023lerf}) and $3$D-Gaussian (e.g., FMGS \citep{zuo2025fmgs}) grounded approaches embed CLIP/DINO features into neural fields for estimating consistent, view-invariant labels. Furthermore, OpenFusion++ \citep{jin2025openfusion++}, OpenVox \citep{deng2025openvox}, and MR-COGraphs \citep{gu2025mr} focus on real-time, open-vocabulary voxel mapping and multi-robot scene graphs for efficient robot exploration.

  \item \underline{Pose estimation and affordance prediction}: Pose estimation aims at aligning an object's local coordinate system to the global, world one, while affordance prediction aims at the detection of the ways that an object can be manipulated; in both cases, FMs are particularly useful due to their inherent ability of linking semantics (language) with spatial geometry (pixels). In particular, OV$9$D \citep{cai2024ov9d} estimates category-agnostic $9$-DoF pose and object size without relying on the use of CAD models. Similarly, Oryon \citep{corsetti2024open} aligns CLIP-guided segments across different views, in order to recover relative $6$D pose for unseen objects. On the other hand, OpenAD \citep{nguyen2023open} models zero-shot $3$D affordances in a shared vision-language embedding space, while OVA-Fields \citep{su2025ova} extends this direction to weakly supervised open-vocabulary affordance fields for robot operational part detection in $3$D scenes. Moreover, one-shot open affordance learning transforms a single example into dense, class-agnostic affordance masks \citep{li2024one}.

  \item \underline{Contact-centric and visuotactile perception}: This category of methods focuses on analyzing and modeling the physics of robot interactions for enhancing perception. In particular, contact-centric approaches consider junction points as the primary state representation, while visuotactile ones integrate exocentric vision with local tactile sensing. In this respect, TLA \citep{hao2025tla} and Tactile-VLA \citep{huang2025tactile} combine tactile information streams with vision and language, in order to improve insertion, assembly, and material reasoning tasks. Large tactile-vision-language models (e.g., TALON \citep{jiang2024talon}) further extend this idea to richer contact semantics. Moreover, visuotactile systems, such as NeuralFeels \citep{suresh2024neuralfeels}, can enhance in-hand pose and shape estimation, when visual cues are uncertain.

  \item \underline{Long-term object tracking}: This aims at locating a given object or point across a sequence of video frames, while maintaining prediction stability over time and capability to recall objects when they become occluded or exit the field of view. The latter is a particular requirement for long-horizon tasks. In this respect, OVTrack \citep{li2023ovtrack} employs language and diffusion priors to generalize multi-object tracking to unseen categories without explicit video pre-training. Similarly, DINO-MOT \citep{lee2024dino} combines DINOv$2$ features with a memory mechanism for robust pedestrian tracking, while COVTrack \citep{qian2025covtrack} further strengthens open-vocabulary tracking through improved temporal association across continuous trajectories.
\end{itemize}

\subsection{Planning}
\label{sup:ssec:robotic_tasks-planning}
Planning serves as the fundamental bridge between high-level (semantic) reasoning and low-level motor control procedures. In particular, the primary usefulness of robot planning lies in long-horizon task decomposition, where a high-level task goal needs to be broken down into multiple, sequential, individual sub-goals over an extended period, prior to their actual execution. The main advantages of the incorporation of FMs in robot planning are \citep{hu2023toward}: a) Increased interpretability, where FMs enable a planned sequence of robot actions to be represented in human-like form, b) Increased generalization	ability, where robots can often adapt to novel settings by leveraging general-purpose, world knowledge stored in a FM, c) Reduced need for training data, where pre-trained FMs exhibit a decreased need for large-scale, expensive, visuomotor robot data, and d) Safety contraints integration, where FMs enable the efficient incorporation of safety constraints directly into the planning loop. On the contrary, the main limitations of the integration of FMs in robot planning are \citep{firoozi2025foundation}: a) Logical gaps, where FM-based planners may estimate action steps that are not physically possible, b) Lack of grounding, where FM inference may result into deviations between a high-level plan and the low-level capacities of the robot at hand, c) Increased latency, where the high-computing nature of FMs may be proven restrictive for high-pace tasks, and d) Increased closed-loop complexity, where FM-based solutions face challenges in adjusting their behavior in real-time settings, as a response to constant environmental changes.

The main categories of planning methods are:
\begin{itemize}
  \item \underline{Language-driven task decomposition}: This aims at breaking down high-level, long-horizon goals into logical sequences of robot primitive actions (atomic skills), often in the form of structured, executable code (program synthesis). In particular, SayCan \citep{saycan} and SayPlan \citep{rana2023sayplan} ground each action step on an affordance map or a $3$D scene graph, respectively, so that abstract sub-goals to correspond to concrete objects and locations. On the other hand, Code-as-Policies \citep{liang2023code} and similar approaches generate directly short, Python-like programs for integrating existing software libraries, rendering planning easier to inspect, to test, and to modify \citep{singh2023progprompt,huang2023voxposer}.

  \item \underline{Neuro-symbolic closed-loop reasoning}: This category of methods combines the general-purpose knowledge of a FM with formal, logical checking procedures, so as to guarantee the successful operation of robot agents in the real world. In this respect, ISR-LLM \citep{zhou2024isr} converts natural language instructions to PDDL ones and iteratively refines the estimated plans using a symbolic validation scheme, until a feasible sequence of actions is determined. Additionally, AutoTAMP \citep{chen2024autotamp} employs an LLM for generating or translating high-level language instructions into intermediate representations suitable for a task-and-motion planning (TAMP) solving mechanism, while making use of autoregressive re-prompting to correct syntactic and semantic errors. Moreover, safety- and feasibility-oriented planners, such as SELP \citep{wu2025selp} and LLM-GROP \citep{zhang2025llm}, translate LLM proposals into explicit constraint checking and task-and-motion reasoning procedures, in order to avoid unsafe or dead-end policies.

  \item \underline{Multi-modal policy generation}: This category focuses on generating high-level plans or robot actions by integrating multiple input modalities, such as language, vision, and embodied state information. In particular, PaLM-E \citep{palme} combines language, vision, and proprioception for embodied reasoning and action generation, while RT-$2$ \citep{rt2} and OpenVLA \citep{openvla} demonstrate that language-aligned visual representations can transfer web-scale semantic knowledge to real-world robot control.

  \item \underline{Execution-time validation and failure recovery}: This category focuses on monitoring generated plans during deployment, verifying action preconditions, detecting inconsistencies or failures, and triggering corrective re-planning when needed. In this respect, Code-as-Monitor \citep{zhou2025code} introduces constraint-aware visual programming for reactive and proactive robotic failure detection, SELP \citep{wu2025selp} incorporates explicit safety and feasibility constraints into the planning loop, while  VLM-based monitoring frameworks, such as Guardian \citep{pacaud2025guardian} and unified real-time failure-handling approaches \citep{ahmad2025unified}, support execution-time failure detection and recovery in robotic manipulation.

  \item \underline{Semantic multi-robot coordination}: This category of methods capitalizes on the broad knowledge base of FMs for coordinating multi-robot setups, where accurate reasoning about task dependencies, resources, and scheduling is needed. In this direction, LiP-LLM \citep{obata2024lip} employs an LLM to create a skill list and a corresponding dependency graph from language instructions, while subsequently relying on linear programming techniques to allocate tasks across robotic platforms. Additionally, SMART-LLM \citep{kannan2024smart} assigns agents task-specific roles based on structured representations of their skills and capabilities, reducing in this way instruction drift and enabling coalition formation and task allocation, based on a single high-level instruction. Moreover, large-scale orchestration systems, such as AutoRT \citep{ahn2024autort}, combine language-based task assignment with human oversight and monitoring, demonstrating that FM-based planners can coordinate dozens of physical robots in real-world environments.
\end{itemize}

\subsection{Navigation}
\label{sup:ssec:robotic_tasks-navigation}
The fundamental usefulness of FMs in robot navigation lies on providing the necessary spatial common-sense knowledge in embodied AI settings. The latter is mainly accomplished due to the capacity of FMs to process the robot's surrounding environment as a high-level semantic space, instead of considering rigid, inflexible spatial maps. The main advantages of the incorporation of FMs in robot navigation are \citep{pan2025mixed}: a) Open-world navigation, where robots can operate in environments including previously unseen entities, b) Cross-embodiment transfer, where a single pretrained model can be deployed in different/diverse hardware platforms, and c) Semantic reasoning capability, which relies on the inherent ability of FMs to combine vision with language understanding for instruction interpretation. On the contrary, the main limitations of the integration of FMs in robot navigation are \citep{firoozi2025foundation}: a) Sim-to-real gap, where FMs trained in simulation may exhibit difficulties in operating in real-world environments, b) Lack of training data, where large-scale, high-quality, diverse $3$D navigation data, required for FM training, is difficult and expensive to collect, c) Increased latency, where the high computational needs of FMs can lead to challenging situations in real-world applications, and d) Safety and ethical concerns, where FMs need to be equipped with appropriate social norms for operating in human-shared spaces.

The main categories of navigation methods are:
\begin{itemize}

  \item \underline{Semantic spatial grounding}: This category of methods aims at registering the objects present in the environment, but also their relative position (with respect to the robots) and a concrete action plan for reaching them (in natural language form). In particular, VLMaps \citep{huang2023visual} builds CLIP-indexed spatial memories that enable the system to query objects and rooms without task-specific retraining, directly ingesting visual-language features into a $3$D map. Zero-shot localization schemes, such as PixNav \citep{cai2024bridging} and VLTNet \citep{wen2025zero}, guide navigation through pixel-level target cues or construct semantic navigation maps and rank exploration frontiers from language prompts. Moreover, open-vocabulary mapping methods (like One Map to Find Them All \citep{busch2025one}, DualMap \citep{jiang2025dualmap}, and scene graph-based approaches \citep{loo2025open}) support multi-object navigation, dynamic environments, and functional queries, so that a single map can accommodate for many language goals and robots.

  \item \underline{Instruction-following policies}: This is based on the increased capability of FMs in interpreting natural language instructions, without requiring a pre-defined map or hard-coded scripts for every possible object present in the environment. In this respect, generalist navigation models, such as GNM \citep{shah2023gnm} and ViNT \citep{shah2023vint}, formalize navigation as a sequence prediction problem over images and poses, relying on a single model across different robots and environments. Additionally, NaviLLM \citep{zheng2024towards} unifies instruction following and embodied QA with schema-tuned prompts, while NavFormer \citep{wang2024navformer} learns target-driven policies in unknown, dynamic environments. Moreover, FASTNav \citep{chen2024fastnav} demonstrates that compact, LoRA-adapted language models can operate in real-time on embedded hardware, offering a practical path from large offline training to on-board controllers.

  \item \underline{End-to-end policies}: This aims at developing a unified architecture that can map raw sensorial data directly to motor commands, instead of adopting the conventional modular design of breaking down navigation into individual steps (like mapping, localization, and path planning). In particular, ViNT \citep{shah2023vint} learns a generalizable visuomotor navigation policy across multiple robots and environments, while NavFoM \citep{zhang2026embodied} extends this idea towards cross-embodiment and cross-task navigation. In driving-oriented settings, end-to-end frameworks, such as DiffusionDrive \citep{liao2025diffusiondrive} and DriveGPT-4 \citep{xu2024drivegpt4}, further demonstrate that multi-modal models can predict low-level control signals directly from visual- and language-conditioned inputs.

\end{itemize}

\subsection{Manipulation}
\label{sup:ssec:robotic_tasks-manipulation}
The primary utility of FMs in robot manipulation lies on providing the required knowledge for implementing the translation from high-level, semantic, human-like instructions to precise physical pressures and movements needed to manipulate an object. The main contribution of FMs for achieving the latter comprises their cross-embodiment learning capability, where the same model pretrained on data from multiple different platforms can be deployed to diverse setups. The main advantages of the incorporation of FMs in robot manipulation are \citep{li2024foundation}: a) Increased generalization, where FMs are shown to be robust in handling previously unseen object types, b) Improved robustness, where FMs can make use of real-time (visual) feedback to adjust their manipulation strategy on the fly, and c) Enhanced embodiment capability, where FM solutions enable the understanding of the physical properties of the objects, prior to their manipulation. On the contrary, the main limitations of the integration of FMs in robot manipulation are \citep{sapkota2025vision}: a) Limited training data, where sufficient quantities of high-quality, labeled robotic manipulation data is difficult to collect, b) Safety concerns, where the robot policies are not always guaranteed to result into feasible and safe manipulations, and c) Low action precision, where the increased generalization ability of FMs is accompanied with corresponding decrease in physical task execution for specialized domains.

The main categories of manipulation methods are:
\begin{itemize}

  \item \underline{Language-to-action models}: This category aims to interpret the semantic meaning of a natural language command and to map it to the physical world setting, without requiring task-specific programming. In particular, RT-$1$ \citep{rt1} and RT-$2$ \citep{rt2} approach manipulation as a sequence modeling problem over multi-modal tokens, while PaLI-X \citep{chen2023pali} aims at incorporating broad visual knowledge. Additionally, OpenVLA \citep{openvla} comprises a large vision-language-action policy that adapts to new robotic platforms, based on small-scale fine-tuning. GR$00$T N$1$ \citep{bjorck2025gr00t} combines a deliberative VLM  with a diffusion motor policy for bimanual humanoid skills acquisition. Moreover, state-space variants, such as RoboMamba \citep{liu2024robomamba}, replace the core transformer component with a Mamba-based selective state-space model for lowering latency, while maintaining increased visuomotor skill performance.

  \item \underline{Retrieval-augmented imitation learning}: This relies on receiving guidance from a large database of relevant previous demonstrations for predicting future actions. In this context, DINOBot \citep{di2024dinobot} detects similar demonstrations based on DINO feature correspondence and subsequently estimates dense manipulation trajectories for realizing one- and few-shot generalization to novel objects. Additionally, STRAP \citep{memmelstrap} retrieves relevant manipulation sub-trajectories from prior demonstrations and uses them to augment few-shot imitation learning, improving generalization to novel objects and tasks.

  \item \underline{Constraint-aware policy synthesis}: This category employs a FM to generate high-level control code or mathematical objectives, which are explicitly bounded by physical, safety, and environmental constraints. In this context, CoPa \citep{huang2024copa} detects task-relevant parts using a multi-modal LLM and estimates spatial constraints that a conventional planner translates into $6$-DoF actions. ReKep \citep{huang2025rekep} represents tasks as sequences of relational keypoint constraints and optimizes them hierarchically for single- and dual-arm assembly tasks. Moreover, part-centric perception methods, like PartSLIP++ \citep{zhou2023partslip++}, estimate part-affordance correlations that are required for robust task execution.

  \item \underline{Semantic spatial maps}: This category aims at generating representations of the environment that combine $3$D spatial geometry with semantic information about the involved objects. In this direction, VoxPoser \citep{huang2023voxposer} estimates constraints and object affordances from natural language inputs, creates $3$D value maps from vision-language information cues, and applies common motion planning routines in a zero-shot fashion. Additionally, AdaRPG \citep{zhang2025adaptive} leverages VLMs to infer part affordances and operational constraints that guide primitive skills for articulated-object manipulation.
\end{itemize}

\subsection{Human-robot interaction}
\label{sup:ssec:robotic_tasks-human-robot-interaction}
The fundamental usefulness of FMs in HRI is grounded on their increased ability for realizing semantic, human-like reasoning and interpreting human intent. In particular, robots are capable of reacting to conversational instructions, asking clarification questions, understanding contextual settings, and providing explanations for their actions, which greatly simplifies communication (especially) with non-expert users. The main advantages of the incorporation of FMs in HRI are \citep{zhao2025multimodal}: a) Rapid generalization, where a human user can demonstrate a new task to a robot and it can adapt instantly, b) Intuitive control, where due to the increased human behavior interpretation capabilities of FMs, robot control becomes more efficient and intuitive, c) Increased safety alignment, where user-provided feedback can boost robots to learn/incorporate social norms, and d) Efficient error recovery, where human provided feedback can be rapidly exploited by a robot for recovering from a fault state. On the contrary, the main limitations of the integration of FMs in HRI are \citep{xiao2025robot}: a) Lack of training data, where data collection involving human feedback/interactions is typically costly, b) Incorporation of human bias, where robots learn in an unconstrained way from their human-provided feedback, and c) Semantic drift occurrences, where human intent is not always easy to interpret as contextual/environmental conditions may change.

The main categories of HRI methods are:
\begin{itemize}

  \item \underline{Conversational policy alignment}: This category of methods focuses on using natural language dialogue to dynamically adjust a robot's behavior in real-time, so that it matches a human's specific intent, preferences, and safety boundaries. In particular, DRAGON \citep{liu2024dragon} comprises a dialogue-based navigation framework that grounds free-form commands in visual landmarks, describes the environment, and asks clarification questions when the reference is unclear. Additionally, PlanCollabNL \citep{izquierdo2024plancollabnl} translates spoken instructions into editable plans so that users can insert, remove, or reorder steps in collaborative manipulation and assembly settings.

  \item \underline{Reciprocal social tuning}: This aims at developing a high-level, semantic communication framework, where both humans and robots continuously adjust their behaviors, social cues, and expectations to harmonize each other. In this context, incremental system updates combine natural-language feedback with on-robot sensing so that the controller can refine prompts, code snippets, or skill graphs after each mistake and also to reuse the acquired knowledge at a later stage \citep{barmann2024incremental}. Additionally, design studies on conversational companion robots provide concrete guidelines for everyday HRI settings, such as clear grounding, turn-taking, and repair under uncertainty \citep{irfan2024recommendations}. TidyBot \citep{wu2023tidybot} demonstrates that LLMs can learn user-specific preferences for household clean-up from a few examples and then generalize these preferences to new objects and scenes. Similarly, LAMS \citep{tao2025lams} extends this idea to assistive teleoperation, by using an LLM to switch control modes based on task context and improving performance over time based on user feedback.

  \item \underline{Active alignment and mitigation}: This targets to ensure that a robot remains synchronized with human intent, while proactively handling errors or deviations. In this respect, RoboVQA \citep{sermanet2024robovqa} queries egocentric video to check preconditions, to verify whether a sub-task has succeeded, and to proactively request human intervention in case of identified failures. Additionally, embodied LLM controllers also incorporate human feedback for adjusting their plans when they detect that the current state drifts from the expected goal \citep{mon2025embodied}. Overall, the common practice relies on combining a latency-aware VLM with a dialogue-based manager, so that the robot can provide concise explanations, ask specific clarifications, and implement targeted re-planning \citep{barmann2024incremental,sermanet2024robovqa}.

\end{itemize}

\section{Application domains}
\label{sec:application_domains_details}

This section provides the full details of the different real-world application domains in which robotic FM-based solutions have been deployed, complementing the condensed presentation of Section~\ref{sec:app_domains} of the main text. For each application domain (namely, agentic mobility, industrial manipulation, supply operations, service robots, medical robots, cognitive agrisystems, crisis agents, maritime robotics, and space robotics), the complete domain definition, as well as the extensive list of representative methods are provided below.

\subsection{Agentic mobility}
\label{sup:ssec:application-domains_mobility}
The integration of FMs has induced a paradigm shift in autonomous movement, transitioning from rigid path-following routines to autonomous agentic solutions, where robots leverage high-level reasoning to understand mission objectives and to adapt to environmental changes.

A core application of these models is in semantic spatial grounding, which replaces traditional, inflexible spatial maps with queryable memories. For instance, VLMaps \citep{huang2023visual} constructs CLIP-indexed spatial memories by directly ingesting visual-language features into $3$D maps; this allows robotic systems to navigate to specific rooms or objects via natural language queries, without requiring task-specific retraining. Similarly, methods like EffoNAV \citep{shen2025effonav} pair CLIP-based goal detection with exploration routines and low-level controllers to handle visual targets in challenging settings.

FMs also address the per-robot engineering bottleneck through cross-embodiment and cross-site generalization. Generalist models, such as GNM \citep{shah2023gnm} and ViNT \citep{shah2023vint}, formalize navigation as a sequence prediction problem over images and poses. This enables a single visual backbone to drive diverse mobile platforms across varied environments, significantly facilitating the rollout of new robots in novel sites. This scalability is further demonstrated by City Walker \citep{liu2025citywalker}, which learns policies that transfer across road networks in different cities.

For ensuring real-time feasibility on embedded hardware, particular focus has been given on model efficiency. In particular, FASTNav \citep{chen2024fastnav} utilizes compact, LoRA-adapted language models to provide instruction-following capabilities that run on-board in real-time. Additionally, frameworks like NaviLLM \citep{zheng2024towards} unify instruction following with embodied question-answering to provide a flexible interface for operators.

In road environments, FMs enhance safety and transparency, by conditioning planning on linguistic reasoning or graph-based visual question answering. Frameworks like DriveGPT-4 \citep{xu2024drivegpt4}, Drive Anywhere \citep{wang2024drive}, and DriveLM \citep{sima2024drivelm} go beyond black-box end-to-end policies, by producing human-readable rationales and intermediate decisions that are easier to inspect, to debug, and to align with safety standards. For assistive scenarios, DRAGON \citep{liu2024dragon} uses dialogue-based navigation to allow users to describe targets and constraints in natural language, where the robot provides verbal explanations of its planned route.

\subsection{Industrial manipulation}
\label{sup:ssec:application-domains_industrial}
FMs are fundamentally redefining industrial automation by replacing rigid, task-specific routines with flexible, general-purpose agents that excel in unstructured manufacturing environments. In particular, models like RFM-$1$ \citep{covariant2024rfm1} are deployed within high-throughput production cells, leveraging visuo-motor backbones trained on millions of real-world pick actions to achieve zero-shot generalization, when encountering novel objects. For large-scale logistics, PRIME-$1$ \citep{ambirobotics2025prime1} provides a foundation tailored to parcel workflows, facilitating site-specific adaptation and the rapid rollout of autonomous systems across diverse distribution centers. In order to address hardware variability, OpenVLA \citep{openvla} utilizes parameter-efficient fine-tuning to simplify transfer across different robotic platforms, effectively reducing the need for extensive per-line retraining, while enabling operators to issue high-level, language-conditioned tasks. Moreover, RT-$2$ \citep{rt2} capitalizes on web-to-robot knowledge transfer to significantly boost robustness against long-tail objects and complex instructions on assembly and kitting lines, ensuring that robots can adapt to items outside their original training data.

\subsection{Supply operations}
\label{sup:ssec:application-domains_supply}
FMs are drastically evolving autonomous supply operations, by providing robots with the flexibility and intelligence needed to handle unstructured, dynamic, and complex logistics tasks. This paradigm shift allows robotic agents to adapt to new items, facility layouts, and operational disruptions without requiring manual reprogramming. At a broader scale, logistics operations utilize FMs for sustainable planning, multi-agent coordination, and end-to-end decision support, specifically targeting `Logistics 5.0' goals, such as greener routing \citep{nicoletti2024green}. By integrating multi-agent systems with foundation backbones, autonomous supply chains can effectively link demand forecasting, resource allocation, and transport routing through shared representations and agentic reasoning \citep{xu2024multi}. Consequently, FM-driven solutions can support a wide array of critical industrial tasks, including demand sensing, inventory positioning, warehouse tasking, and transport orchestration \citep{nicoletti2025foundation,agaskar2025deepfleet}.

\subsection{Service robots}
\label{sup:ssec:application-domains_service}
The integration of FMs represents a major paradigm shift for service robots, transforming them to adaptable agents, capable of navigating through unstructured domestic environments and interpreting natural language instructions. In this context, TidyBot \citep{wu2023tidybot} exemplifies the use of few-shot LLM reasoning to learn personalized tidying preferences directly from user conversations, allowing a mobile manipulator to ground abstract `put-away' commands into specific physical actions across novel scenes. Extending these capabilities to more complex, multi-step procedures, ELLMER \citep{mon2025embodied} performs tasks such as coffee brewing, by integrating visual, force, and linguistic feedback; specifically, ELLMER utilizes a retrieval-augmented memory mechanism to adapt its plans on the fly if environmental conditions change or failures are detected, ensuring robust execution under uncertainty.

\subsection{Medical robots}
\label{sup:ssec:application-domains_medical}
FMs in the medical domain serve as a critical bridge between high-level clinical knowledge and low-level robotic execution. These models are increasingly integrated into surgical perception and decision-making pipelines to handle high-stake, high-variability tasks. A key application is monocular depth estimation; for instance, Surgical-DINO \citep{cui2024surgical} adapts DINOv$2$ using a LoRA adapter specifically for endoscopic imagery. Similarly, DARES \citep{zeinoddin2024dares} tailors a Depth Anything Model, using self-supervised Vector-LoRA, to better align with surgical scene statistics. Additionally, EndoDDC \citep{lin2026endoddc} addresses sparse-to-dense depth reconstruction for endoscopic robotic navigation through diffusion-based depth completion, supporting more reliable $3$D reconstruction and safe instrument guidance. These advancements allow for the robust implementation of surgical assistance systems, including real-time instrument tracking, tool detection, and intra-operative guidance within hospital settings \citep{he2024foundation}. Ultimately, this integration provides context-aware intelligence and precise assistance, enabling robotic platforms to robustly support complex clinical procedures.

\subsection{Cognitive agrisystems}
\label{sup:ssec:application-domains_agrisystems}
The integration of FMs into cognitive agrisystems marks a transition towards adaptive, embodied intelligence in field operations. By utilizing VLA architectures, these systems can ingest high-level semantic instructions and translate them into precise motor outputs in real-time. Unlike traditional agricultural robots that rely on hard-coded rules, FMs bridge the semantic gap, by using massive pre-trained datasets to reason through the unpredictable variability of biological environments, such as shifting light, overlapping foliage, and irregular crop shapes \citep{yin2025foundation}. For instance, HarvestFlex \citep{zhao2026harvestflex} employs VLA policies for real greenhouse tabletop strawberry harvesting, a long-horizon, unstructured task, challenged by occlusion and specular reflections. Additionally, FM-based reasoning supports task planning and action selection in crop monitoring and field-management scenarios \citep{cuaran2026visual}.

\subsection{Crisis agents}
\label{sup:ssec:application-domains_crisis}
FMs are fundamentally reforming disaster response and public safety, by enabling crisis agents to transition from rigid, remote-controlled setups to autonomous systems capable of high-level reasoning in unpredictable and hazardous environments. In particular, SafeGuard ASF \citep{canh2026safeguard} combines multi-modal hazard perception with agentic reasoning for real-time fire-risk detection and disaster recovery. Additionally, a robotic fire-risk detection system based on dynamic knowledge graphs and LLM-enhanced multi-modal reasoning is developed \citep{pan2025robotic}, demonstrating how FM-based reasoning can support emergency response in safety-critical settings.

\subsection{Maritime robotics}
\label{sup:ssec:application-domains_maritime}
The incorporation of FM intelligence has significantly bolstered the capabilities of autonomous maritime systems, allowing them to perceive, to reason, and to act more effectively within complex aquatic environments. These models are specifically engineered to navigate typical underwater challenges, such as high turbidity, limited visibility, and severe communication constraints that often degrade traditional robotic sensors. In particular, UnderwaterVLA \citep{wang2025underwatervla} introduces a dual-brain VLA architecture for autonomous underwater navigation, combining multi-modal reasoning with embodied control for improving robustness under degraded visual and communication conditions. Additionally, MarineInst \citep{zheng2024marineinst} and MarineGPT \citep{zheng2023marinegpt} demonstrate FM capabilities in bridging raw marine visual data, semantic understanding, and domain-specific natural-language knowledge, thereby supporting richer perception and reasoning modules for maritime robotic platforms.

\subsection{Space robotics}
\label{sup:ssec:application-domains_space}
The integration of FMs is critically transforming the field of astro-embodied intelligence, enabling robotic agents to reason, to adapt, and to perceive within unstructured, off-world environments, where human intervention is physically impossible. These models provide strong priors and zero-shot generalization capabilities that are crucial for operating under the extreme conditions and data scarcity typical of planetary missions. A primary application involves the usage of SAM for universal crater detection \citep{giannakis2023deep}, which utilizes promptable segmentation to identify features across diverse planetary imagery without requiring domain-specific retraining. Beyond basic detection, such FMs are being extended to facilitate autonomous terrain understanding and complex geological analysis \citep{giannakis2023deep, zhao2024crater,holden2026vision}, allowing robots to make high-stake decisions independently in remote and hazardous space settings.

\section{Public datasets}
\label{sec:datasets_details}

This section provides the complete dataset catalog summarized in Section~\ref{sec:Datasets} of the main text. In particular, Tables \ref{tab:datasets_perception_updated}-\ref{tab:datasets_hri} group the various benchmarks with respect to the main robotic task concerned, while they also include information about the following aspects for each entry: a) Year, b) Scale, c) Semantic classes, d) Modalities involved, e) Annotation type, f) Domain, g) Environment type (Simulation (S), Real (R)), h) Temporality (Static (ST), Video (V), Sequential (SE)), i) Embodiment (Yes (Y), No (N)), and j) Short description. Apart from per task remarks, the following global observations can be made:
\begin{itemize}
    \item \underline{Focus on massive cross-embodiment datasets}: In order to address the need for constructing generalist policies that can perform robustly under different kinematic structures and environmental conditions, dataset creation activities have shifted from specialized, single-robot setups to large-scale, multi-robot, heterogeneous ones. For example, Open X-Embodiment \citep{openx} and AgiBot World \citep{bu2025agibot} contain over a million trajectories across dozens of robotic platforms.
    \item \underline{Emphasis on vision-language-action capturings}: Following the research trend of developing embodied VLA solutions, data collection procedures increasingly support sensorial types that aim at integrating perception, reasoning, and control actions. For example, BridgeData V$2$ \citep{walke2023bridgedata} and CALVIN \citep{mees2022calvin} include synchronized information streams of RGB-D video, natural language instructions, and low-level action tokens.
    \item \underline{Translation to high-fidelity, real-world capturing settings}: In order to robustly address the sim-to-real gap in the FM era, intense efforts have been devoted on creating massive, real-world benchmarks, instead of simulation ones. In this respect, datasets like Ego$4$D \citep{grauman2022ego4d} and DROID \citep{khazatsky2024droid} support geographic and environmental diversity at unprecedented scale.
    \item \underline{Incorporation of human reasoning aspects}: In order to enable FMs to develop common-sense, human-like reasoning and interaction capabilities, such aspects are increasingly incorporated in the dataset formation procedures. For instance, RoboVQA \citep{sermanet2024robovqa} and TEACh \citep{padmakumar2022teach} include hundreds of thousands of video-text pairs and clarification-oriented Q\&A dialogues. 
    \item \underline{Lack of tactile sensing data}: Despite the high availability of visual and textual data, tactile (touch) and force-sensing information streams remain critically under-represented in the current benchmarks. Incorporation of the aforementioned modalities is essential for developing efficient, high-precision dexterous manipulation. 
    \item \underline{Lack of failure and recovery data}: A common observation across all datasets comprises the typical lack of recordings corresponding to rare failure modes and successful recovery actions. The latter is also essential during the training phase for developing models that are likely to be robust in real-world execution settings.
\end{itemize}

\begin{table}[!h]
\caption{Public datasets concerning in principle perception tasks. Abbreviations used: Environment type (Simulation (S), Real (R)), Temporality (Static (ST), Video (V), Sequential (SE)), and Embodiment (Yes (Y), No (N)).}
\label{tab:datasets_perception_updated}
\centering
\scriptsize

\setlength{\aboverulesep}{0pt}
\setlength{\belowrulesep}{0pt}
\setlength{\tabcolsep}{2pt}
\renewcommand{\arraystretch}{0.9}

\setlist*[tabitem]{before=\vspace{2.2pt}\justifying, after=\vspace{2.2pt}}

\rowcolors{2}{gray!25}{white}

\newlength{\Wds}\setlength{\Wds}{1.5cm}
\newlength{\Wyr}\setlength{\Wyr}{0.78cm}
\newlength{\Wscale}\setlength{\Wscale}{2.5cm}
\newlength{\Wclass}\setlength{\Wclass}{1.5cm}
\newlength{\Wmoda}\setlength{\Wmoda}{1.7cm}
\newlength{\Wann}\setlength{\Wann}{2.5cm}
\newlength{\Wdom}\setlength{\Wdom}{1.25cm}
\newlength{\Wenv}\setlength{\Wenv}{0.7cm}
\newlength{\Wtemp}\setlength{\Wtemp}{0.98cm}
\setlength{\Wemb}{0.86cm}
\newlength{\Wdesc}\setlength{\Wdesc}{3.0cm}

\resizebox{\textwidth}{!}{%
\begin{tabular}{@{}|
  >{\justifying\arraybackslash}m{\Wds}|
  >{\centering\arraybackslash}m{\Wyr}|
  >{\raggedright\arraybackslash}m{\Wscale}|
  >{\raggedright\arraybackslash}m{\Wclass}|
  >{\raggedright\arraybackslash}m{\Wmoda}|
  >{\raggedright\arraybackslash}m{\Wann}|
  >{\centering\arraybackslash}m{\Wdom}|
  >{\centering\arraybackslash}m{\Wenv}|
  >{\centering\arraybackslash}m{\Wtemp}|
  >{\centering\arraybackslash}m{\Wemb}|
  >{\justifying\arraybackslash}m{\Wdesc}|
@{}}
\toprule
\rowcolor{gray!40}
\headerbreak{Dataset} &
\headerbreak{Year} &
\headerbreak{Scale} &
\headerbreak{Classes} &
\headerbreak{Modalities} &
\headerbreak{Annotation\\type} &
\headerbreak{Domain} &
\headerbreak{Env.\\type} &
\headerbreak{Temp.} &
\headerbreak{Emb.} &
\headerbreak{Description} \\
\midrule

\textbf{Matter-port3D} \citep{chang2017matterport3d} & 
2017 & 
\begin{tabitem}
\item 90 building scenes
\item 10{,}800 panoramas
\item 194{,}400 RGB-D images
\item 50{,}811 object instances
\item 38{,}328 3D bounding boxes
\end{tabitem} & 
\begin{tabitem}
\item 20 structural
\item 20 objects
\end{tabitem} & 
\begin{tabitem}
\item RGB-D panoramas
\item 3D meshes
\item Camera poses
\end{tabitem} & 
\begin{tabitem}
\item 3D reconstruction
\item 2D masks
\item 3D semantic \& instance labels
\end{tabitem} & 
Indoors & 
R & 
ST & 
N & 
RGB-D dataset of entire buildings for global scene understanding \\ 
\midrule

\textbf{2D-3D-S} \citep{armeni2017joint} & 
2017 & 
\begin{tabitem}
\item 6K $m^2$ coverage
\item 270 rooms / 6 areas
\item 70{,}496 regular RGB images
\item 1{,}413 equirectangular RGB images
\item 695M 3D points
\end{tabitem} & 
\begin{tabitem}
\item 7 structural
\item 6 objects
\end{tabitem} & 
\begin{tabitem}
\item RGB-D images
\item Surface normals
\item 3D meshes \& point clouds
\item Camera poses \& metadata
\end{tabitem} & 
\begin{tabitem}
\item 2D \& 3D semantic \& instance labels
\item 3D bounding boxes
\item Room categorizations
\end{tabitem} & 
Indoors & 
R & 
ST & 
N & 
Jointly registered 2D and 3D data for office-style indoor scene understanding \\ 
\midrule

\textbf{Semantic--KITTI} \citep{behley2019semantickitti} & 
2019 & 
\begin{tabitem}
\item 22 sequences
\item 43{,}552 LiDAR scans
\item 360$^{\circ}$ FOV coverage
\end{tabitem} & 
\begin{tabitem}
\item 11 structural
\item 9 objects
\item 8 actors
\end{tabitem} & 
\begin{tabitem}
\item LiDAR point clouds
\item 3D point trajectories
\item Sensor poses
\end{tabitem} & 
\begin{tabitem}
\item Point-wise semantic labels
\item Sequence-level ID tracking
\item Scene completion targets
\end{tabitem} & 
Driving & 
R & 
SE & 
Y & 
Large-scale LiDAR driving dataset with point-level semantic labels for sequences \\ 
\midrule

\textbf{BDD100K} \citep{yu2020bdd100k} & 
2020 & 
\begin{tabitem}
\item 100K HD video clips
\item 40M video frames
\item 100M total distance (km)
\item 10 vision tasks
\end{tabitem} & 
\begin{tabitem}
\item 11 structural
\item 8 objects
\end{tabitem} & 
\begin{tabitem}
\item RGB video
\item GPS/IMU metadata
\item Time/ Weather/ Lighting tags
\end{tabitem} & 
\begin{tabitem}
\item 2D bounding boxes
\item Instance \& driveable masks
\item Lane marking
\item Object tracking
\end{tabitem} & 
Driving & 
R & 
V & 
Y & 
Diverse driving dataset covering varying weather, times, and city environments \\ 
\midrule

\textbf{nuScenes} \citep{caesar2020nuscenes} & 
2020 & 
\begin{tabitem}
\item 1K scenes
\item 1.4M images
\item 390K LiDAR
\item 1.4M radar sweeps
\item 1.4M 3D object boxes
\end{tabitem} & 
\begin{tabitem}
\item 9 structural
\item 23 objects
\end{tabitem} & 
\begin{tabitem}
\item 6 RGB cams (360$^{\circ}$)
\item 32-beam LiDAR
\item 5 long-range radars
\item GPS/IMU metadata
\end{tabitem} & 
\begin{tabitem}
\item 3D bounding boxes
\item LiDAR semantic labels
\end{tabitem} & 
Driving & 
R & 
SE & 
N & 
Driving dataset providing full AV sensor suite (LiDAR, radar, 6 cameras) \\ 
\midrule

\textbf{Waymo Open} \citep{sun2020scalability} & 
2020 & 
\begin{tabitem}
\item 1K sequences
\item 200K LiDAR frames
\item 1M RGB images
\item 12M 3D boxes
\item 12.6M 2D boxes
\end{tabitem} & 
\begin{tabitem}
\item 15 structural
\item 13 objects
\item 8 actors
\end{tabitem} & 
\begin{tabitem}
\item 5 LiDAR
\item 5 RGB cameras
\item Sensor poses
\end{tabitem} & 
\begin{tabitem}
\item 2D/3D tracking boxes
\item Global 3D point IDs
\item Cross-camera 2D labels
\end{tabitem} & 
Driving & 
R & 
SE & 
N & 
High-resolution, multi-sensor driving data focused on perception and motion prediction \\ 
\midrule

\textbf{Ego4D} \citep{grauman2022ego4d} & 
2022 & 
\begin{tabitem}
\item 3{,}670 hours of video
\item 931 participants
\item 74 locations
\item 9 countries
\item 5M+ episodic annotations
\end{tabitem} & 
\begin{tabitem}
\item N/A (open)
\end{tabitem} & 
\begin{tabitem}
\item Egocentric RGB video
\item Multi-channel audio
\item 3D eye gaze
\item IMU
\item Stereo
\end{tabitem} & 
\begin{tabitem}
\item Episodic memory tags
\item Hand-object interactions
\item Audio-visual diarization
\item Social interaction labels
\end{tabitem} & 
Daily life & 
R & 
V & 
N & 
Large-scale egocentric video captured by hundreds of people worldwide \\ 
\midrule

\textbf{HM3D-SEM} \citep{yadav2023habitat} & 
2022 & 
\begin{tabitem}
\item 216 3D spaces
\item 3{,}100 rooms
\item 142{,}646 object instances
\item 14{,}200+ human hours
\end{tabitem} & 
\begin{tabitem}
\item 12 structural
\item 28 objects
\end{tabitem} & 
\begin{tabitem}
\item Textured 3D meshes
\item Semantic textures
\item Room-level metadata
\end{tabitem} & 
\begin{tabitem}
\item Pixel-level semantics
\item 40 Matterport categories
\item Instance-level labels
\end{tabitem} & 
Indoors & 
R & 
ST & 
N & 
Dataset of semantically annotated building-scale 3D indoor reconstructions \\ 
\midrule

\textbf{ScanNet++} \citep{yeshwanth2023scannet++} & 
2023 & 
\begin{tabitem}
\item 460 scenes
\item 1{,}858 laser scans
\item 280K 33MP DSLR images
\item 3.7M RGB-D frames
\end{tabitem} & 
\begin{tabitem}
\item 50 structural
\item 950+ objects
\end{tabitem} & 
\begin{tabitem}
\item RGB-D video
\item 3D meshes
\item Camera poses
\end{tabitem} & 
\begin{tabitem}
\item 3D semantic \& instance labels
\item Multi-label ambiguity tags
\end{tabitem} & 
Indoors & 
R & 
ST & 
N & 
Sub-millimeter fidelity indoor scans paired with high-resolution 33MP DSLR imagery \\ 
\bottomrule
\end{tabular}%
}
\end{table}

\begin{table}[!h]
\caption{Public datasets concerning in principle planning tasks. Abbreviations used: Environment type (Simulation (S), Real (R)), Temporality (Static (ST), Video (V), Sequential (SE)), and Embodiment (Yes (Y), No (N)).}
\label{tab:datasets_planning_final}
\centering
\scriptsize

\setlength{\aboverulesep}{0pt}
\setlength{\belowrulesep}{0pt}
\setlength{\tabcolsep}{2pt}
\renewcommand{\arraystretch}{0.9}

\setlist*[tabitem]{before=\vspace{2.2pt}\justifying, after=\vspace{2.2pt}}

\rowcolors{2}{gray!25}{white}

\setlength{\Wds}{1.5cm}
\setlength{\Wyr}{0.78cm}
\setlength{\Wscale}{2.5cm}
\setlength{\Wclass}{1.5cm}
\setlength{\Wmoda}{1.7cm}
\setlength{\Wann}{2.5cm}
\setlength{\Wdom}{1.25cm}
\setlength{\Wenv}{0.7cm}
\setlength{\Wtemp}{0.98cm}
\setlength{\Wemb}{0.86cm}
\setlength{\Wdesc}{3.0cm}

\resizebox{\textwidth}{!}{%
\begin{tabular}{@{}|
  >{\justifying\arraybackslash}m{\Wds}|
  >{\centering\arraybackslash}m{\Wyr}|
  >{\raggedright\arraybackslash}m{\Wscale}|
  >{\raggedright\arraybackslash}m{\Wclass}|
  >{\raggedright\arraybackslash}m{\Wmoda}|
  >{\raggedright\arraybackslash}m{\Wann}|
  >{\centering\arraybackslash}m{\Wdom}|
  >{\centering\arraybackslash}m{\Wenv}|
  >{\centering\arraybackslash}m{\Wtemp}|
  >{\centering\arraybackslash}m{\Wemb}|
  >{\justifying\arraybackslash}m{\Wdesc}|
@{}}
\toprule
\rowcolor{gray!40}
\headerbreak{Dataset} &
\headerbreak{Year} &
\headerbreak{Scale} &
\headerbreak{Classes} &
\headerbreak{Modalities} &
\headerbreak{Annotation\\type} &
\headerbreak{Domain} &
\headerbreak{Env.\\type} &
\headerbreak{Temp.} &
\headerbreak{Emb.} &
\headerbreak{Description} \\
\midrule

\textbf{AI2-THOR} \citep{kolve2017ai2} & 
2017 & 
\begin{tabitem}
\item 120 room scenes
\item 4 room categories
\item 3{,}578 interactive objects
\end{tabitem} & 
\begin{tabitem}
\item 100+ object types
\end{tabitem} & 
\begin{tabitem}
\item Egocentric RGB-D videos
\item Object metadata
\end{tabitem} & 
\begin{tabitem}
\item Object state changes
\item Navigation actions
\item Arm manipulation
\end{tabitem} & 
House-hold & 
S & 
SE & 
Y & 
Near photo-realistic 3D environments for visual AI agents navigation and interaction \\ 
\midrule

\textbf{Virtual-Home} \citep{puig2018virtualhome} & 
2018 & 
\begin{tabitem}
\item 2{,}821 action programs
\item 6 furnished houses
\item 357 objects per house
\end{tabitem} & 
\begin{tabitem}
\item $\sim$300 objects
\item 70 actions
\end{tabitem} & 
\begin{tabitem}
\item Natural language
\item Synthetic video
\item 3D poses
\end{tabitem} & 
\begin{tabitem}
\item Action programs
\item Timestamps
\item Atomic interactions
\end{tabitem} & 
House-hold & 
S & 
SE & 
Y & 
Simulation of complex daily activities via executable programs and scripts \\ 
\midrule

\textbf{BabyAI} \citep{chevalier2018babyai} & 
2018 & 
\begin{tabitem}
\item 19 difficulty levels
\item $2.48\times10^{19}$ instructions
\end{tabitem} & 
\begin{tabitem}
\item 6 object types
\end{tabitem} & 
\begin{tabitem}
\item Synthetic language
\item 2D grid
\end{tabitem} & 
\begin{tabitem}
\item Expert demonstrations
\item Sub-goal decompositions
\end{tabitem} & 
Indoors & 
S & 
SE & 
Y & 
Dataset focused on sample efficiency and grounded language learning in grid worlds \\ 
\midrule

\textbf{ALFRED} \citep{shridhar2020alfred} & 
2020 & 
\begin{tabitem}
\item 120 indoor scenes
\item 25{,}743 language directives
\item 8{,}055 expert demos
\end{tabitem} & 
\begin{tabitem}
\item 7 task types
\item 80 objects
\end{tabitem} & 
\begin{tabitem}
\item Egocentric RGB videos
\item Natural language
\end{tabitem} & 
\begin{tabitem}
\item High/low-level instructions
\item Action sequences
\item Pixel-wise interaction masks
\end{tabitem} & 
House-hold & 
S & 
V,SE & 
Y & 
Mapping of natural language to sequences of actions for visual AI agents \\ 
\midrule

\textbf{CALVIN} \citep{mees2022calvin} & 
2022 & 
\begin{tabitem}
\item 4 environments
\item 389 instructions
\end{tabitem} & 
\begin{tabitem}
\item 34 tasks
\end{tabitem} & 
\begin{tabitem}
\item RGB-D videos
\item Vision-based tactile
\item Proprioce-ption
\item Natural language
\end{tabitem} & 
\begin{tabitem}
\item Language goals
\item Pre-task locomotion behavior
\item Precomputed MiniLM embeddings
\end{tabitem} & 
Tabletop & 
S & 
SE & 
Y & 
Long-horizon, language-conditioned policy learning for continuous control \\ 
\midrule

\textbf{BEHAV-IOR-1K} \citep{li2023behavior} & 
2024 & 
\begin{tabitem}
\item 50 scenes
\end{tabitem} & 
\begin{tabitem}
\item 1K activities
\item 1.2K objects
\end{tabitem} & 
\begin{tabitem}
\item Egocentric RGB-D videos
\item Proprioce-ption
\end{tabitem} & 
\begin{tabitem}
\item Logic (BDDL)
\item Transition rules
\item Semantic properties
\end{tabitem} & 
House-hold & 
S & 
SE & 
Y & 
Human-centered activities with realistic physics and state changes \\ 
\midrule

\textbf{LAMBDA} \citep{jaafar2025lambda} & 
2024 & 
\begin{tabitem}
\item 31 rooms
\item 8 environments
\end{tabitem} & 
\begin{tabitem}
\item 571 tasks
\end{tabitem} & 
\begin{tabitem}
\item Natural language
\item Egocentric RGB-D videos
\item Robot poses \& actions
\end{tabitem} & 
\begin{tabitem}
\item Human-collected trajectories
\item Semantic maps
\end{tabitem} & 
House-hold & 
S,R & 
SE & 
Y & 
Focus on data-efficiency for multi-room, multi-floor mobile manipulation \\ 
\bottomrule
\end{tabular}%
}
\end{table}

\begin{table}[!h]
\caption{Public datasets concerning in principle navigation tasks. Abbreviations used: Environment type (Simulation (S), Real (R)), Temporality (Static (ST), Video (V), Sequential (SE)), and Embodiment (Yes (Y), No (N)).}
\label{tab:datasets_navigation}
\centering
\scriptsize

\setlength{\aboverulesep}{0pt}
\setlength{\belowrulesep}{0pt}
\setlength{\tabcolsep}{2pt}
\renewcommand{\arraystretch}{0.9}

\setlist*[tabitem]{before=\vspace{2.2pt}\justifying, after=\vspace{2.2pt}}

\rowcolors{2}{gray!25}{white}

\setlength{\Wds}{1.5cm}
\setlength{\Wyr}{0.78cm}
\setlength{\Wscale}{2.5cm}
\setlength{\Wclass}{1.5cm}
\setlength{\Wmoda}{1.7cm}
\setlength{\Wann}{2.5cm}
\setlength{\Wdom}{1.25cm}
\setlength{\Wenv}{0.7cm}
\setlength{\Wtemp}{0.98cm}
\setlength{\Wemb}{0.86cm}
\setlength{\Wdesc}{3.0cm}

\resizebox{\textwidth}{!}{%
\begin{tabular}{@{}|
  >{\justifying\arraybackslash}m{\Wds}|
  >{\centering\arraybackslash}m{\Wyr}|
  >{\raggedright\arraybackslash}m{\Wscale}|
  >{\raggedright\arraybackslash}m{\Wclass}|
  >{\raggedright\arraybackslash}m{\Wmoda}|
  >{\raggedright\arraybackslash}m{\Wann}|
  >{\centering\arraybackslash}m{\Wdom}|
  >{\centering\arraybackslash}m{\Wenv}|
  >{\centering\arraybackslash}m{\Wtemp}|
  >{\centering\arraybackslash}m{\Wemb}|
  >{\justifying\arraybackslash}m{\Wdesc}|
@{}}
\toprule
\rowcolor{gray!40}
\headerbreak{Dataset} &
\headerbreak{Year} &
\headerbreak{Scale} &
\headerbreak{Classes} &
\headerbreak{Modalities} &
\headerbreak{Annotation\\type} &
\headerbreak{Domain} &
\headerbreak{Env.\\type} &
\headerbreak{Temp.} &
\headerbreak{Emb.} &
\headerbreak{Description} \\
\midrule

\textbf{REVERIE} \citep{qi2020reverie} & 
2020 & 
\begin{tabitem}
\item 10{,}567 panoramas
\item 86 scenes
\item 23{,}536 instructions
\item 4{,}140 objects
\end{tabitem} & 
\begin{tabitem}
\item 489 objects
\end{tabitem} & 
\begin{tabitem}
\item Natural language
\item RGB panoramas
\end{tabitem} & 
\begin{tabitem}
\item Instructions
\item Target boxes
\item Nav-graph sequences
\end{tabitem} & 
Indoors & 
S & 
SE & 
Y & 
Remote embodied visual referring expressions in unseen environments \\ 
\midrule

\textbf{VLN-CE} \citep{krantz2020beyond} & 
2020 & 
\begin{tabitem}
\item 90 scenes
\item 7{,}189 trajectories
\item 21{,}567 instructions
\end{tabitem} & 
\begin{tabitem}
\item N/A (path-based)
\end{tabitem} & 
\begin{tabitem}
\item Natural language
\item Egocentric RGB-D videos
\end{tabitem} & 
\begin{tabitem}
\item Low-level continuous actions
\item Navigation paths
\end{tabitem} & 
Indoors & 
S & 
SE & 
Y & 
Navigation in continuous environments, emphasizing fine-grained control \\ 
\midrule

\textbf{RxR} \citep{ku2020room} & 
2020 & 
\begin{tabitem}
\item 126{,}069 instructions
\item 16.5M total words
\item 3 languages
\end{tabitem} & 
\begin{tabitem}
\item N/A (path-based)
\end{tabitem} & 
\begin{tabitem}
\item Multilingual text
\item Virtual pose traces
\item RGB panoramas
\end{tabitem} & 
\begin{tabitem}
\item Dense spatiotemporal grounding
\item Time-aligned text-to-pose
\end{tabitem} & 
Indoors & 
S & 
SE & 
Y & 
Large-scale multilingual vision-and-language navigation \\ 
\midrule

\textbf{Habitat-Web} \citep{ramrakhya2022habitat} & 
2022 & 
\begin{tabitem}
\item 80K navigation trajectories
\item 12K Pick\&Place trajectories
\item 29.3M actions
\item 22{,}600 hours
\end{tabitem} & 
\begin{tabitem}
\item 21 object trajectories
\end{tabitem} & 
\begin{tabitem}
\item Egocentric RGB-D videos
\item Teleopera-tion traces
\end{tabitem} & 
\begin{tabitem}
\item Human task trajectories
\item Implicit search heuristics
\end{tabitem} & 
Indoors & 
S & 
SE & 
Y & 
Imitation learning from large-scale human demonstrations collected on the web \\ 
\midrule

\textbf{ScaleVLN} \citep{wang2023scaling} & 
2023 & 
\begin{tabitem}
\item 1.2K+ scenes
\item 4.9M trajectories
\end{tabitem} & 
\begin{tabitem}
\item N/A (path-based)
\end{tabitem} & 
\begin{tabitem}
\item Synthesized language
\item Egocentric RGB-D videos
\end{tabitem} & 
\begin{tabitem}
\item Synthesized trajectory-instruction pairs
\end{tabitem} & 
Indoors & 
S & 
SE & 
Y & 
Navigation using automatically generated large-scale synthetic data \\ 
\midrule

\textbf{GOAT-Bench} \citep{khanna2024goat} & 
2024 & 
\begin{tabitem}
\item 90 scenes
\item 9 tasks
\item 3K+ goal entities
\end{tabitem} & 
\begin{tabitem}
\item N/A (open-vocabulary)
\end{tabitem} & 
\begin{tabitem}
\item Multi-modal goals
\item Egocentric RGB-D videos
\end{tabitem} & 
\begin{tabitem}
\item Target object locations
\item Sequential goal sequences
\end{tabitem} & 
Indoors & 
S & 
SE & 
Y & 
Lifelong navigation to open-vocabulary goals \\ 
\midrule

\textbf{HM3D-OVON} \citep{yokoyama2024hm3d} & 
2024 & 
\begin{tabitem}
\item 216 scenes
\item 15K+ object instances
\end{tabitem} & 
\begin{tabitem}
\item 379 objects
\end{tabitem} & 
\begin{tabitem}
\item Free-form language
\item Egocentric RGB-D videos
\end{tabitem} & 
\begin{tabitem}
\item Open-vocabulary object goals
\item 3D bounding boxes
\end{tabitem} & 
Indoors & 
S & 
SE & 
Y & 
Open-vocabulary object goal navigation, based on free-form natural language \\ 
\bottomrule
\end{tabular}%
}
\end{table}

\begin{table}[!h]
\caption{Public datasets concerning in principle manipulation tasks. Abbreviations used: Environment type (Simulation (S), Real (R)), Temporality (Static (ST), Video (V), Sequential (SE)), and Embodiment (Yes (Y), No (N)).}
\label{tab:datasets_manipulation}
\centering
\scriptsize

\setlength{\aboverulesep}{0pt}
\setlength{\belowrulesep}{0pt}
\setlength{\tabcolsep}{2pt}
\renewcommand{\arraystretch}{0.9}

\setlist*[tabitem]{before=\vspace{2.2pt}\justifying, after=\vspace{2.2pt}}

\rowcolors{2}{gray!25}{white}

\setlength{\Wds}{1.5cm}
\setlength{\Wyr}{0.78cm}
\setlength{\Wscale}{2.5cm}
\setlength{\Wclass}{1.5cm}
\setlength{\Wmoda}{1.7cm}
\setlength{\Wann}{2.5cm}
\setlength{\Wdom}{1.25cm}
\setlength{\Wenv}{0.7cm}
\setlength{\Wtemp}{0.98cm}
\setlength{\Wemb}{0.86cm}
\setlength{\Wdesc}{3.0cm}
\resizebox{\textwidth}{!}{%
\begin{tabular}{@{}|
  >{\justifying\arraybackslash}m{\Wds}|
  >{\centering\arraybackslash}m{\Wyr}|
  >{\raggedright\arraybackslash}m{\Wscale}|
  >{\raggedright\arraybackslash}m{\Wclass}|
  >{\raggedright\arraybackslash}m{\Wmoda}|
  >{\raggedright\arraybackslash}m{\Wann}|
  >{\centering\arraybackslash}m{\Wdom}|
  >{\centering\arraybackslash}m{\Wenv}|
  >{\centering\arraybackslash}m{\Wtemp}|
  >{\centering\arraybackslash}m{\Wemb}|
  >{\justifying\arraybackslash}m{\Wdesc}|
@{}}
\toprule
\rowcolor{gray!40}
\headerbreak{Dataset} &
\headerbreak{Year} &
\headerbreak{Scale} &
\headerbreak{Classes} &
\headerbreak{Modalities} &
\headerbreak{Annotation\\type} &
\headerbreak{Domain} &
\headerbreak{Env.\\type} &
\headerbreak{Temp.} &
\headerbreak{Emb.} &
\headerbreak{Description} \\
\midrule

\textbf{RoboNet} \citep{dasari2019robonet} & 
2019 & 
\begin{tabitem}
\item 15M frames
\item 162K trajectories
\item 4 locations
\end{tabitem} & 
\begin{tabitem}
\item N/A (pushing)
\end{tabitem} & 
\begin{tabitem}
\item RGB videos
\item Robot actions
\item Gripper states
\end{tabitem} & 
\begin{tabitem}
\item Action trajectories
\item Video targets
\end{tabitem} & 
Tabletop & 
R & 
SE & 
Y & 
Multi-robot dataset on learning visual foresight and video prediction for non-prehensile manipulation \\ 
\midrule

\textbf{RLBench} \citep{james2020rlbench} & 
2020 & 
\begin{tabitem}
\item Infinite demonstrations
\end{tabitem} & 
\begin{tabitem}
\item 100+ tasks
\end{tabitem} & 
\begin{tabitem}
\item RGB-D videos
\item Segmenta-tion
\item Proprio-ception
\end{tabitem} & 
\begin{tabitem}
\item Motion-planned trajectories
\item Target way-points
\end{tabitem} & 
Tabletop & 
S & 
SE & 
Y & 
Tasks algorithmically generated using ground-truth state information \\ 
\midrule

\textbf{CALVIN} \citep{mees2022calvin} & 
2022 & 
\begin{tabitem}
\item 4 environments
\item 2.4M interaction steps
\end{tabitem} & 
\begin{tabitem}
\item 34 tasks
\end{tabitem} & 
\begin{tabitem}
\item RGB-D videos
\item Tactile
\item Proprio-ception
\end{tabitem} & 
\begin{tabitem}
\item Language goals
\item Pre-task locomotion
\end{tabitem} & 
Tabletop & 
S & 
SE & 
Y & 
Long-horizon language-conditioned continuous control \\ 
\midrule

\textbf{LIBERO} \citep{liu2023libero} & 
2023 & 
\begin{tabitem}
\item 6.5K trajectories
\end{tabitem} & 
\begin{tabitem}
\item 130 tasks
\end{tabitem} & 
\begin{tabitem}
\item RGB video
\item Proprio-ception
\item Language instructions
\end{tabitem} & 
\begin{tabitem}
\item Expert demonstrations
\item Task completion tags
\end{tabitem} & 
Tabletop & 
S & 
SE & 
Y & 
Evaluation of knowledge transfer across sequentially learned task suites \\ 
\midrule

\textbf{RH20T} \citep{fang2024rh20t} & 
2023 & 
\begin{tabitem}
\item 110K trajectories
\end{tabitem} & 
\begin{tabitem}
\item 150+ skills
\end{tabitem} & 
\begin{tabitem}
\item RGB video
\item Force
\item Tactile
\item Audio
\end{tabitem} & 
\begin{tabitem}
\item Action trajectories
\item Language descriptions
\end{tabitem} & 
Tabletop & 
R & 
SE & 
Y & 
Multi-modal dataset including force and audio, targeting contact-rich skills \\ 
\midrule

\textbf{BridgeData V2} \citep{walke2023bridgedata} & 
2023 & 
\begin{tabitem}
\item 60{,}096 trajectories
\item 24 environments
\end{tabitem} & 
\begin{tabitem}
\item 13 skills
\end{tabitem} & 
\begin{tabitem}
\item RGB videos
\item Proprioception
\end{tabitem} & 
\begin{tabitem}
\item Goal images
\item Language instructions
\end{tabitem} & 
Tabletop & 
R & 
SE & 
Y & 
Use of low-cost robots across 24 environments to boost generalization \\ 
\midrule

\textbf{Open X-Embodiment} \citep{openx} & 
2024 & 
\begin{tabitem}
\item 1M+ episodes
\item 60 datasets
\end{tabitem} & 
\begin{tabitem}
\item 500+ skills
\end{tabitem} & 
\begin{tabitem}
\item RGB video
\item End-effector poses
\item Language instructions
\end{tabitem} & 
\begin{tabitem}
\item Action trajectories
\end{tabitem} & 
Multi-domain & 
R,S & 
SE & 
Y & 
Aggregation of 60+ datasets and 22 robotic platforms into a unified format for cross-embodiment scenarios \\ 
\midrule

\textbf{DROID} \citep{khazatsky2024droid} & 
2024 & 
\begin{tabitem}
\item 76K trajectories
\item 564 scenes
\end{tabitem} & 
\begin{tabitem}
\item 86 tasks
\end{tabitem} & 
\begin{tabitem}
\item RGB-D videos
\item Proprio-ception
\end{tabitem} & 
\begin{tabitem}
\item Teleoperated actions
\item Language instructions
\end{tabitem} & 
In-the-wild & 
R & 
SE & 
Y & 
In-the-wild dataset collected by 50 people across 52 buildings to maximize scene and lighting diversity \\ 
\midrule

\textbf{RoboMIND} \citep{wu2024robomind} & 
2024 & 
\begin{tabitem}
\item 107K trajectories
\item 96 objects
\end{tabitem} & 
\begin{tabitem}
\item 279 tasks
\end{tabitem} & 
\begin{tabitem}
\item RGB-D videos
\item Proprio-ception
\item Natural language
\end{tabitem} & 
\begin{tabitem}
\item Expert teleoperation
\item Failure demonstrations
\item Fine-grained instructions
\end{tabitem} & 
Indoors & 
S,R & 
SE & 
Y & 
Unified-standard dataset covering humanoids and dual-arm robots for multi-embodiment intelligence \\ 
\midrule

\textbf{AgiBot World} \citep{bu2025agibot} & 
2025 & 
\begin{tabitem}
\item 1{,}001{,}552 trajectories
\item 3K+ objects
\end{tabitem} & 
\begin{tabitem}
\item 217 tasks
\end{tabitem} & 
\begin{tabitem}
\item RGB-D video
\item Visuo-tactile
\item Proprio-ception
\end{tabitem} & 
\begin{tabitem}
\item Human-in-the-loop teleoperation actions
\end{tabitem} & 
Multi-domain & 
S,R & 
SE & 
Y & 
Large-scale facility-based platform using 100 humanoid robots to collect high-fidelity, bi-manual, long-horizon task data \\ 
\bottomrule
\end{tabular}%
}
\end{table}

\begin{table}[!h]
\caption{Public datasets concerning in principle human-robot interaction tasks. Abbreviations used: Environment type (Simulation (S), Real (R)), Temporality (Static (ST), Video (V), Sequential (SE)), and Embodiment (Yes (Y), No (N)).}
\label{tab:datasets_hri}
\centering
\scriptsize

\setlength{\aboverulesep}{0pt}
\setlength{\belowrulesep}{0pt}
\setlength{\tabcolsep}{2pt}
\renewcommand{\arraystretch}{0.9}

\setlist*[tabitem]{before=\vspace{2.2pt}\justifying, after=\vspace{2.2pt}}

\rowcolors{2}{gray!25}{white}

\setlength{\Wds}{1.5cm}
\setlength{\Wyr}{0.78cm}
\setlength{\Wscale}{2.5cm}
\setlength{\Wclass}{1.5cm}
\setlength{\Wmoda}{1.7cm}
\setlength{\Wann}{2.5cm}
\setlength{\Wdom}{1.25cm}
\setlength{\Wenv}{0.7cm}
\setlength{\Wtemp}{0.98cm}
\setlength{\Wemb}{0.86cm}
\setlength{\Wdesc}{3.0cm}

\resizebox{\textwidth}{!}{%
\begin{tabular}{@{}|
  >{\justifying\arraybackslash}m{\Wds}|
  >{\centering\arraybackslash}m{\Wyr}|
  >{\raggedright\arraybackslash}m{\Wscale}|
  >{\raggedright\arraybackslash}m{\Wclass}|
  >{\raggedright\arraybackslash}m{\Wmoda}|
  >{\raggedright\arraybackslash}m{\Wann}|
  >{\centering\arraybackslash}m{\Wdom}|
  >{\centering\arraybackslash}m{\Wenv}|
  >{\centering\arraybackslash}m{\Wtemp}|
  >{\centering\arraybackslash}m{\Wemb}|
  >{\justifying\arraybackslash}m{\Wdesc}|
@{}}
\toprule
\rowcolor{gray!40}
\headerbreak{Dataset} &
\headerbreak{Year} &
\headerbreak{Scale} &
\headerbreak{Classes} &
\headerbreak{Modalities} &
\headerbreak{Annotation\\type} &
\headerbreak{Domain} &
\headerbreak{Env.\\type} &
\headerbreak{Temp.} &
\headerbreak{Emb.} &
\headerbreak{Description} \\
\midrule

\textbf{CVDN} \citep{thomason2020vision} & 
2020 & 
\begin{tabitem}
\item 2{,}050 human dialogues
\item 7K+ trajectories
\item 83 scenes
\end{tabitem} & 
\begin{tabitem}
\item N/A (goal-driven)
\end{tabitem} & 
\begin{tabitem}
\item Natural language dialogues
\item RGB panoramas
\end{tabitem} & 
\begin{tabitem}
\item Dialogue history
\item Shortest-path actions
\item Navigation traces
\end{tabitem} & 
House-hold & 
S & 
SE & 
Y & 
Multi-turn human-human dialogues for navigation, where agents ask for help \\ 
\midrule

\textbf{TEACh} \citep{padmakumar2022teach} & 
2022 & 
\begin{tabitem}
\item 3{,}047 dialogues
\item 39.5K utterances
\end{tabitem} & 
\begin{tabitem}
\item 12 tasks
\end{tabitem} & 
\begin{tabitem}
\item Natural language dialogues
\item Egocentric RGB videos
\item Discrete actions
\end{tabitem} & 
\begin{tabitem}
\item Dialogue history
\item Human demonstrations
\item Object state changes
\end{tabitem} & 
House-hold & 
S & 
V,SE & 
Y & 
Task-driven agents that communicate to complete complex household tasks \\ 
\midrule

\textbf{DialFRED} \citep{gao2022dialfred} & 
2022 & 
\begin{tabitem}
\item 53K Q\&A pairs
\end{tabitem} & 
\begin{tabitem}
\item 25 sub-goals
\end{tabitem} & 
\begin{tabitem}
\item Natural language dialogues
\item Egocentric RGB videos
\end{tabitem} & 
\begin{tabitem}
\item Q\&A pairs
\item Action sequences
\item Oracle responses
\end{tabitem} & 
House-hold & 
S & 
V,SE & 
Y & 
Active questioning framework where agents ask humans for clarifications to solve household tasks \\ 
\midrule

\textbf{RoboVQA} \citep{sermanet2024robovqa} & 
2023 & 
\begin{tabitem}
\item 829{,}502 video-text pairs
\item 29{,}520 instructions
\end{tabitem} & 
\begin{tabitem}
\item N/A (open-ended QA)
\end{tabitem} & 
\begin{tabitem}
\item RGB videos
\item Natural language
\item Robot actions
\end{tabitem} & 
\begin{tabitem}
\item VQA pairs
\item Multi-embodiment demonstrations
\end{tabitem} & 
Daily life & 
R & 
V & 
Y & 
Large-scale reasoning dataset using interleaved vision-text-action for long-horizon robot planning \\ 
\midrule

\textbf{NatSGD} \citep{shrestha2024natsgd} & 
2024 & 
\begin{tabitem}
\item 1{,}143 commands
\item 18 participants
\end{tabitem} & 
\begin{tabitem}
\item 11 actions
\item 20 objects
\end{tabitem} & 
\begin{tabitem}
\item Speech
\item Audio
\item Gestures
\item Robot actions
\end{tabitem} & 
\begin{tabitem}
\item Intent labels
\item Time-aligned behavior
\end{tabitem} & 
Cooking & 
S,R & 
V,SE & 
Y & 
Synchronized speech, gestures, and robot demonstrations for natural human-robot interaction \\ 
\midrule

\textbf{HA-R2R} \citep{li2024human} & 
2024 & 
\begin{tabitem}
\item 21{,}567 instructions
\item 486 motion sequences
\end{tabitem} & 
\begin{tabitem}
\item 145 activities
\end{tabitem} & 
\begin{tabitem}
\item Natural language
\item RGB-D \& fisheye videos
\item Human activity data
\end{tabitem} & 
\begin{tabitem}
\item Human activity descriptions
\item Navigation trajectories
\end{tabitem} & 
Indoors & 
S,R & 
V,SE & 
Y & 
Human-aware navigation focusing on social constraints and dynamic human interactions \\ 
\bottomrule
\end{tabular}%
}
\end{table}

\end{document}

%% file: math_commands.tex
\usepackage{amsmath,amsfonts,bm}

\def\eqref#1{equation~\ref{#1}}

\def\1{\bm{1}}

\DeclareMathAlphabet{\mathsfit}{\encodingdefault}{\sfdefault}{m}{sl}
\SetMathAlphabet{\mathsfit}{bold}{\encodingdefault}{\sfdefault}{bx}{n}

